\PassOptionsToPackage{noend}{algpseudocode} % optional, if using options
\PassOptionsToPackage{x11names,dvipsnames}{xcolor}
\documentclass[pmlr,twocolumn,10pt]{jmlr} % W&CP article

\usepackage{booktabs}
\usepackage{siunitx}
\usepackage{enumitem}  
\usepackage{multirow} 
\usepackage[capitalise]{cleveref}

\usepackage{microtype}
\usepackage{bm}
\usepackage{algpseudocode}
\usepackage{algorithm}
\usepackage{adjustbox}

\usepackage[switch]{lineno}
\newcommand{\bftab}{\fontseries{b}\selectfont}

\theorembodyfont{\upshape}
\theoremheaderfont{\scshape}
\theorempostheader{:}
\theoremsep{\newline}

\jmlrvolume{287}
\jmlryear{2025}
\jmlrsubmitted{} % LEAVE UNSET
\jmlrpublished{} % LEAVE UNSET
\jmlrworkshop{Conference on Health, Inference, and Learning (CHIL) 2025}

\title[Causal Survival Analysis: The Impact of Medication Non-adherence on Adverse Outcomes]{The Impact of Medication Non-adherence on Adverse Outcomes: \linebreak Evidence from Schizophrenia Patients via Survival Analysis}
\author{%
\Name{Shahriar Noroozizadeh} \Email{snoroozi@cmu.edu}\\
\addr Carnegie Mellon University, USA 
\AND
\Name{Pim Welle} \Email{pdw@andrew.cmu.edu}\\
\addr Allegheny County Department of Human Services, USA \\
\addr Carnegie Mellon University, USA 
\AND
\Name{Jeremy C. Weiss} \Email{jeremy.weiss@nih.gov}\\
\addr National Institutes of Health, USA 
\AND
\Name{George H. Chen} \Email{georgechen@cmu.edu}\\
\addr Carnegie Mellon University, USA 
\vspace{-1.5em}
}

\begin{document}
% \vspace{-2cm}
\maketitle
\begin{abstract}
This study quantifies the association between non-adherence to antipsychotic medications and adverse outcomes in individuals with schizophrenia. 
We frame the problem using survival analysis, focusing on the time to the earliest of several adverse events (early death, involuntary hospitalization, jail booking).
We extend standard causal inference methods (T-learner, S-learner, nearest neighbor matching) to utilize various survival models to estimate individual and average treatment effects, where treatment corresponds to medication non-adherence. 
Analyses are repeated using different amounts of longitudinal information (3, 6, 9, and 12 months). 
Using data from Allegheny County in western Pennsylvania, we find strong evidence that non-adherence advances adverse outcomes by approximately~1 to~4 months. 
Ablation studies confirm that county-provided risk scores adjust for key confounders, as their removal amplifies the estimated effects. 
Subgroup analyses by medication formulation (injectable vs.~oral) and medication type consistently show that non-adherence is associated with earlier adverse events. 
These findings highlight the clinical importance of adherence in delaying psychiatric crises and show that integrating survival analysis with causal inference tools can yield policy-relevant insights. 
We caution that although we apply causal inference, we only make associative claims and discuss assumptions needed for causal interpretation.
\end{abstract}

% \vspace{-1em}
\paragraph*{Data and Code Availability}
This paper utilizes administrative data from the Allegheny County Department of Human Services in the United States. 
% \textcolor{red}{For the purposes of peer review, we have anonymized the specific county and the full department name.} The dataset consists of de-identified records, where part of the de-identification process involved coarsening the time resolution to the month level. %, %, where personally identifiable information (PII) has been removed,
% and its time resolution has been intentionally coarsened to the month level.
Due to confidentiality agreements, the dataset is not publicly available.

The implementation of our proposed method is available at our GitHub repository: \linebreak
\url{https://github.com/Shahriarnz14/causal-meta-learner-survival-analysis}.

% \vspace{-2em}
\paragraph*{Institutional Review Board (IRB)}
This study does not require IRB approval, as it conducts secondary analysis of de-identified data provided by the Allegheny County Department of Human Services.

\setlength{\abovedisplayskip}{3pt plus 1pt}
\setlength{\belowdisplayskip}{3pt plus 1pt}
\setlength{\abovedisplayshortskip}{2pt plus 1pt}
\setlength{\belowdisplayshortskip}{2pt plus 1pt}

% \vspace{-1em}
\section{Introduction}
\label{sec:intro}
% \vspace{-0.25em}
Mental health disorders remain a leading contributor to disability worldwide, incurring substantial individual and societal costs \citep{whiteford2013global, rehm2019global, gbd2022global}. 
Severe mental illnesses (SMIs) such as schizophrenia, schizoaffective disorder, and bipolar disorder pose particular challenges for long-term management. 
In these conditions, antipsychotic medications constitute a cornerstone of treatment, mitigating psychotic symptoms, reducing relapse risk, and facilitating social and functional stability \citep{yatham2018canadian, keepers2020american, leucht2012antipsychotic, tiihonen2017real}. 
Unfortunately, non-adherence to antipsychotic regimens is widespread and has been associated with a host of adverse outcomes, including psychiatric hospitalization, suicidality, homelessness, and involvement with the criminal justice system \citep{ascher2006medication, velligan2009expert, velligan2017psychiatric, correll2018comparison}.

A variety of factors, ranging from medication side effects and associated stigma to cognitive impairment and limited social support, can shape adherence behaviors \citep{kane2016comprehensive, semahegn2020psychotropic}. 
While it is well-established that non-adherence heightens risks of adverse outcomes, many studies focus on a single outcome (e.g., hospitalization), thereby overlooking broader, multifaceted challenges \citep{walter2019multiple, mok2024multiple}.
Given the substantial clinical and public health implications of concurrent risks such as mortality, involuntary hospitalization, and criminal justice involvement, a more comprehensive survival framework is needed \citep{chen2022influencing}.

A recent Allegheny County report has underscored rising mortality rates and the considerable burden of involuntary hospitalizations among individuals with SMIs \citep{welle2023involuntary}. 
Despite highlighting these issues, the report did not isolate or quantify how adherence influences such events.
Accordingly, there is a need for survival and causal inference methods that account for right-censoring, selection bias, and time-varying exposures in understanding how non-adherence might affect the timing of critical outcomes.
In the present study, we develop a rigorous, real-world application of these methods to examine the first occurrence of mortality, involuntary hospitalization, or jail booking in a de-identified county dataset.
Our approach leverages both established survival models and well-studied causal meta-learning techniques, offering a nuanced view of how non-adherence may exacerbate early adverse events in schizophrenia.

The population under study faces a high risk of severe adverse events within a short period.  
Among 6,827 individuals in our analysis, 19\% experienced an adverse event within the first year, which includes 595 (53\%) involuntary hospitalizations, 434 (38\%) jail bookings, and 100 (9\%) deaths. These figures highlight the urgency of figuring out who might be at risk of experiencing an adverse event and, subsequently, finding interventions for high-risk individuals that would lead to positive outcomes. Our underlying hypothesis is that medication non-adherence is an indicator for whether an individual will experience an adverse event.
To this end, we aim to rigorously quantify the association between medication non-adherence and adverse event timing. Our hope is that this quantification provides a first step toward future targeted interventions and policy decisions that benefit individuals with schizophrenia.

Our main contributions are summarized as follows:
\begin{enumerate}[itemsep=0pt]
    \item \textbf{Longitudinal survival analysis framework for a composite adverse outcome.}
    Unlike existing investigations that restrict attention to an endpoint defined by a single event, we define a composite adverse event comprised of mortality, involuntary hospitalization, or jail booking. We model the time-to-first-event for this composite outcome; we refer to this time duration as the \emph{adverse event time}. We measure this adverse event time starting from different time origins depending on how much of an individual's longitudinal data we get to observe. Our framework directly extends the longitudinal data analysis approach of \citet{fitzmaurice2009longitudinal} to survival analysis.
    \item \textbf{Adaptation of standard causal inference methods to survival analysis.}
    Building on meta-learner strategies (T-learner and S-learner from \citet{kunzel2019metalearners}) and nearest-neighbor matching from causal inference (e.g., \citealt{stuart2010matching}), we quantify the effect of medication non-adherence on differences in mean adverse event times.
    In contrast to prior causal survival analysis work that often focuses on proportional hazards models or that uses parametric assumptions, our analysis shows how meta-learners can estimate individual and average treatment effects when censoring and time-varying exposures arise.

    \item \textbf{Subgroup and sensitivity analyses.}
    To illuminate the role of medication adherence across specific treatment contexts, we conduct subgroup analyses by antipsychotic formulation (injectable vs.\ oral) and by specific medication type (based on generic drug name).
    In addition, we investigate how risk scores provided by the county, serving as proxies for unmeasured confounders, modify both survival predictions and treatment effect estimates.
    These analyses shed light on model robustness and underscore the importance of rich covariate information in observational studies.
\end{enumerate}
Overall, this work demonstrates how established survival and causal inference techniques can be combined to investigate multifaceted, policy-relevant questions about medication adherence in schizophrenia.
By leveraging de-identified administrative and pharmacy data, we show that non-adherence remains a robust risk factor for earlier occurrences of life-threatening and socially destabilizing outcomes.
These findings support the clinical importance of promoting medication adherence and contribute methodological guidance for future analyses that seek to address similar real-world complexities.

Our experimental framework is designed to reflect real-world complexities, where medication adherence is naturally non-randomized, making direct counterfactual comparisons infeasible due to confounding, selection bias, immortal time bias, and potential reverse causality. 
A naive comparison of adherent versus non-adherent groups can lead to biased estimates. 
By explicitly modeling adherence patterns and incorporating a composite adverse outcome, our study provides a more comprehensive and realistic assessment of a patient's risk of experiencing adverse outcomes due to medication non-adherence. 
This approach enables us to obtain estimates that are robust, interpretable, and actionable, equipping clinicians and policymakers with the evidence needed to make informed decisions about adherence-promoting interventions.

% \vspace{-1em}
\section{Related Work}
\label{sec:related_works}
% \vspace{-0.25em}
Survival analysis in healthcare has been widely studied, spanning classical statistical methods and modern machine learning.
The Cox Proportional Hazards (CoxPH) model \citep{cox1972regression} remains popular for its interpretability but assumes a log-linear relationship with proportional hazards.
To capture complex patterns, nonparametric tree-based methods like random survival forests (RSFs) \citep{ishwaran2008random} and causal survival forests (CSFs) \citep{cui2023estimating} account for nonlinearities and interactions without explicit model assumptions.
Deep learning models, such as DeepSurv \citep{katzman2018deepsurv} and DeepHit \citep{lee2018deephit}, offer flexible individual survival distribution estimation and improved feature learning. 
As the goal of our paper is not to develop a new survival model, our experiments later use a variety of survival models (CoxPH, RSF, CSF, DeepSurv, DeepHit). A key message of our experiments is that our findings are robust to the choice of survival model so long as the survival model used is sufficiently accurate.

A standard approach to analyzing longitudinal data emphasizes conditioning on the complete history of treatment assignments and covariates, thereby addressing time-dependent confounding and repeated measures \citep{fitzmaurice2009longitudinal, kennedy2019nonparametric}. 
In this framework, causal inference is performed at discrete time snapshots by conditioning on the observed adherence history up to each point, allowing for the estimation of time-specific treatment effects. 
Following these principles, we adopt a snapshot-based method in our analysis, conditioning on each individual's adherence and covariate trajectory at regular intervals and effectively performing causal inference at multiple points in time. 
This setup provides a straightforward approach for investigating how adherence patterns may relate to subsequent adverse outcomes under time-varying exposures.

Meta-learner strategies from causal inference \citep{kunzel2019metalearners} have recently been extended to survival analysis.
For example, \citet{bo2024meta} estimate conditional average treatment effects (CATEs) in a survival analysis setting, focusing on individualized survival probabilities at specific time points.
Their flexible framework captures heterogeneous treatment effects without restrictive parametric assumptions, making it well-suited for complex real-world survival data.
Similar to this earlier work, we also extend meta-learners to survival analysis but directly estimate individualized treatment effects (ITE) using differences in restricted mean adverse event times, providing a more interpretable measure of treatment impact over a predefined horizon.
This formulation better aligns with clinical and policy decision-making, where we suspect that absolute time differences (measured in months in our case) are more straightforward to interpret than probability estimates. 
Our work contributes to the broader goal of estimating heterogeneous treatment effects in time-to-event data without strong parametric assumptions.
Unlike standard survival models, which predict observed event times, our framework integrates survival function estimators into meta-learners, enabling principled estimation of counterfactual time-to-event distributions. This adaptation allows for robust causal inference in censored settings, offering greater interpretability and clinical relevance compared to hazard ratio-based methods that can be challenging to interpret clinically and may obscure absolute treatment benefits or harms \citep{hernan2010hazards, aalen2015does}.

Separately, in the realm of medication adherence for severe mental illness, prior work consistently shows that non-adherence contributes to adverse outcomes, including relapse and criminal justice involvement \citep{semahegn2020psychotropic, lin2022time, correll2018comparison}.
Nevertheless, many investigations restrict attention to an endpoint defined by a single event---often hospitalization---and rely on parametric frameworks that may fail when key assumptions do not hold \citep{campos2021measuring, tannous2024predictive}.
Consequently, existing studies sometimes struggle with modeling multifaceted outcomes or precisely isolating the causal impact of adherence in real-world datasets that have right-censoring, moderate sample sizes, and only partial confounder information.

In this paper, we build on these strands of research. Unlike prior studies that focus on an outcome defined by a single event or that focus on methodological innovation, we integrate various survival analysis models with causal inference estimators in a real-world dataset to assess the impact of medication adherence on a composite adverse endpoint (mortality, involuntary hospitalization, or jail booking).
We adapt standard meta-learners to estimate restricted mean adverse event time differences, demonstrating how widely used causal methods can be applied in survival contexts.
While our observational design does not permit definitive causal conclusions, the consistent sign of our Average Treatment Effect (ATE) estimates across multiple approaches suggests that adherence may delay adverse outcomes, highlighting its clinical benefits, informing policy-driven studies, and motivating further refinements in causal survival analysis.

% \vspace{-1em}
\section{Methods}
\label{sec:methods}
% \vspace{-0.25em}

% \vspace{-1em}
\subsection{Data Description}
\label{sec:methods_data}
% \vspace{-0.25em}
We leverage longitudinal data obtained from Allegheny County Department of Human Services.
The dataset is defined as:
\[
\mathcal{D} = \{(T_i, \delta_i, \{A_{it}\}_{t=1}^{M_i}, \{\mathbf{X}_{it}\}_{t=1}^{M_i}) \mid i = 1,2,\dots,N\},
\]
where we explain what $T_i$, $\delta_i$, $A_{it}$, and $\mathbf{X}_{it}$ refer to shortly. The cohort consists of \(N = \text{6,827}\) adult patients diagnosed with a schizophrenia, each indexed by \(i \in \{1, \dots, N\}\), who are observed on a monthly basis for up to \(M = 96\) months. 
For each patient \(i\), \(M_i \leq 96\) denotes the number of months of observation.

\paragraph{Outcome variables $\bm{(T_{i},\delta_{i})}$.}
For each training patient \(i\), if the patient experiences an adverse event, then the event indicator is set to $\delta_i = 1$. In this case, the observed time $T_i\in \{1,\dots,M_i\}$ is equal to the true adverse event time, defined to be the month in which the earliest adverse event occurs (any of mortality, involuntary hospitalization, or jail booking), measured starting from the month in which the patient first fills an antipsychotic medication (for example, if $\delta_i=1$ and $T_i=1$, then this means that the $i$-th training patient experienced an adverse event 1 month after their first recorded antipsychotic prescription fill). If the patient does not experience an adverse event, then $\delta_i = 0$, and the observed time $T_i$ is referred to as a \emph{censoring time}; the true adverse event time is unknown but is sometime \emph{after} the censoring time.

Classically, what we call the adverse event time is instead called the ``survival'' time, but we avoid this latter wording in most of our exposition since for our application, the adverse event is \emph{not} necessarily death.

\paragraph{Treatment variables $\bm{A_{it}}$.} 
We consider medication adherence to be the treatment and define it monthly using prescription refill records.
For patient \(i\) at month \(t\), we define the binary treatment indicator as:
\[
A_{it} =
\begin{cases}
1, & \text{if non-adherent} \\
   & \text{(} \leq\text{10 days of prescription coverage)}, \\
0, & \text{if adherent} \\
   & \text{(} >\text{10 days of prescription coverage)}.
\end{cases}
\]
We experimented with threshold values ranging from 6 to 24 days and found that our downstream results remained largely unchanged, with only minor differences across this range.

\paragraph{Covariates $\mathbf{X}_{\bm{it}}$.}  
At each month \(t\), patient \(i\) has a covariate vector \(\mathbf{X}_{it}\) that includes both static and time-varying features. 
The static time-independent demographic covariates \( \mathbf{X}_i^{\text{(static)}} \) include age, race, gender, ethnicity, and education level.
The county also provides time-varying individualized risk scores:
\[
\mathbf{R}_{it} = \big(r_{it}^{(1)}, r_{it}^{(2)}, r_{it}^{(3)}, r_{it}^{(4)}, r_{it}^{(5)}\big),
\]
with
\(
r_{it}^{(k)} = \Pr(\text{event}_k \mid \widetilde{\mathbf{X}}_{it}),
\)
for \(k \in \{1,2,3,4,5\}\),
where \(\text{event}_k \in \{\)mortality, jail booking, shelter entry, involuntary hospitalization, drug overdose\(\}\) within 12 months.
These risk scores are derived from a rich set of covariates \(\mathbf{\widetilde{X}}_{it}\) that we do \emph{not} get to observe directly due to limitations of what data the county can share with us; this richer set of covariates include administrative records such as prior interactions with the criminal justice system, past hospitalizations, and comorbidity burden.
In more detail, the $\mathbf{R}_{it}$ risk scores that we do have access to represent the predicted probabilities obtained from a logistic regression model trained on \(\mathbf{\widetilde{X}}_{it}\), where the outcome variable corresponds to the occurrence of each respective adverse event within the next 12 months. In short, even though we do not get to observe the richer \(\mathbf{\widetilde{X}}_{it}\) covariates, we observe the county-provided $\mathbf{R}_{it}$ risk scores that serve as proxies for the full covariate set.

In Appendix~\ref{apd:risk_score_evaluations}, we demonstrate that for mortality, jail booking, and involuntary hospitalization, the provided risk scores exhibit strong predictive performance. This suggests that these scores effectively capture the underlying unmeasured covariates that we do not have direct access to, reinforcing their validity as proxies within our dataset. 

% \vspace{-1em}
\subsection{Prediction Setup}
\label{sec:methods_prediction}
% \vspace{-0.25em}
We formulate a time-to-event prediction task using longitudinal patient data at snapshot times \(\tau \in \{3,6,9,12\}\) months, following a standard setup (see, e.g., Section 6.2 of \citet{chen2024introduction}). %, following the standard approach of \citet{fitzmaurice2009longitudinal}.
For each snapshot \(\tau\), patient information up to that time is used to predict the occurrence of an adverse event after time $\tau$ (we measure the adverse event time starting from time $\tau$). 

\paragraph{Cohort selection at each time snapshot.}  
At snapshot time \(\tau\), we include patients who have an observed time $T_i$ that is at least \(\tau+1\) months. 
Namely, the cohort for each \(\tau\) is defined as:
\[
\mathcal{D}_\tau = \{(T_i, \delta_i, \{A_{it}\}_{t=1}^{\tau}, \mathbf{X}_{i\tau}) \mid M_i \geq \tau+1\}.
\]

\paragraph{Feature construction.}  
At snapshot \(\tau\), the feature vector for patient \(i\) is constructed as follows:
\begin{enumerate}[label=(\roman*),itemsep=0pt]
    \item \text{Static Demographics:} \( \mathbf{X}_i^{\text{(static)}} \) (age, race, gender, education).
    \item \text{Medication Adherence History:} The sequence \(\mathbf{A}_{i,1:\tau-1} = (A_{i1}, A_{i2}, \dots, A_{i,\tau-1})\). (patient's longitudinal adherence pattern leading up to snapshot time \(\tau\)).
    \item \text{Risk Scores:} The county-provided risk scores \(\mathbf{R}_{i\tau}\) at time \(\tau\).
    \item \text{Current Adherence (Treatment  Indicator):} The non-adherence indicator \(A_{i\tau}\) at snapshot \(\tau\).
\end{enumerate}

We define the patient’s full feature vector at snapshot time \( \tau \) as:
\[
\mathbf{Z}_{i\tau} = \left[ \mathbf{X}^{\text{(static)}}_{i}, \mathbf{R}_{i\tau}, \mathbf{A}_{i,1:\tau-1} \right].
\]

\paragraph{Prediction task.}  
For a test patient with feature vector \(\mathbf{Z}_{\tau}\) at snapshot time \(\tau\), the goal is to predict the time until an adverse event occurs.  
Specifically, we define: 
\[\widetilde{T} = T - \tau,\]
where \(T\) is the ground-truth adverse event time, and 
\(\widetilde{T}\) represents the time to the event from \(\tau\).  
Thus, the prediction target is \(\widetilde{T}\), focusing solely on estimating the time to the adverse event.

% \vspace{-1em}
\subsection{Survival Analysis Models}
\label{sec:methods_survival}
% \vspace{-0.25em}
To estimate the adverse event time, we consider four survival analysis models that vary in their underlying assumptions and modeling flexibility. 
These models include both traditional statistical approaches and more recently developed deep learning methods and can be plugged into meta-learning causal inference approaches:
% Below, we describe the four survival models used in our experiments:
\begin{enumerate}[itemsep=0pt]
    \item \textbf{Cox Proportional Hazards (CoxPH)} \citep{cox1972regression} – A semiparametric model that assumes a log-linear relationship between covariates and the hazard function, with a time-invariant hazard ratio:
    \(
    h(t \mid \mathbf{Z}_{i\tau}) = h_0(t) \exp(\boldsymbol{\beta}^{\top} \mathbf{Z}_{i\tau}),
    \)
    where \( h_0(t) \) is the baseline hazard and \( \boldsymbol{\beta} \) the regression coefficients. CoxPH is widely used for its interpretability and computational efficiency.

    \item \textbf{Random Survival Forest (RSF)} \citep{ishwaran2008random} – A nonparametric ensemble method that extends decision trees to survival data. It constructs multiple survival trees and aggregates them to estimate the survival function:
    \(
    \widehat{S}(t \mid \mathbf{Z}_{i\tau}) = \frac{1}{B} \sum_{b=1}^{B} S^{(b)}(t \mid \mathbf{Z}_{i\tau}),
    \)
    where \( S^{(b)}(t \mid \mathbf{Z}_{i\tau}) \) is the survival function from the \( b \)-th tree. RSF captures non-linear relationships and complex interactions without parametric assumptions.

    \item \textbf{DeepSurv} \citep{katzman2018deepsurv} – A deep learning extension of CoxPH that replaces its linear assumption with a neural network:
    \(
    h(t \mid \mathbf{Z}_{i\tau}) = h_0(t) \exp(f_{\theta}(\mathbf{Z}_{i\tau})),
    \)
    where \( f_{\theta} \) is a neural network. DeepSurv learns non-linear covariate effects while maintaining the proportional hazards framework.

    \item \textbf{DeepHit} \citep{lee2018deephit} – A neural network-based model that directly estimates the probability mass function (PMF) of survival time:
    \(
    P(T = t \mid \mathbf{Z}_{i\tau}) = f_{\theta}(t, \mathbf{Z}_{i\tau}),
    \)
    where \( f_{\theta} \) predicts the likelihood of event occurrence at each time \( t \). DeepHit allows for modeling multimodal survival distributions and competing risks.
\end{enumerate}
Our experiments later also uses a Causal Survival Forest (CSF) \citep{cui2023estimating}, which explicitly accounts for treatment effects in survival analysis. However, since a CSF is inherently a causal model that estimates treatment effects, it is not meant for being plugged into a meta-learning causal inference framework. We defer discussing it to the next section, where we describe the causal inference setup we consider. 

% \vspace{-1em}
\subsection{Causal Inference Setup}
\label{sec:methods_causal}
% \vspace{-0.25em}
For our observational study, we adopt a causal inference framework to quantify the association between medication non-adherence and adverse event time.
Under the potential outcomes framework, these estimates approximate causal effects if unmeasured confounding is limited and positivity holds.
To help satisfy some of these assumptions, we apply preprocessing techniques such as trimming, detailed in Appendix~\ref{apd:preprocessing}.
In Appendix~\ref{adp:from_assoc_to_causal}, we outline the assumptions required for a valid causal interpretation, including those specific to survival analysis.

\paragraph{Restricted Mean Event Time (RMET).}  
RMET measures the expected time until an adverse event occurs, restricted to a predefined follow-up period.
Intuitively, it represents the average time a patient remains event-free within the observation window.\footnote{In survival analysis literature, this quantity is commonly referred to as the Restricted Mean Survival Time (RMST), but we adapt the terminology to emphasize its focus on adverse events.}
For patient \(i\), the RMET is defined as:
\[
\bar{T_i} = \mathbb{E}[\widetilde{T_i} | A_{i\tau}, \mathbf{Z}_{i\tau}] = \int_{0}^{M} {S}(u | A_{i\tau}, \mathbf{Z}_{i\tau}) \, du,
\]
where \( {S}(u | A_{i\tau}, \mathbf{Z}_{i\tau}) \) is the true survival function and \( M = 96 \) months denotes the maximum follow-up duration. In practice, we replace the true survival function with an estimated version using a survival model from Section~\ref{sec:methods_survival}.

\paragraph{Potential outcomes.} 
Under the potential outcomes framework, we denote the potential adverse event time under treatment \(a \in \{0,1\}\) by \(\widetilde{T_i}(a)\).
For potential outcomes, we can then define:
\[
\bar{T}_i(a) = \mu_a(\mathbf{Z}_{i\tau}) =\int_{0}^{M} S(u | A_{i\tau} = a, \mathbf{Z}_{i\tau})\, du,
\]
and the individual treatment effect (ITE) as:
\[
\text{ITE}_i = \bar{T}_i(1) - \bar{T}_i(0).
\]
By a standard derivation (see Appendix~\ref{apd:ate-ite}), the average treatment effect (ATE) is given by:
\(
\psi = \mathbb{E}\big[\bar{T}_i(1) - \bar{T}_i(0)\big].
\)
Finally, we can estimate the ATE empirically as:
\[
\widehat{\psi} = \frac{1}{N} \sum_{i=1}^{N}\left(\widehat{\mu}_1(\mathbf{Z}_{i\tau}) - \widehat{\mu}_0(\mathbf{Z}_{i\tau})\right),
\]
with
\(
\widehat{\mu}_a(\mathbf{Z}_{i\tau}) = \int_{0}^{M} \widehat{S}(u | A_{i\tau} = a, \mathbf{Z}_{i\tau})\, du.
\)

\paragraph{Causal estimators.}  
We extend meta-learning approaches (T-learner and S-learner from \citet{kunzel2019metalearners}) and matching methods to survival analysis, with algorithms provided in Appendix~\ref{apd:causal-meta-learner-algorithms}, introducing novel extensions for estimating treatment effects under censoring:
% Additionally, we employ causal survival forests as a nonparametric baseline.
\begin{itemize}[itemsep=0pt]
    \item \textbf{T-Learner:}
    Fit separate survival models for the treated (\(a=1\)) and control (\(a=0\)) groups. 
    The survival function is estimated independently for each treatment condition:
    \(
    \widehat{S}_a (u | \mathbf{Z}_{i\tau}) = {\widehat{S}(u | A_{i\tau} = a, \mathbf{Z}_{i\tau})},
    \)
    yielding two RMET estimators:
    \(
    \widehat{\mu}_a (\mathbf{Z}_{i\tau}) = \int_0^{M} \widehat{S}_a (u | \mathbf{Z}_{i\tau}) \, du,
    \)
    with \(a\in\{0,1\}\). The ATE is then estimated as:
    \[
    \widehat{\psi}_{\text{T-learner}} = \frac{1}{N} \sum_{i=1}^{N} \left( \widehat{\mu}_1 (\mathbf{Z}_{i\tau}) - \widehat{\mu}_0 (\mathbf{Z}_{i\tau}) \right).
    \]

    \item \textbf{S-Learner:} Fit a single survival model incorporating treatment as an additional covariate:
    \(
    \widehat{S}(u | \mathbf{Z}_{i\tau}, A_{i\tau}) = \widehat{S}(u | A_{i\tau}, \mathbf{Z}_{i\tau}).
    \)
    The single RMET estimator is then:
    \(
    \widehat{\mu} (\mathbf{Z}_{i\tau}, a) = \int_0^{M} \widehat{S}(u | \mathbf{Z}_{i\tau}, a) \, du,
    \)
    and the ATE is estimated by plugging \(a=0\text{ and } 1\) in the single estimator:
    \[
    \widehat{\psi}_{\text{S-learner}} = \frac{1}{N} \sum_{i=1}^{N} \left( \widehat{\mu} (\mathbf{Z}_{i\tau},1) - \widehat{\mu} (\mathbf{Z}_{i\tau},0) \right).
    \]

    \item \textbf{Matching (k-Nearest Neighbors)} \citep{stuart2010matching}\textbf{:}  
    Match individuals based on baseline covariates. 
    First set: \(
    \widehat{\mu}_{A_{i\tau}} (\mathbf{Z}_{i\tau}) = \widehat{\mu} (\mathbf{Z}_{i\tau}, A_{i\tau})
    \).
    Then for patient \(i\), let \(J_K(i)\) denote the set of \(K\) nearest neighbors from the opposite treatment group. The estimated counterfactual RMET is
    \[
    \widehat{\mu}_{1-A_{i\tau}}(\mathbf{Z}_{i\tau}) = \frac{1}{K} \sum_{j \in J_K(i)} \widehat{\mu}(\mathbf{Z}_{j\tau}, A_{j\tau}),
    \]
    and the ATE is estimated as:
    \footnotesize
    \[
    \widehat{\psi}_{\text{match}} = \frac{1}{N} \sum_{i=1}^{N} \left(\widehat{\mu}_{A_{i\tau}}(\mathbf{Z}_{i\tau}) - \widehat{\mu}_{1-A_{i\tau}}(\mathbf{Z}_{i\tau})\right)(2A_{i\tau} - 1).
    \]
    \normalsize
    We consider \(K \in \{1,5,20\}\).
    See Appendix~\ref{apd:matching_derivation} for a more detailed derivation.
    
    \item \textbf{Causal Survival Forest (CSF)} \citep{cui2023estimating}\textbf{:} A nonparametric method that estimates heterogeneous treatment effects in censored survival data:
    % CSF partitions the feature space into homogeneous subgroups and estimates ITEs within each partition:
    \[
    \widehat{\theta}_{\text{CSF}}(\mathbf{Z}_{i\tau}) = \widehat{\mu}_1 (\mathbf{Z}_{i\tau}) - \widehat{\mu}_0 (\mathbf{Z}_{i\tau}),
    \]
    with the ATE computed as:
    \[
    \widehat{\psi}_{\text{CSF}} = \frac{1}{N} \sum_{i=1}^{N} \widehat{\theta}_{\text{CSF}}(\mathbf{Z}_{i\tau}).
    \]
    (Appendix~\ref{apd:csf_ate_derivation} provides a more detailed derivation of CSF).
\end{itemize}

% \vspace{-1em}
\subsection{Evaluation Metrics}
\label{sec:eval_metrics}
% \vspace{-0.25em}
\begin{table*}[!ht]
    \setlength{\tabcolsep}{4pt}
    % \scriptsize
    \floatconts
      {tab:survival_3_6}
      {\caption{Prediction Performance Across Time Snapshots (3 and 6 months)}\vspace{-.75em}}
      {\adjustbox{max width=.9\textwidth}{\begin{tabular}{l  c c c @{\hspace{0.5cm}} c c c }
    \toprule
    \textbf{Snapshot Time} &\multicolumn{3}{c}{3 months} & \multicolumn{3}{c}{6 months} \\
    \midrule
     & \(C^{td}\) & IBS & \(\text{AUC}^{td}\) & \(C^{td}\) & IBS & \(\text{AUC}^{td}\) \\
    \midrule
    CoxPH & 0.639 ± 0.008 & 0.198 ± 0.004 & 0.681 ± 0.012 & 0.633 ± 0.011 & 0.189 ± 0.003 & 0.682 ± 0.011 \\
    Random Survival Forest & {\bftab 0.659} ± 0.008 & {\bftab 0.190} ± 0.003 & {\bftab 0.705} ± 0.010 & {\bftab 0.661} ± 0.012 & {\bftab 0.181} ± 0.004 & {\bftab 0.705} ± 0.011 \\
    DeepSurv & 0.646 ± 0.018 & 0.196 ± 0.005 & 0.689 ± 0.022 & 0.630 ± 0.010 & 0.190 ± 0.005 & 0.670 ± 0.010 \\
    DeepHit & 0.570 ± 0.036 & 0.240 ± 0.008 & 0.588 ± 0.044 & 0.575 ± 0.050 & 0.219 ± 0.018 & 0.613 ± 0.043 \\
    \bottomrule
    \end{tabular}}\vspace{-1em}}
\end{table*}

We assess the performance of survival models using three key evaluation metrics, computed on the test set across five independent experimental repeats. 
Each repeat involves a different randomized train-validation-test split (\(60/20/20\)). 
Reported values include the mean and standard deviation across repeats.

\paragraph{Survival Model Metrics.}  
The following metrics are used to evaluate time-to-event predictions:
\begin{enumerate}[label=(\roman*),itemsep=0pt]
    \item \textbf{Time-dependent Concordance Index (\(C^{td}\))} \citep{antolini2005time}:  
    Measures the model’s ability to correctly rank adverse event times over different time horizons.
    Higher values indicate better discriminative performance.
    
    \item \textbf{Integrated Brier Score (IBS)} \citep{graf1999assessment}:  
    Assesses the overall prediction accuracy survival probability by integrating the Brier score over time:
    \[
    \text{IBS} = \int_{0}^{M} \text{BS}(t)\, w(t)\, dt,
    \]
    where \(\text{BS}(t)\) is the Brier score at time \(t\) and \(w(t)\) is a weighting function. Lower values denote better calibration.
    
    \item \textbf{Time-dependent Area Under the Curve (\(\text{AUC}^{td}\))}: Quantifies the model’s discriminative ability by evaluating how well it distinguishes individuals experiencing events at different time points \citep{uno2007evaluating}. 
    We report the mean across all prediction time points. 
    A higher values suggest improved predictive performance.
\end{enumerate}

\paragraph{Causal Inference Evaluation.}  
At each snapshot \(\tau\), the ATE is estimated on the full cohort.
The survival models used for ATE estimation are trained on different randomized subsets across experimental repeats, employing cross-validation to avoid overfitting. 
The ATE is reported along with its standard deviation across experimental repeats. The standard deviation gives an empirical measure of uncertainty in the~ATE.

In our experimental setup, a negative ATE suggests that non-adherence is associated with an earlier occurrence of adverse events (in months), and a positive ATE implies a delay in adverse events occurring.

% \vspace{-1em}
\section{Experiments and Results}
\label{sec:experiments_results}
% \vspace{-0.25em}
\begin{table*}[!htp]
    \setlength{\tabcolsep}{4pt}
    % \footnotesize
    \floatconts
      {tab:table_ate_3_6}
      {\caption{Causal Inference ATE Estimates at Snapshots of 3 and 6 Months}\vspace{-.5em}}
      {\adjustbox{max width=.825\textwidth}{\begin{tabular}{l c c c c c }
    \toprule
    &T-learner & S-learner & Matching (1) & Matching (5) & Matching (20) \\
    \midrule
    \textit{(Snapshot: 3 months)} \\
    CoxPH & -3.524 ± 0.274 & -3.644 ± 0.393 & -4.245 ± 0.005 & -4.280 ± 0.003 & -4.406 ± 0.002 \\
    Random Survival Forest & -2.317 ± 0.602 & -1.183 ± 0.363 & -1.615 ± 0.073 & -1.538 ± 0.054 & -1.978 ± 0.050 \\
    DeepSurv & -2.366 ± 0.603 & -1.986 ± 0.382 & -2.160 ± 1.014 & -2.101 ± 1.051 & -2.312 ± 1.223 \\
    DeepHit & -2.956 ± 6.663 & 0.417 ± 0.967 & 0.215 ± 0.481 & 0.241 ± 0.476 & 0.200 ± 0.501 \\
    \midrule
    Causal Survival Forest &  \multicolumn{5}{c}{-3.045 ± 0.128 (Not a meta-learner method)}  \\
    \midrule \\
    \textit{(Snapshot: 6 months)} \\
    CoxPH & -2.910 ± 0.791 & -2.118 ± 0.751 & -2.578 ± 0.009 & -2.639 ± 0.006 & -2.900 ± 0.007 \\
    Random Survival Forest & -2.148 ± 0.957 & -0.925 ± 0.378 & -1.762 ± 0.191 & -1.569 ± 0.135 & -1.807 ± 0.109 \\
    DeepSurv & -3.685 ± 1.998 & -0.516 ± 1.284 & -1.877 ± 0.708 & -1.927 ± 0.594 & -2.152 ± 0.422 \\
    DeepHit & 11.551 ± 3.806 & -1.022 ± 0.428 & 0.007 ± 0.727 & 0.054 ± 0.804 & 0.140 ± 0.876 \\
    \midrule
    Causal Survival Forest &  \multicolumn{5}{c}{-2.831 ± 0.101 (Not a meta-learner method)}  \\
    \bottomrule
    \end{tabular}}\vspace{-1em}}
    % \footnotetext{Not a meta-learner method.}
\end{table*}

In this section, we present our findings on survival analysis prediction performance and on causal inference treatment effect estimates.
Performance metrics are computed at snapshots of \(\tau \in \{3, 6, 9, 12\}\) months.
The main body highlights results at 3 and 6 months, while detailed outcomes for 9 and 12 months are included in Appendix~\ref{apd:survival_analysis} and Appendix~\ref{apd:causal_inference}. 
A comparison of unadjusted survival curves is provided in Appendix~\ref{apd:unadjusted_survival}, demonstrating that the observed Average Treatment Effects (ATEs) cannot be fully attributed to baseline differences between groups.

For transparency and reproducibility\footnote{Code is available in at \url{https://github.com/Shahriarnz14/causal-meta-learner-survival-analysis}}, we provide details on data preprocessing, training protocols, and model hyperparameters in Appendices~\ref{apd:preprocessing} and~\ref{apd:model_hyperparams}.
Additionally, Appendix~\ref{apd:cohort_information} provides a comprehensive breakdown of the study cohort, including demographics, adverse events, adherence patterns, and prescribing trends, offering essential context on the patient population and factors influencing adherence. 

% \vspace{-1em}
\subsection{Results: Adverse Event Time Prediction}
\label{sec:04_results_survival}
% \vspace{-0.25em}

We begin by looking at the prediction performance of the survival models from Section~\ref{sec:methods_survival}.
Tables~\ref{tab:survival_3_6} and~\ref{tab:survival_9_12} report the time-dependent Concordance Index (\(C^{td}\)), Integrated Brier Score (IBS), and time-dependent Area Under the Curve (\(\text{AUC}^{td}\)) for each survival model--Cox proportional hazards (CoxPH), Random Survival Forest (RSF), DeepSurv, and DeepHit--across the four snapshot horizons. 

Across different prediction time snapshots, RSF consistently outperforms other models in terms of \(C^{td}\) and \(\text{AUC}^{td}\) while achieving lower IBS values, suggesting superior calibration and predictive reliability.
CoxPH and DeepSurv exhibit similar performance trends, with CoxPH maintaining slightly higher discriminative ability at later time points, whereas DeepSurv performs better at shorter horizons.

DeepHit underperforms across all time points, with lower \(C^{td}\) and \(\text{AUC}^{td}\) values and higher IBS scores, indicating weaker discrimination and calibration.
At 3 months, its \(C^{td}\) is 0.570 and \(\text{AUC}^{td}\) is 0.588, the lowest among all models, and this trend persists at later horizons, with \(C^{td}\) dropping to 0.540 by 12 months.

The superior performance of RSF suggests its strength in capturing nonlinear relationships and complex interactions among risk scores, demographic features, and adherence patterns.
In contrast, CoxPH is interpretable but limited by its linearity and proportional hazards assumptions, reducing its flexibility in capturing complex dependencies.
DeepSurv and DeepHit exhibit larger variability performance metrics, possibly due to hyperparameter sensitivity.

% \vspace{-1em}
\subsection{Results: ATE Estimation}
\label{sec:04_results_causal}
% \vspace{-0.25em}

Tables~\ref{tab:table_ate_3_6} and~\ref{tab:table_ate_9_12} present the Average Treatment Effect (ATE) estimates for non-adherence at 3, 6, 9, and 12 months\footnote{Appendix~\ref{adp:from_assoc_to_causal} outlines the assumptions required for causal interpretation of our results.}.
Negative ATE values indicate that non-adherence is associated with a shorter adverse event time (i.e., an earlier occurrence of an adverse event) compared to adherence.
Figure~\ref{fig:graphical_ATE_trend} shows the ATE trends for each method across time snapshots.

Several key observations emerge from the ATE estimates. 
At early time points (3, 6, and 9 months), most models indicate that non-adherence is associated with earlier adverse events, with ATE estimates generally ranging from -0.5 to -4 months. 
Matching-based estimators tend to yield the most negative estimates, reinforcing the association between non-adherence and a shorter time to adverse event. 
The Causal Survival Forest (CSF) provides relatively stable estimates, consistently showing a negative ATE.

DeepHit exhibits high variability in its ATE estimates, sometimes producing values that deviate significantly from those of other models. 
Notably, at 6 months, DeepHit reports an ATE of \(11.551 \pm 3.806\), far from the estimates given by other methods. 
This instability aligns with its poor performance in survival analysis, where it consistently showed the lowest \(C^{td}\) and \(\text{AUC}^{td}\) scores and the highest IBS values.

In contrast, Random Survival Forest (RSF) not only achieves strong predictive performance in survival analysis but also provides more stable ATE estimates, consistently reporting negative values across time points, aligning with broader trends observed in other models. 

The negative ATE estimates at early time points align with clinical expectations that sustained medication adherence can stabilize patients with severe mental illness, thereby reducing the risk of crisis events.
Additionally, robust models such as RSF and CoxPH produce relatively consistent ATE estimates across different causal estimation approaches, including T-learners, S-learners, and matching-based estimators.
This consistency further supports the reliability of these estimates.
However, despite adjustments using demographic variables and county-provided risk scores, residual time-varying confounding may still influence these estimates, suggesting a need for more advanced causal modeling approaches that explicitly account for dynamic adherence and confounding effects.

\begin{figure*}[!ht]
\floatconts
  {fig:ite_injectable}
  {\vspace{-2em} \caption{
  Distribution of Estimated Individual Treatment Effects (ITE) for different medication adherence groups: \textcolor{ForestGreen}{injectable}, \textcolor{orange}{non-injectable}, and \textcolor{purple}{not covered} at time snapshot \(\tau=3\) months. 
    Each plot's legend highlights the Average Treatment Effect (ATE) in months for the groups covered by \textcolor{ForestGreen}{injectable} medication, \textcolor{orange}{non-injectable} medication, and those \textcolor{purple}{not covered} by any medication. 
  (a)-(c): T-learner ITEs for CoxPH, DeepSurv, and Random Survival Forest. 
  (d): ITE for Causal Survival Forest.}\vspace{-1em}}
  {%
    \subfigure[\footnotesize (CoxPH, T-Learner)]{\label{fig:ite_injectable_coxph}%
      \includegraphics[width=.22\linewidth]{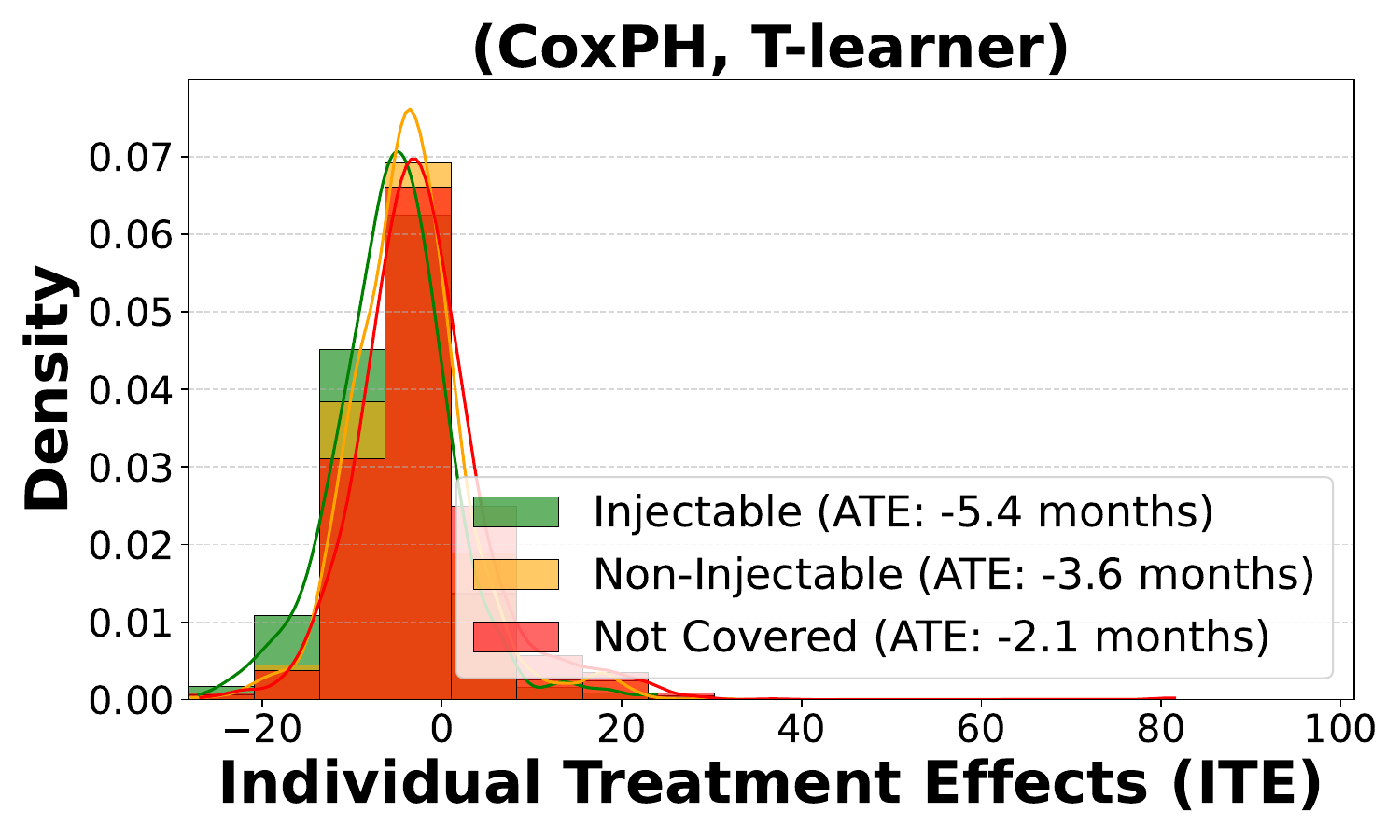}}%
    \qquad
    \subfigure[\footnotesize (DeepSurv, T-Learner)]{\label{fig:ite_injectable_deepsurv}%
      \includegraphics[width=.22\linewidth]{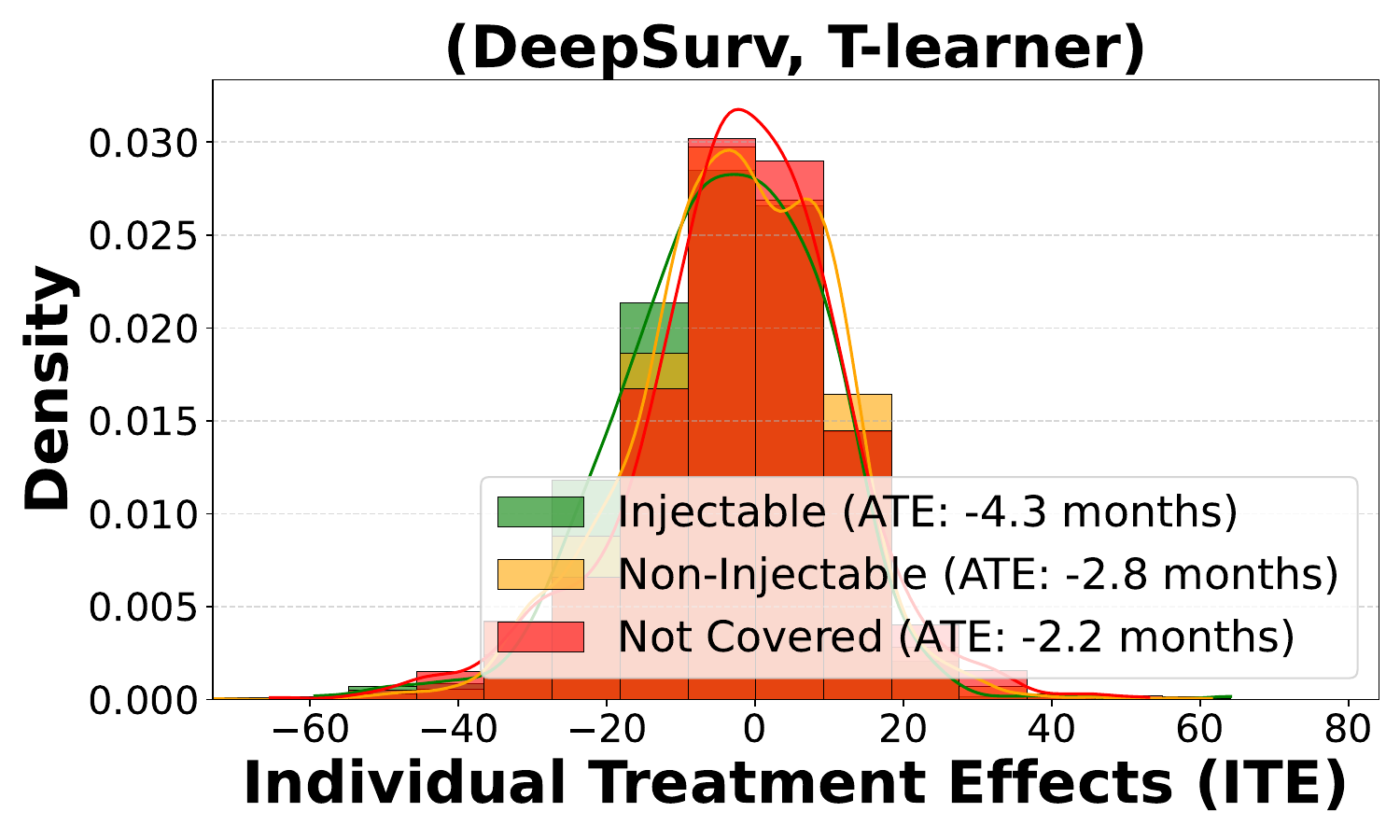}}%
    \qquad
    \subfigure[\footnotesize (RandomSurvivalForest, T-Learner)]{\label{fig:ite_injectable_random_survival_forest}%
      \includegraphics[width=.22\linewidth]{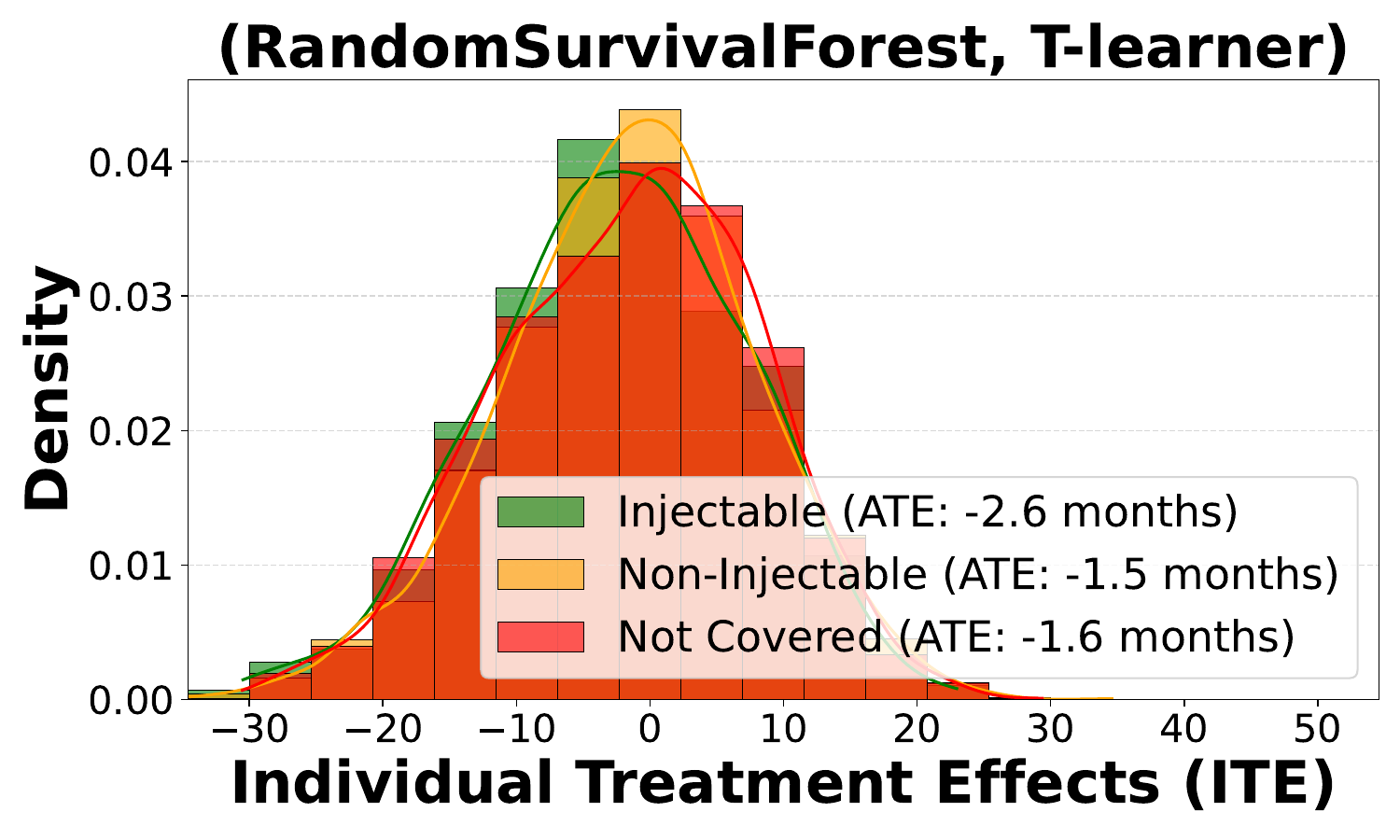}}%
    \qquad
    \subfigure[\footnotesize CausalSurvivalForest]{\label{fig:ite_injectable_causal_survival_forest}%
      \includegraphics[width=.21\linewidth]{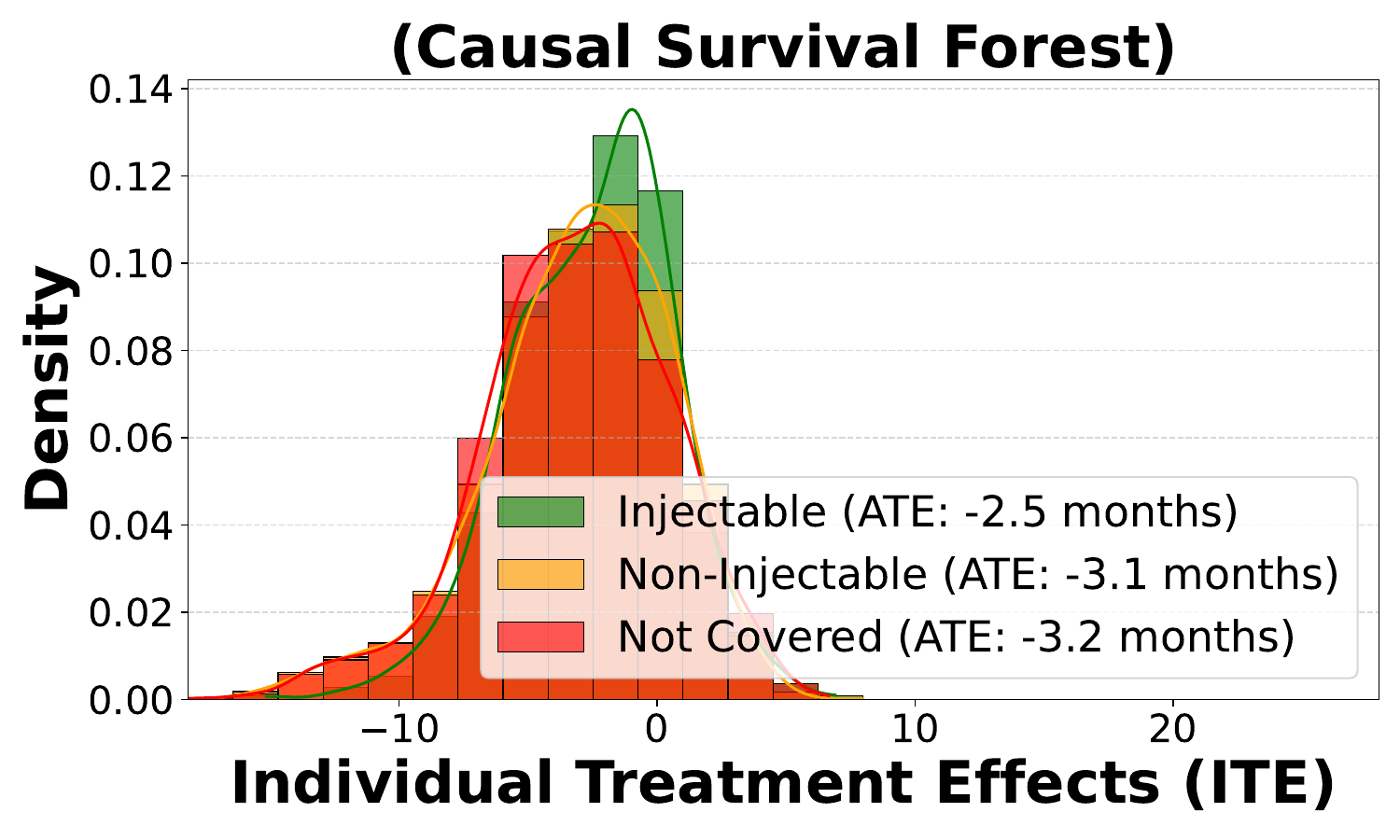}}%
  }\vspace{-1.25em}
\end{figure*}

By 12 months, some ATE estimates, including those from RSF and CSF, shift toward zero or even become positive. 
For instance, CSF reports an ATE of \(0.345 \pm 0.097\), suggesting that adverse event time differences between adherent and non-adherent groups diminish at later snapshots. 
Rather than reflecting a genuine weakening of the treatment effect, this trend likely arises from selection dynamics, as each snapshot only includes individuals who have not had an adverse event up to that point, potentially biasing the estimates toward a ``healthier'' subset. 
Additionally, residual time-varying confounding may contribute to this shift if certain survival-influencing factors are not fully accounted for in the estimators.

% \vspace{-1.5em}
\subsection{Ablation Study: How ATE Estimates Change When Risk Scores are Removed}
\label{sec:04_results_ablation}
% \vspace{-0.25em}

We next examine the impact of removing the county-provided risk scores on the ATE estimates for different survival analysis model and causal inference method pairs. 
\cref{tab:table_ablation_CoxPH,tab:table_ablation_RandomSurvivalForest,tab:table_ablation_DeepSurv,tab:table_ablation_DeepHit,tab:table_ablation_RandomSurvivalForest} in Appendix~\ref{apd:ablation} compare the estimates obtained with the full model versus those from the ablated model (without risk scores) at the snapshot time 3, 6, 9, and 12 months. 
Notably, the magnitude of the negative ATE increases in the ablated models across all time horizons, suggesting that the risk scores partially capture confounding factors. 
Their removal amplifies the estimated impact of non-adherence, suggesting that these scores encode important latent risk factors that, when included, help account for confounding effects.

% \vspace{-1.5em}
\subsection{ITE Distribution for Different Medication}
\label{sec:04_results_ite_analysis}
% \vspace{-.5em}

We conducted a subgroup analysis to compare the distribution of Estimated Individual Treatment Effects (ITE) across three adherence groups: patients covered by injectable medications, non-injectable medications, and those not covered by any medication at the time snapshot \(\tau=3\) months.
Figure~\ref{fig:ite_injectable} presents the ITE distributions for each adherence group using various survival models and meta-learners.
The ITEs reflect the change in Restricted Mean Event Time (RMET) when comparing a non-adherence scenario to adherence within each category.

For the T-learner-based approaches using survival models, the ATEs for injectable medications appear slightly more negative compared to the non-injectable or not-covered groups.
This trend is observed in Figures~\ref{fig:ite_injectable_coxph}, \ref{fig:ite_injectable_deepsurv}, and \ref{fig:ite_injectable_random_survival_forest}, corresponding to CoxPH, DeepSurv, and Random Survival Forest, respectively.
However, the Causal Survival Forest results in Figure~\ref{fig:ite_injectable_causal_survival_forest} do not show a similar distinction, with ATEs for injectables being closer to the non-injectable and not-covered groups.

Overall, the ATEs across all survival models remain close within each formulation group, suggesting no meaningful differences in RMET across injectable, non-injectable, and not-covered groups in our dataset.
The ITE distributions also highlight some variability across models, with Random Survival Forest showing relatively narrower distributions compared to the other methods.
Further subgroup analysis is detailed in Appendix~\ref{apd:ite_injectable_all} for all models and meta-learners and follow the same pattern observed in Figure~\ref{fig:ite_injectable}.

These consistent negative ATE values align with the results reported in Section~\ref{sec:04_results_causal}, indicating shorter adverse event times associated with non-adherence across all adherence categories and medication types.
If nuanced differences existed in ITEs for injectable versus non-injectable formulations, they might have highlighted heterogeneity in treatment effects based on the delivery mechanism of medications.
However, the lack of such variation suggests that the adherence effect may be largely uniform across these groups in the context of our dataset.

In Section~\ref{apd:ite_medication}, we conducted a similar analysis for ITEs by medication type, observing close ATE values across the top four antipsychotic medications --risperidone, aripiprazole, olanzapine, and haloperidol-- with minimal differences in RMETs for non-adherence within each medication, showing no heterogeneity across medication type as well.

These findings indicate no significant heterogeneity in treatment effects across medication formulations or specific antipsychotic medications.
While slight variations in ATE rankings are observed across models, the magnitudes remain similar, suggesting that the adherence effect is largely stable regardless of medication type or delivery mechanism.

% \vspace{-1.5em}
\section{Discussion}
\label{sec:discussion}
% \vspace{-0.5em}
% \paragraph{Overall Findings.}
Our analyses reveal a robust association between medication non-adherence and the earliest occurrence of any of the three adverse events (mortality, involuntary hospitalization, jail booking) in the short- to medium-term (predicting time until the earliest adverse event starting from 3, 6, or 9 months after the first prescription fill recorded).
Our findings demonstrate that, at each time snapshot, medication non-adherence is associated with earlier adverse event time among patients who have not yet experienced any adverse outcome.
Moreover, at these earlier time frames (3, 6, and 9 months), this effect is particularly pronounced, with non-adherence advancing the onset of adverse outcomes by approximately 3 months on average across our models.
Over the first year, these earlier onsets culminate in an estimated 9\% increase in composite adverse events, underscoring the practical significance of adherence in mitigating severe outcomes.

% \vspace{.1em}
% \noindent
% \textbf{Methodological insights: multiple survival models.}~
\vspace{-.25em}
\paragraph{Methodological insights: multiple survival models.}
A central methodological insight from this work is the importance of evaluating multiple survival models in our framework, both for prediction and subsequently in treatment effect estimation.
While Random Survival Forest performed well in our setting, different datasets---especially those with higher dimensional features or more complex event patterns---may favor alternative approaches.
Because survival probabilities and hazard functions form the foundation for counterfactual risk predictions in our framework, careful model selection is indispensable for producing stable Average Treatment Effect (ATE) estimates.

Indeed, our findings show that if a chosen survival model fails to capture the underlying risk dynamics well (as observed with DeepHit in our analysis, where its adverse event prediction performance was noticeably poor), subsequent causal estimates can become unreliable.
Moreover, consistency of ATE estimates across different snapshot times serves as an additional check on a model’s robustness.
In our study, methods with stronger predictive performance (e.g. Random Survival Forest, and Causal Survival Forest) yielded more stable ATE estimates, highlighting a broader analytical insight: reliable causal inferences with right-censored data hinge on trustworthy survival predictions, rendering performance evaluation at multiple time points a critical step in model validation.

% \vspace{.1em}
% \noindent
% \textbf{``Unbundling'' the composite outcome.}~
\vspace{-.25em}
\paragraph{``Unbundling'' the composite outcome.}
In this study, we treat mortality, involuntary hospitalization, and jail booking as a single composite outcome to leverage well-established causal survival analysis methodologies.
An important next step would be disentangling this composite adverse event and actually modeling the three event types to be distinct.
While competing risks models are widely used in the multiple event type setting, our scenario partially involves events that could happen concurrently at least at the time resolution of the data we have access to (e.g., jail booking and involuntary hospitalization can both happen in the same month, where we do not know which actually happens earlier). 
Also, jail booking and involuntary hospitalization could each happen multiple times for an individual, whereas death of course prevents any future adverse event from happening.
Properly modeling how these three adverse events interact in the presence of censoring would necessitate developing an entirely new analytical framework, as current methods in causal survival analysis do not readily support multiple events like the ones we have (without just bundling them into a larger composite adverse event as we have done).

% \vspace{-.5em}
% \noindent
% \textbf{Policy implications and medication type.}~
\vspace{-.25em}
\paragraph{Policy implications and medication type.}
While our subgroup analyses in Section~\ref{sec:04_results_ite_analysis} and Appendix~\ref{apd:ite_medication} revealed consistently negative ATE estimates across multiple medication types, we found minimal evidence of meaningful effect heterogeneity in adherence-related outcomes among the top four antipsychotics.
These findings suggest that interventions aiming to enhance medication adherence may not need to target specific medications in isolation.
Instead, policies may achieve greater impact by focusing on the identification and support of high-risk individuals—irrespective of their prescribed medication.
Nevertheless, certain formulations could be inherently more conducive to adherence (e.g., long-acting injections for patients with specific preferences or support systems), so tailoring strategies to individual circumstances remains a fruitful direction for future research.
Ultimately, interventions improving adherence, whether through better communication, usability enhancements, or policy adjustments, may be broadly beneficial, given the uniformity of adherence effects observed across medication types.

From a clinical standpoint, our findings equip healthcare providers with precise, quantitative tools to clearly communicate the specific risks of medication non-adherence to patients. In our study, given the consistently observed delays in adverse events associated with adherence across different medication types and administration methods, clinicians can leverage this evidence to select medication formulations (injectable versus oral) tailored to individual patient preferences and adherence challenges. This direct and informed communication can significantly enhance patient awareness and engagement by providing concrete, personalized data that clearly illustrate the potential consequences of non-adherence and benefits of adherence-focused interventions. In practical terms, clinicians can use these results to better guide treatment discussions, set realistic patient expectations, and design individualized adherence support plans aimed explicitly at incentivizing and improving medication adherence, thus potentially reducing the incidence and severity of adverse outcomes.

From a policy perspective, our results underscore the effectiveness and efficiency of investing in comprehensive adherence-promoting programs rather than narrower, medication-specific interventions. Policymakers and health administrators can utilize our quantified adherence impacts to prioritize resource allocation towards robust, scalable, and broadly applicable adherence support initiatives. Examples of such initiatives include structured adherence counseling programs and advanced digital adherence technologies. These strategically targeted programs can leverage routinely collected administrative and pharmacy data to identify and proactively engage high-risk individuals, such as reaching out to them to check in on their well-being or possibly more directly incentivizing adherence by providing some sort of financial incentive such as a discount or rebate on specific medications. These targeted problems could significantly mitigate adverse outcomes at the population level and improve overall healthcare quality and efficiency.

% \vspace{.1em}
% \noindent
% \textbf{Actionable interventions informed by adherence quantification.}~
% \vspace{-.25em}
\paragraph{Actionable interventions informed by adherence quantification.}
Our findings underscore the need for proactive, targeted interventions to enhance adherence. Effective strategies could include personalized outreach programs, such as scheduled follow-up calls, digital medication reminders, or telehealth sessions designed around patient preferences and needs. Transitioning suitable patients to long-acting injectable formulations may also mitigate adherence challenges by simplifying medication routines. Additionally, policy adjustments aimed at streamlining prescription renewals could reduce practical barriers to sustained adherence. Previous studies have consistently demonstrated that structured adherence interventions significantly lower hospitalization rates and associated healthcare costs \citep{velligan2009expert, kane2016comprehensive}. Thus, the quantified impact of adherence presented here can directly guide clinicians and policymakers in designing and implementing interventions optimized for maximal clinical benefit and resource efficiency.

% \vspace{.1em}
% \noindent
% \textbf{Future research directions.}~
\vspace{-.25em}
\paragraph{Other future research directions.}
Although our observational design cannot definitively establish causality, the consistently negative ATE estimates indicate a strong link between non-adherence and earlier adverse events.
Nonetheless, unmeasured factors correlated with adherence may drive this relationship, so these findings do not guarantee that improving adherence will necessarily prevent adverse outcomes.
Given the practical challenges of directly randomizing adherence, a promising avenue is to design RCTs focused on communication-based interventions rather than attempting to force medication use, which would raise ethical concerns.
Such interventions could test whether tailored outreach effectively increase adherence and reduce adverse events.
Future studies should also collect additional clinical and behavioral data, incorporate time-varying confounders, and investigate whether specific subgroups (e.g., by comorbidities or socioeconomic status) are more vulnerable.

While our investigation focuses on individuals with schizophrenia, the general principles extend to other severe mental illnesses that demand sustained medication adherence.
Ultimately, ensuring adherence could delay the onset of life-threatening or socially destabilizing outcomes, reinforcing the importance of ongoing research and intervention development in this critical public health arena.
Although our study is based on data from a single county, the methodological framework we present is broadly applicable. Future studies could replicate this analysis using data from multiple regions or healthcare systems to assess the consistency of the observed associations across diverse settings. Additionally, incorporating data from different demographic groups or healthcare infrastructures could provide insights into how contextual factors influence the relationship between medication non-adherence and adverse outcomes.

% \vspace{-1.5em}
\newpage
\section*{Acknowledgments}
\label{sec:acknowledgments}
% \vspace{-0.5em}
This research was supported in part by the Division of Intramural Research (DIR) of the National Library of Medicine (NLM), National Institutes of Health (NIH).
G. H. C. was supported by NSF CAREER award \#2047981.
S. N. was supported by Carnegie Mellon University TCS Presidential Fellowship, and Natural Sciences and Engineering Research Council of Canada (NSERC) PGS-D award.
S.N was also supported in part by an appointment to the National Library of Medicine Research Participation Program administered by the Oak Ridge Institute for Science and Education (ORISE) through an interagency agreement between the U.S. Department of Energy (DOE) and the National Library of Medicine, National Institutes of Health. 
ORISE is managed by ORAU under DOE contract number DE-SC0014664. 
All opinions expressed in this paper are the authors' and do not necessarily reflect the policies and views of NIH, NLM,
DOE, or ORAU/ORISE.

The authors would also like to thank Mr. Ethan Goode, Dr. Brian Kovak, Dr. Eli Ben-Michael, Dr. Akshaya Jha, Dr. Edward Kennedy, and Dr. Zachary Lipton for helpful discussions.

% \clearpage
\bibliography{references}

\clearpage

\appendix
\section*{Appendix}
\label{apd:outline}
The appendix is organized into five main sections.
Appendix~\ref{apd:causal_derivations} presents causal derivations and assumptions underlying our analysis. Appendix~\ref{apd:full_results} reports the full experimental results across different model snapshots and causal estimation methods. Appendix~\ref{apd:medication_subgroup_analysis} provides subgroup analyses by medication. Appendix~\ref{apd:dataset_information} describes the study cohort characteristics. Lastly, Appendix~\ref{apd:preprocessing_and_modeling_hyperparameters} details data preprocessing steps and modeling hyperparameters.

\noindent\emph{Each section is further organized as follows:}

In Appendix~\ref{apd:ate-ite}, we derive the relationship between the Average Treatment Effect (ATE) and the Individual Treatment Effect (ITE) under the potential outcomes framework for restricted mean adverse event time.
In Appendix~\ref{adp:from_assoc_to_causal}, we discuss the assumptions required for causal interpretation.
We outline the potential outcomes framework assumptions and the survival analysis assumptions necessary for making causal claims in our setup.
We also briefly highlight challenges with time-varying confounding and suggest directions for improvement.  

Appendix~\ref{apd:causal-meta-learner-algorithms} provides the pseudo-code for the adaptation of causal meta-learners (T-Learner, S-Learner, and Nearest Neighbor Matching Method) to survival analysis.
In Appendix~\ref{apd:matching_derivation} and \ref{apd:csf_ate_derivation}, we illustrate how the treatment effect estimator is derived for the matching meta-learner and Causal Survival Forest respectively. 
Finally, in Appendix~\ref{apd:in_context_of_new_research}, we highlight emerging methodologies that could complement or extend our approach and position our study within a larger research trajectory 

In Appendix~\ref{apd:survival_analysis}, we present the performance of survival analysis models at 9- and 12-month snapshots.  
We summarize model discriminative ability and calibration, highlighting the superiority of the Random Survival Forest.  
In Appendix~\ref{apd:causal_inference}, we report the full causal inference results for 9- and 12-month snapshots.  
We analyze shifts in Average Treatment Effect (ATE) estimates and discuss the role of survival model performance and selection bias. 
The graphical representation of ATE estimates as snapshot time of the prediction is progresses is presented in Appendix~\ref{apd:graphical_ate}.

In Appendix~\ref{apd:unadjusted_survival}, we present Kaplan-Meier survival curves for adherent and non-adherent patients.  
We report restricted mean event time (RMET) estimates as a baseline for later analyses which puts into context that the treatment effects found is not due raw difference of the two cohorts.

In Appendix~\ref{apd:ablation}, we evaluate the effect of removing county-provided risk scores on ATE estimates.  
We analyze how these scores act as proxies for unmeasured confounders across different survival models.  

In Appendix~\ref{apd:ite_medication}, we report Individual Treatment Effect (ITE) distributions for different medications and in Appendix~\ref{apd:ite_injectable_all}, we provide the ITE estimates by medication administration type for all survival model and meta-learner pairs. 

In Appendix~\ref{apd:cohort_information}, we first present a visual overview of the modeling setup and cohort structure (Figure~\ref{fig:data_overview}), illustrating how the first adverse event is defined relative to the prediction snapshot and how monthly adherence is binarized. 
We then describe the study cohort, including demographics, adherence patterns, and prescribing trends.  
We detail demographics in Appendix~\ref{apd:cohort_demographics}, first adverse events timing and types in Appendix~\ref{apd:cohort_first_event},  
non-adherence patterns in Appendix~\ref{apd:cohort_nonadherence}, and antipsychotic medication trends in Appendix~\ref{adp:cohort_medications}.  
In Appendix~\ref{apd:risk_score_evaluations}, we assess the predictive accuracy of county-provided risk scores.  
We present ROC curves and AUC values for predicting specific adverse events within 12 months. 

In Appendix~\ref{apd:preprocessing}, we outline data preprocessing steps.  
We describe cohort filtering, covariate construction, feature normalization, train-test splitting, and trimming strategy for the positivity assumption of our causal analysis.  
In Appendix~\ref{apd:model_hyperparams}, we summarize hyperparameter settings for all survival models.  

\clearpage
\section{Causal Derivations and Assumptions}
\label{apd:causal_derivations}
\numberwithin{equation}{section}
\numberwithin{figure}{section}
\numberwithin{table}{section}

\subsection{Derivation Relating ATE and ITE}
\label{apd:ate-ite}

In the potential outcomes framework, under consistency, randomization (no unmeasured confounders), and positivity assumption, we have
\[
\mathbb{E}[\bar{T_i}|A=a, \mathbf{Z}_{i\tau}]=\mathbb{E}[\bar{T_i}(a) \mid \mathbf{Z}_{i\tau}].
\]
Hence, we can write our setup as:
\begin{align*}
\psi
    &= \mathbb{E}[\bar{T_i}(1) - \bar{T_i}(0)] \\
    &= \mathbb{E}\big[\mathbb{E}[\bar{T_i}(1) - \bar{T_i}(0) \mid A_{i\tau} = 1]\\
    &\quad \quad- \mathbb{E}[\mathbb{E}[\bar{T_i}(1) - \bar{T_i}(0) \mid A_{i\tau} = 0 ] ~\big|~ \mathbf{Z}_{i\tau}\big] \\
    &=\mathbb{E}[\mu_1(\mathbf{Z}_{i\tau})-\mu_0(\mathbf{Z}_{i\tau})] \\
    &=\mathbb{E}[\text{ITE}_i].
\end{align*}

\subsection{Assumptions for Causal Interpretation}
\label{adp:from_assoc_to_causal}
Although our study leverages a comprehensive set of covariates and sophisticated modeling strategies, drawing causal conclusions in an observational setting requires additional assumptions beyond those used for association-based inferences. 
If these assumptions hold, our estimates can be interpreted as causal effects rather than associations; however, we acknowledge that some assumptions may not fully hold in cohort of study. 
To mitigate potential violations --particularly in the potential outcomes framework-- we employ trimming, as outlined in \ref{apd:preprocessing}. 
Trimming excludes patients with a very low probability of being either adherent or non-adherent, improving the plausibility of the positivity assumption and reducing extrapolation biases in estimating causal effects.

In this section, we detail the assumptions necessary for a causal interpretation, including those related to the potential outcomes framework, survival analysis, and longitudinal data.

\paragraph{Potential outcomes framework.} 
Under the potential outcomes framework, we invoke the following core assumptions:
\textit{(i)} \emph{Consistency}, which implies that for any individual, the potential outcome under the observed treatment exposure coincides with the observed outcome;
\textit{(ii)} \emph{Positivity}, meaning that each patient has a nonzero probability of being either adherent or non-adherent across relevant strata of covariates;
\textit{(iii)} \emph{Ignorability} (or no unmeasured confounding), which holds that all confounders of the relationship between adherence and outcomes are measured and properly accounted for in the model; and 
\textit{(iv)} \emph{Stable Unit Treatment Value Assumption (SUTVA)}, which requires that one patient’s potential outcomes are unaffected by other patients’ treatments. 
In principle, if these assumptions were satisfied and our models were correctly specified, the estimates presented would represent unbiased estimators of causal effects, rather than mere associations.

\paragraph{Survival analysis assumptions.}
To make causal claims from our underlying survival models, we need to make additional assumptions.
First, we assume \emph{non-informative censoring}, which states that the probability of being censored is independent of the adverse event, conditional on observed covariates.
If censoring is informative, then our standard survival estimators may be biased.
Second, we require \emph{consistency} of the underlying survival models,  ensuring that, given a sufficiently large sample, the estimated survival function converges to the true survival function.
Finally, we assume \emph{independence of survival times}, meaning that the survival times of different individuals are independent.
These assumptions are necessary to ensure that our underlying survival models yield valid causal estimates rather than biased associations driven by censoring mechanisms or dependent survival times in our causal framework.

\paragraph{Longitudinal data assumptions.}
Lastly, when dealing with longitudinal data, causal interpretations become more nuanced when treatment and covariates vary over time.
In principle, addressing \emph{time-varying confounding} requires specialized methods (e.g., marginal structural models or joint modeling approaches) to ensure that changing adherence patterns are not themselves driven by evolving unmeasured factors.
To approximate these conditions in our setup, we leveraged extensive administrative data and county-provided risk scores, but the possibility of residual confounding remains.
Future work could further mitigate bias by refining the measurement of dynamic exposures, extending models to incorporate time-varying covariates in greater detail, and conducting sensitivity analyses to quantify the impact of potential violations of these assumptions.

\subsection{Causal Meta-Learner Algorithms}
\label{apd:causal-meta-learner-algorithms}

\begin{algorithm}[ht]
\caption{T-Learner for Restricted Mean Event Time}
\label{alg:t-learner}
\KwIn{Cohort $\mathcal{D}_\tau = \{(\mathbf{Z}_{i\tau}, A_{i\tau}, T_i, \delta_i)\}_{i=1}^N$, time horizon $M$}
\KwOut{Estimate $\widehat{\psi}_{\mathrm{T\text{-}learner}}$}

Partition into treated and control groups:\;
\[
\mathcal{D}_a \leftarrow \{(\mathbf{Z}_{i\tau}, T_i, \delta_i) : A_{i\tau} = a\}, \quad a \in \{0,1\}
\]

\For{$a \in \{0,1\}$}{
    Fit a survival model on $\mathcal{D}_a$ to obtain $\widehat{S}_a(u \mid \mathbf{Z}_{i\tau})$\;
    \For{$i = 1$ \KwTo $N$}{
        \[
        \widehat{\mu}_a(\mathbf{Z}_{i\tau}) \gets \int_0^M \widehat{S}_a(u \mid \mathbf{Z}_{i\tau})\, du
        \]
    }
}
\Return{
\[
\widehat{\psi}_{\mathrm{T\text{-}learner}} = \frac{1}{N} \sum_{i=1}^N \Bigl(\widehat{\mu}_1(\mathbf{Z}_{i\tau}) - \widehat{\mu}_0(\mathbf{Z}_{i\tau})\Bigr)
\]
}
\end{algorithm}

We provide detailed pseudocode for how we adapted the causal meta-learner algorithms to survival analysis to be used to estimate the treatment effect on restricted mean event time (RMET). 
Specifically, we implement T-Learner, S-Learner, and \(K\)-Nearest-Neighbors Matching approaches adapted for survival analysis under a finite time horizon. 
Each algorithm estimates individual and average treatment effects by modeling survival functions conditioned on baseline covariates and treatment assignments. 

Algorithm~\ref{alg:t-learner} describes the T-Learner, which fits separate survival models for treated and control groups. Algorithm~\ref{alg:s-learner} details the S-Learner, which fits a single survival model incorporating the treatment as an additional covariate. 
Algorithm~\ref{alg:matching} outlines the \(K\)-Nearest-Neighbors Matching Estimator, which imputes counterfactual outcomes by averaging over nearest neighbors from the opposite treatment group in the covariate space.

All methods operate on the cohort \(\mathcal{D}_\tau\), which contains baseline covariates \(\mathbf{Z}_{i\tau}\), treatment assignments \(A_{i\tau}\), observed event times \(T_i\), and event/censoring indicators \(\delta_i\). 
Each approach uses a time horizon \(M\) to compute restricted mean estimates of survival time and treatment effects.

\begin{algorithm}[t]
\caption{S-Learner for Restricted Mean Event Time}
\label{alg:s-learner}
\KwIn{Cohort $\mathcal{D}_\tau = \{(\mathbf{Z}_{i\tau}, A_{i\tau}, T_i, \delta_i)\}_{i=1}^N$, time horizon $M$}
\KwOut{Estimate $\widehat{\psi}_{\mathrm{S\text{-}learner}}$}

Fit a single survival model on all data:\;
\[
\widehat{S}(u \mid \mathbf{Z}_{i\tau}, A_{i\tau})
\]

\For{$i = 1$ \KwTo $N$}{
    \For{$a \in \{0,1\}$}{
        \[
        \widehat{\mu}(\mathbf{Z}_{i\tau}, a) \gets \int_0^M \widehat{S}(u \mid \mathbf{Z}_{i\tau}, a)\, du
        \]
    }
}

\Return{
\[
\widehat{\psi}_{\mathrm{S\text{-}learner}} = \frac{1}{N} \sum_{i=1}^N \Bigl(\widehat{\mu}(\mathbf{Z}_{i\tau},1) - \widehat{\mu}(\mathbf{Z}_{i\tau},0)\Bigr)
\]
}
\end{algorithm}

\begin{algorithm}[t]
\caption{\(K\)-Nearest-Neighbors Matching Estimator}
\label{alg:matching}
\KwIn{Cohort $\mathcal{D}_\tau = \{(\mathbf{Z}_{i\tau}, A_{i\tau}, T_i, \delta_i)\}_{i=1}^N$, time horizon $M$, neighbors $K$}
\KwOut{Estimate $\widehat{\psi}_{\mathrm{match}}$}

Fit a joint survival model to compute factual RMETs:\;
\[
\widehat{S}(u \mid \mathbf{Z}, A)
\]

\For{$i = 1$ \KwTo $N$}{
    \[
    \widehat{\mu}_{A_{i\tau}}(\mathbf{Z}_{i\tau}) \gets \int_0^M \widehat{S}(u \mid \mathbf{Z}_{i\tau}, A_{i\tau})\, du
    \]
    
    Find opposite-group neighbors:\;
    \begin{align*}
    J_K(i) \gets \text{K nearest in covariate space among } \\ 
    \{ j : A_{j\tau} = 1 - A_{i\tau} \}
    \end{align*}
    
    Estimate counterfactual RMET:\;
    \[
    \widehat{\mu}_{1-A_{i\tau}}(\mathbf{Z}_{i\tau}) \gets \frac{1}{K} \sum_{j \in J_K(i)} \widehat{\mu}_{A_{j\tau}}(\mathbf{Z}_{j\tau})
    \]
    
    Compute individual treatment effect:\;
    \[
    \widehat{\mathrm{ITE}}_i \gets \widehat{\mu}_1(\mathbf{Z}_{i\tau}) - \widehat{\mu}_0(\mathbf{Z}_{i\tau})
    \]
}

\Return{
\[
\widehat{\psi}_{\mathrm{match}} = \frac{1}{N} \sum_{i=1}^N \widehat{\mathrm{ITE}}_i
\]
}
\end{algorithm}

\subsection{Matching Estimator Derivation}
\label{apd:matching_derivation}
The matching estimator aims to estimate the Individual Treatment Effect (ITE) for each patient $ i $, defined as the difference in RMET under treatment (non-adherence, $ A_{i\tau} = 1 $) versus control (adherence, $ A_{i\tau} = 0 $):
$$ \text{ITE}_i = \bar{T}_i(1) - \bar{T}_i(0), $$
where $ \bar{T}_i(a) = \mu_a(\mathbf{Z}_{i\tau}) = \int_{0}^{M} S(u \mid A_{i\tau} = a, \mathbf{Z}_{i\tau}) \, du $ represents the true RMET under treatment $ a $, and $ \mathbf{Z}_{i\tau} $ is the covariate vector at time $ \tau $. The upper bound $ M $ is a fixed time horizon (e.g., the maximum follow-up time).

In practice, we observe only one of these potential outcomes for each patient:
\begin{itemize}
    \item If $ A_{i\tau} = 1 $, we observe $ \bar{T}_i(1) $, but $ \bar{T}_i(0) $ is counterfactual.
    \item If $ A_{i\tau} = 0 $, we observe $ \bar{T}_i(0) $, but $ \bar{T}_i(1) $ is counterfactual.
\end{itemize}

The matching estimator uses a survival model and nearest-neighbor matching to estimate both the factual and counterfactual RMETs.

Steps of the Matching Estimator:
\begin{enumerate}
\item Estimate the Factual RMET:
   We fit a survival model (e.g., Random Survival Forest or Cox Proportional Hazards) to estimate the survival function $ \widehat{S}(u \mid \mathbf{Z}_{i\tau}, A_{i\tau}) $ for the observed treatment $ A_{i\tau} $. The factual RMET is:
   $$ \widehat{\mu}_{A_{i\tau}}(\mathbf{Z}_{i\tau}) = \int_{0}^{M} \widehat{S}(u \mid \mathbf{Z}_{i\tau}, A_{i\tau}) \, du. $$
   This represents the estimated RMET under the treatment the patient actually received.\\
(Note that this is similar to S-learner’s $\widehat{\mu}(\mathbf{Z}_{i\tau}, a)$ with $a=A_{i\tau}$ the observed treatment).

\item Estimate the Counterfactual RMET:
   For each patient $ i $, we identify the $ K $ nearest neighbors (based on $ \mathbf{Z}_{i\tau} $) from the opposite treatment group. The counterfactual RMET is approximated by averaging the factual RMETs of these neighbors:
   \begin{itemize}
   \item If $ A_{i\tau} = 1 $ (non-adherent), the counterfactual RMET under adherence ($ a = 0 $) is:
     $$ \widehat{\mu}_{0}(\mathbf{Z}_{i\tau}) = \frac{1}{K} \sum_{j \in J_K(i)} \widehat{\mu}_{A_{j\tau}}(\mathbf{Z}_{j\tau}),$$
     where $ A_{j\tau} = 0, $
     and $ J_K(i) $ is the set of $ K $ nearest neighbors with $ A_{j\tau} = 0 $.
   \item If $ A_{i\tau} = 0 $ (adherent), the counterfactual RMET under non-adherence ($ a = 1 $) is:
     $$ \widehat{\mu}_{1}(\mathbf{Z}_{i\tau}) = \frac{1}{K} \sum_{j \in J_K(i)} \widehat{\mu}_{A_{j\tau}}(\mathbf{Z}_{j\tau}),$$ where $A_{j\tau} = 1. $
    \end{itemize}
\item Estimate the ITE:
   The ITE for each patient is the difference between the estimated RMETs under treatment and control:
   \begin{itemize}
   \item For $ A_{i\tau} = 1 $:
     $ \widehat{\text{ITE}}_i = \widehat{\mu}_{1}(\mathbf{Z}_{i\tau}) - \widehat{\mu}_{0}(\mathbf{Z}_{i\tau}) = \widehat{\mu}_{A_{i\tau}}(\mathbf{Z}_{i\tau}) - \frac{1}{K} \sum_{j \in J_K(i)} \widehat{\mu}_{1-A_{j\tau}}(\mathbf{Z}_{j\tau}). $
   \item For $ A_{i\tau} = 0 $:
     $ \widehat{\text{ITE}}_i = \widehat{\mu}_{1}(\mathbf{Z}_{i\tau}) - \widehat{\mu}_{0}(\mathbf{Z}_{i\tau}) = \frac{1}{K} \sum_{j \in J_K(i)} \widehat{\mu}_{1-A_{j\tau}}(\mathbf{Z}_{j\tau}) - \widehat{\mu}_{A_{i\tau}}(\mathbf{Z}_{i\tau}). $
    \end{itemize}
\item Estimate the ATE:
   The ATE is the average of the ITEs across all $ N $ patients:
   $$ \widehat{\psi}_{\text{match}} = \frac{1}{N} \sum_{i=1}^{N} \widehat{\text{ITE}}_i. $$
   This can be written compactly using the indicator $ (2 A_{i\tau} - 1) $ to adjust the subtraction direction:
   {\footnotesize
   $$ \widehat{\psi}_{\text{match}} = \frac{1}{N} \sum_{i=1}^{N} \left( \widehat{\mu}_{A_{i\tau}}(\mathbf{Z}_{i\tau}) - \widehat{\mu}_{1 - A_{i\tau}}(\mathbf{Z}_{i\tau}) \right) (2 A_{i\tau} - 1). $$}
   \begin{itemize}
   \item When $ A_{i\tau} = 1 $, $ (2 A_{i\tau} - 1) = 1 $, so the term inside the sum is $$ \widehat{\mu}_{1}(\mathbf{Z}_{i\tau}) - \widehat{\mu}_{0}(\mathbf{Z}_{i\tau}). $$
   \item When $ A_{i\tau} = 0 $, $ (2 A_{i\tau} - 1) = -1 $, so the term inside the sum becomes 
   \begin{align*} -(\widehat{\mu}_{0}(\mathbf{Z}_{i\tau}) - \widehat{\mu}_{1}(\mathbf{Z}_{i\tau})) \\ = \widehat{\mu}_{1}(\mathbf{Z}_{i\tau}) - \widehat{\mu}_{0}(\mathbf{Z}_{i\tau}).
   \end{align*}
    \end{itemize}
\end{enumerate}

Role of $ \widehat{\mu} $ and $ \widehat{S}(\cdot) $:
\begin{itemize}
\item $ \widehat{\mu}_{A_{i\tau}}(\mathbf{Z}_{i\tau}) $: The factual RMET, computed directly from the survival function $ \widehat{S}(u \mid \mathbf{Z}_{i\tau}, A_{i\tau}) $ via integration.
\item $ \widehat{\mu}_{1 - A_{i\tau}}(\mathbf{Z}_{i\tau}) $: The counterfactual RMET, approximated by averaging the $ \widehat{\mu}_{A_{j\tau}}(\mathbf{Z}_{j\tau}) $ of $ K $ nearest neighbors from the opposite treatment group, where each $ \widehat{\mu}_{A_{j\tau}}(\mathbf{Z}_{j\tau}) $ is also derived from $ \widehat{S}(\cdot) $.
\item $ \widehat{S}(\cdot) $: The survival function is foundational, as it is used to compute all $ \widehat{\mu} $ values, linking survival analysis to causal estimation.
\end{itemize}

\subsection{ATE Estimate Derivation of Causal Survival Forest}
\label{apd:csf_ate_derivation}
The Causal Survival Forest (CSF), as introduced by \citet{cui2023estimating}, is an inherently causal method designed to estimate heterogeneous treatment effects in survival data. Unlike meta-learners that adapt existing survival models (e.g., matching), CSF directly estimates the treatment effect within a random forest framework tailored to survival outcomes. We used CSF as another baseline to validate our treatment effect findings.

CSF builds an ensemble of survival trees. Each tree splits the covariate space $ \mathbf{Z}_{i\tau} $ using a criterion that maximizes the difference in survival outcomes between treatment groups.
\begin{itemize}
\item For a patient $ i $, the estimate $ \widehat{\mu}_a(\mathbf{Z}_{i\tau}) $ is computed by averaging the RMSTs from the leaf nodes across all trees where $ \mathbf{Z}_{i\tau} $ falls under treatment $ a $.
\item Internally, CSF estimates the survival function $ \widehat{S}(u \mid \mathbf{Z}_{i\tau}, a) $ for each treatment level and computes:
  $ \widehat{\mu}_a(\mathbf{Z}_{i\tau}) = \int_{0}^{M} \widehat{S}(u \mid \mathbf{Z}_{i\tau}, a) \, du. $
\item CSF estimates the ITE for each patient $ i $ as the difference in Restricted Mean Survival Time (RMST, equivalent to RMET in our context) between treatment arms:
  $ \widehat{\theta}_{i}(\mathbf{Z}_{i\tau}) = \widehat{\mu}_1(\mathbf{Z}_{i\tau}) - \widehat{\mu}_0(\mathbf{Z}_{i\tau}), $
  where $ \widehat{\mu}_a(\mathbf{Z}_{i\tau}) $ is the estimated RMST under treatment $ a \in \{0, 1\} $.
\item The ATE is the average of these ITEs:
  \[ \widehat{\psi}_{\text{CSF}} = \frac{1}{N} \sum_{i=1}^{N} \widehat{\theta}_{i}(\mathbf{Z}_{i\tau}). \]
\item Unlike the matching estimator, CSF does not rely on external survival models or post-hoc aggregation; it optimizes directly for treatment effect estimation within the forest structure.
\end{itemize}

Role of $ \widehat{\mu} $ and $ \widehat{S}(\cdot) $:
\begin{itemize}
\item $ \widehat{\mu}_a(\mathbf{Z}_{i\tau}) $: The RMST for treatment $ a $, estimated directly by CSF using tree-based survival predictions.
\item $ \widehat{S}(\cdot) $: The survival function is estimated internally within CSF and integrated to produce $ \widehat{\mu}_a(\mathbf{Z}_{i\tau}) $, enabling the computation of treatment effects.
\end{itemize}

\subsection{Emerging Methodologies and Extensions}
\label{apd:in_context_of_new_research}
Beyond the standard meta-learner and matching approaches utilized here, newer developments in survival analysis and causal inference offer promising directions for methodological refinement.
For instance, techniques such as double machine learning for causal inference on survival functions may provide enhanced flexibility and reduced bias under certain assumptions \citep{kabata4875226double}.
Recent work on handling covariate-dependent censoring and left truncation similarly represents an exciting area of inquiry \citep{wang2024learning}.
Although these methods as of writing of this paper are still not published, integrating them into analyses like ours could further strengthen causal inferences and accommodate complex real-world data nuances.
\newpage

\clearpage
\section{Full Experimental Results}
\label{apd:full_results}
\numberwithin{equation}{section}
\numberwithin{figure}{section}
\numberwithin{table}{section}

\subsection{Survival Analysis Model Performance across 9 and 12 months Snapshots}
\label{apd:survival_analysis}

The prediction performance of the survival analysis models across 9 and 12 months snapshots is summarized in Table~\ref{tab:survival_9_12}.
Similar to earlier time points presented in Table~\ref{tab:survival_3_6}, the Random Survival Forest consistently outperforms other models in terms of discriminative ability and calibration.
At 9 months, it achieves the highest Concordance Index (\(C^{td} = 0.648 \pm 0.015\)) and time-dependent AUC (\(\text{AUC}^{td} = 0.694 \pm 0.015\)), along with the lowest Integrated Brier Score (\(\text{IBS} = 0.180 \pm 0.003\)).
At 12 months, the Random Survival Forest maintains its superior performance with \(C^{td} = 0.658 \pm 0.011\), \(\text{AUC}^{td} = 0.706 \pm 0.011\), and \(\text{IBS} = 0.174 \pm 0.004\).

CoxPH and DeepSurv display competitive performance at 9 months, with CoxPH maintaining slightly higher \(C^{td}\) scores (\(0.625 \pm 0.023\)) and comparable calibration as indicated by its \(\text{IBS} = 0.187 \pm 0.004\).
DeepSurv exhibits strong calibration at 9 months (\(\text{IBS} = 0.185 \pm 0.004\)), though it trails behind Random Survival Forest in overall predictive metrics.
By 12 months, DeepSurv demonstrates modest gains, but CoxPH results are unavailable for this time point.

DeepHit continues to underperform across these later snapshots, with \(C^{td}\) values of \(0.525 \pm 0.069\) at 9 months and \(0.540 \pm 0.026\) at 12 months, indicating weaker discriminative ability.
Similarly, its higher \(\text{IBS}\) values (\(0.234 \pm 0.026\) at 9 months and \(0.243 \pm 0.021\) at 12 months) suggest poorer calibration.
These trends highlight DeepHit's limitations in handling the low-dimensional and moderate-sized dataset used in this study.
Given these limitations, the use of DeepHit in our causal inference meta-learner approach for predicting restricted mean time (RMET) is unreliable, as its survival estimates lack the necessary stability and accuracy.

\begin{table*}[!htbp]
    \setlength{\tabcolsep}{2pt}
    \floatconts
      {tab:survival_9_12}
      {\caption{Prediction Performance Across Time Snapshots (9 and 12 months)}\vspace{-.75em}}
      {\adjustbox{max width=.9\textwidth}{\begin{tabular}{l  c c c @{\hspace{0.5cm}} c c c }
    \toprule
    \textbf{Snapshot Time}&\multicolumn{3}{c}{9 months} & \multicolumn{3}{c}{12 months} \\
    \midrule
     & \(C^{td}\) & IBS & \(\text{AUC}^{td}\) & \(C^{td}\) & IBS & \(\text{AUC}^{td}\) \\
    \midrule
    CoxPH & 0.625 ± 0.023 & 0.187 ± 0.004 & 0.664 ± 0.025 & 0.630 ± 0.024 & 0.195 ± 0.010 & 0.672 ± 0.023 \\
    Random Survival Forest & {\bftab 0.648} ± 0.015 & {\bftab 0.180} ± 0.003 & {\bftab 0.694} ± 0.015 & {\bftab 0.658} ± 0.011 & {\bftab 0.174} ± 0.004 & {\bftab 0.706} ± 0.011 \\
    DeepSurv & 0.630 ± 0.009 & 0.185 ± 0.004 & 0.668 ± 0.011 & 0.636 ± 0.013 & 0.187 ± 0.010 & 0.675 ± 0.016 \\
    DeepHit & 0.525 ± 0.069 & 0.234 ± 0.026 & 0.556 ± 0.065 & 0.540 ± 0.026 & 0.243 ± 0.021 & 0.558 ± 0.014 \\
    \bottomrule
    \end{tabular}}}
\end{table*}

Overall, the results reinforce the performance advantage of the Random Survival Forest across varying time horizons, making it the most reliable model evaluated for predicting time-to-first-event in our dataset.
However, even with \(C^{td}\) values consistently above 0.6, there remains significant room for improvement in survival modeling.
This gap in predictive accuracy introduces a caveat for our causal inference analyses, as the reliability of causal estimates is contingent on the underlying survival model's performance.
Addressing this limitation is critical for advancing both predictive and causal modeling in similar clinical contexts.

\subsection{Full Causal Inference Results across 9 and 12 months Snapshots}
\label{apd:causal_inference}

\begin{table*}[!hbtp]
    \setlength{\tabcolsep}{4pt}
    % \footnotesize
    \floatconts
      {tab:table_ate_9_12}
      {\caption{Causal Inference ATE Estimates at 9 and 12 Months}\vspace{-.5em}}
      {\adjustbox{max width=.825\textwidth}{\begin{tabular}{l c c c c c }
    \toprule
    &T-learner & S-learner & Matching (1) & Matching (5) & Matching (20) \\
    \midrule
    \textit{(Snapshot: 9 months)} \\
    CoxPH & -2.285 ± 0.578 & -1.384 ± 0.512 & -1.658 ± 0.006 & -1.790 ± 0.002 & -2.032 ± 0.003 \\
    Random Survival Forest & -1.468 ± 0.518 & -0.546 ± 0.106 & -1.159 ± 0.042 & -1.242 ± 0.050 & -1.540 ± 0.055 \\
    DeepSurv & -2.144 ± 2.706 & 0.837 ± 0.767 & -0.923 ± 0.446 & -0.928 ± 0.436 & -1.021 ± 0.558 \\
    DeepHit & 6.887 ± 5.930 & 0.169 ± 0.692 & -0.160 ± 0.344 & -0.168 ± 0.328 & -0.163 ± 0.368 \\
    \midrule
    Causal Survival Forest &  \multicolumn{5}{c}{-1.610 ± 0.057 (Not a meta-learner method)}  \\
    \midrule \\
    \textit{(Snapshot: 12 months)} \\
    % CoxPH & -- & -- & -- & -- & -- \\
    Random Survival Forest & 0.099 ± 0.430 & 0.050 ± 0.173 & -0.427 ± 0.148 & -0.544 ± 0.134 & -0.829 ± 0.127 \\
    DeepSurv & -2.650 ± 1.553 & 1.116 ± 0.605 & 0.262 ± 1.353 & -0.025 ± 1.381 & -0.347 ± 1.399 \\
    DeepHit & 0.822 ± 3.215 & 1.021 ± 0.535 & 0.355 ± 0.424 & 0.256 ± 0.420 & 0.148 ± 0.415 \\
    \midrule
    Causal Survival Forest &  \multicolumn{5}{c}{0.345 ± 0.097 (Not a meta-learner method)}  \\
    \bottomrule
    \end{tabular}}}
\end{table*}

The Average Treatment Effect (ATE) estimates for 9 and 12 months are summarized in Table~\ref{tab:table_ate_9_12}.
At 9 months, we observe predominantly negative ATEs across most models, suggesting that non-adherence is associated with shorter adverse event times to adverse events, consistent with expectations.
However, by 12 months, there is a notable increase in positive ATE estimates, indicating that adherence might correspond to shorter adverse event times to adverse events, which contradicts clinical expectations.

Examining the standard errors of the ATE estimates across different experimental repeats provides additional context for the unreliability of these numbers for longer time snapshots.
For example, while some negative ATEs at 9 months align with expectations, large standard errors, such as those seen for DeepSurv (\(-2.144 \pm 2.706\)), reduce confidence in these estimates.
Similarly, small ATE magnitudes close to zero, such as those reported by Random Survival Forest using the S-learner (\(-0.546 \pm 0.106\)), suggest minimal difference between adherence and non-adherence.

This unexpected trend of positive ATEs at 12 months can be attributed to two primary factors.
First, the underlying survival models at these later snapshots demonstrate weaker performance compared to earlier snapshots, as discussed in Appendix~\ref{apd:survival_analysis}.
This degradation in survival model accuracy likely propagates to the causal inference estimates, introducing additional unreliability.
Second, there is an inherent bias in the sampled cohort for these later snapshots.
By definition, for each time snapshot, its corresponding cohort consists of individuals who have not experienced an adverse event up to the later time points, regardless of their previous adherence history.
This selection process results in a subgroup that may not be representative of the full cohort, as it comprises patients with inherently better survival probabilities.
Specifically, individuals who survive to later time points without experiencing an adverse event are likely to have more favorable unmeasured health characteristics or environmental factors that contribute to their prolonged survival.
This could include better baseline health, fewer comorbid conditions, or access to additional social or medical resources not captured in the data.
As a result, the sampled cohort at these later time points is disproportionately composed of patients who are less likely to experience adverse events, irrespective of their adherence behavior.
Consequently, the observed ATE estimates at later snapshots are influenced by this selection bias, further complicating their interpretation.

From these results, we observe that the performance and reliability of causal inference analyses at later time points are diminished.
For instance, at 12 months, Random Survival Forest reports ATEs close to zero (\(0.099 \pm 0.430\) for the T-learner and \(0.050 \pm 0.173\) for the S-learner), indicating negligible differences between adherence and non-adherence.
Additionally, the large standard errors in estimates such as those from DeepSurv (\(1.116 \pm 0.605\)) and DeepHit (\(1.021 \pm 0.535\)) highlight the lack of reliability in these results.

The increased presence of positive ATEs highlights the compounded effects of weaker survival models and cohort selection bias.
While these findings provide some insights, they emphasize the need for cautious interpretation and underscore the importance of addressing these limitations in future analyses.

\subsection{Graphical Representation of ATE Trends Across Time Snapshots}
\label{apd:graphical_ate}

Figure~\ref{fig:graphical_ATE_trend} presents the Average Treatment Effect (ATE) estimates at different time snapshots \(\tau\) for the survival models experimented with (Cox Proportional Hazards (CoxPH), Random Survival Forest, DeepSurv, and DeepHit). Each subfigure illustrates the ATE trends over time for a specific survival model, incorporating multiple meta-learning approaches. To facilitate comparisons, the Causal Survival Forest (CSF) that is inherently a causal method is included in all plots.

\begin{figure*}[!htp]
\floatconts
  {fig:graphical_ATE_trend}
  {\caption{Estimated Average Treatment Effect (ATE) of non-adherence across time snapshots \(\tau\) for different survival models. Each panel corresponds to a specific survival model (CoxPH, Random Survival Forest, DeepSurv, and DeepHit), with ATE estimates reported for multiple meta-learning approaches (T-learner, S-learner, and Matching). The Causal Survival Forest (CSF) is included in all plots as a baseline for comparison. Shaded regions represent standard deviations, illustrating the uncertainty in the estimates.}}
  {%
    \subfigure[CoxPH]{\label{fig:graphical_ate_coxph}%
      \includegraphics[width=.45\linewidth]{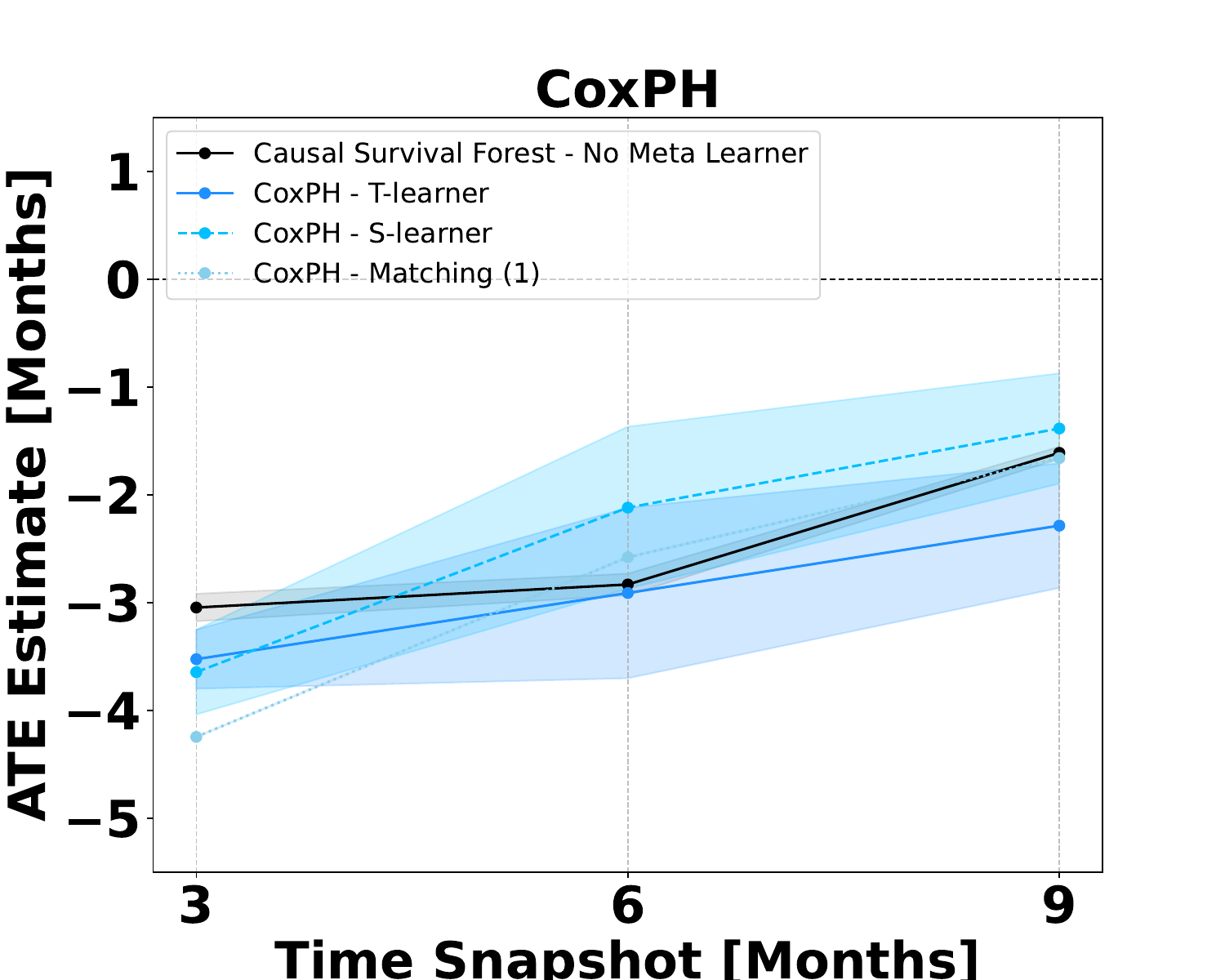}}%
    \subfigure[RandomSurvivalForest]{\label{fig:graphical_ate_rsf}%
      \includegraphics[width=.45\linewidth]{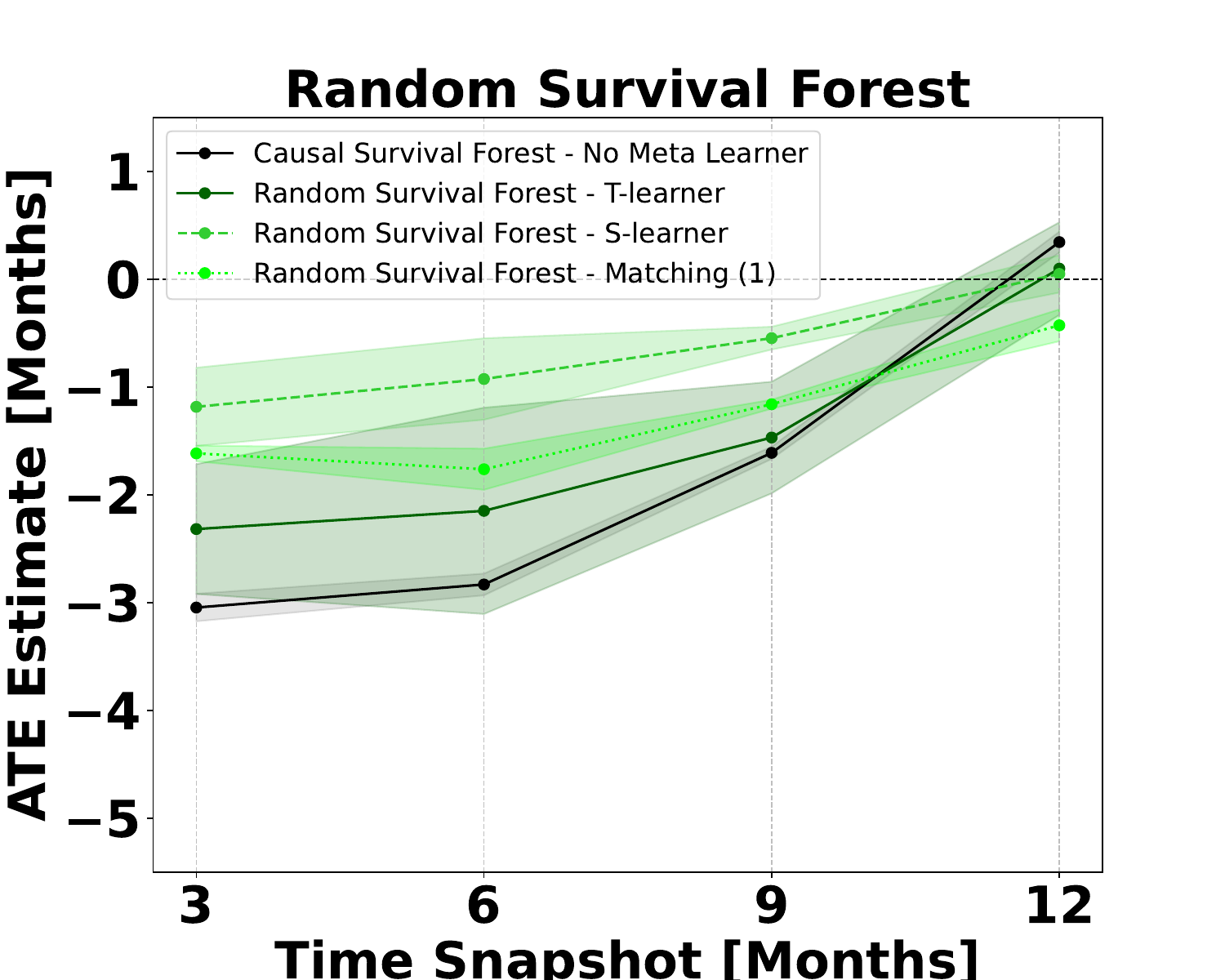}}%
      \\
    \subfigure[DeepSurv]{\label{fig:graphical_ate_deepsurv}%
      \includegraphics[width=.45\linewidth]{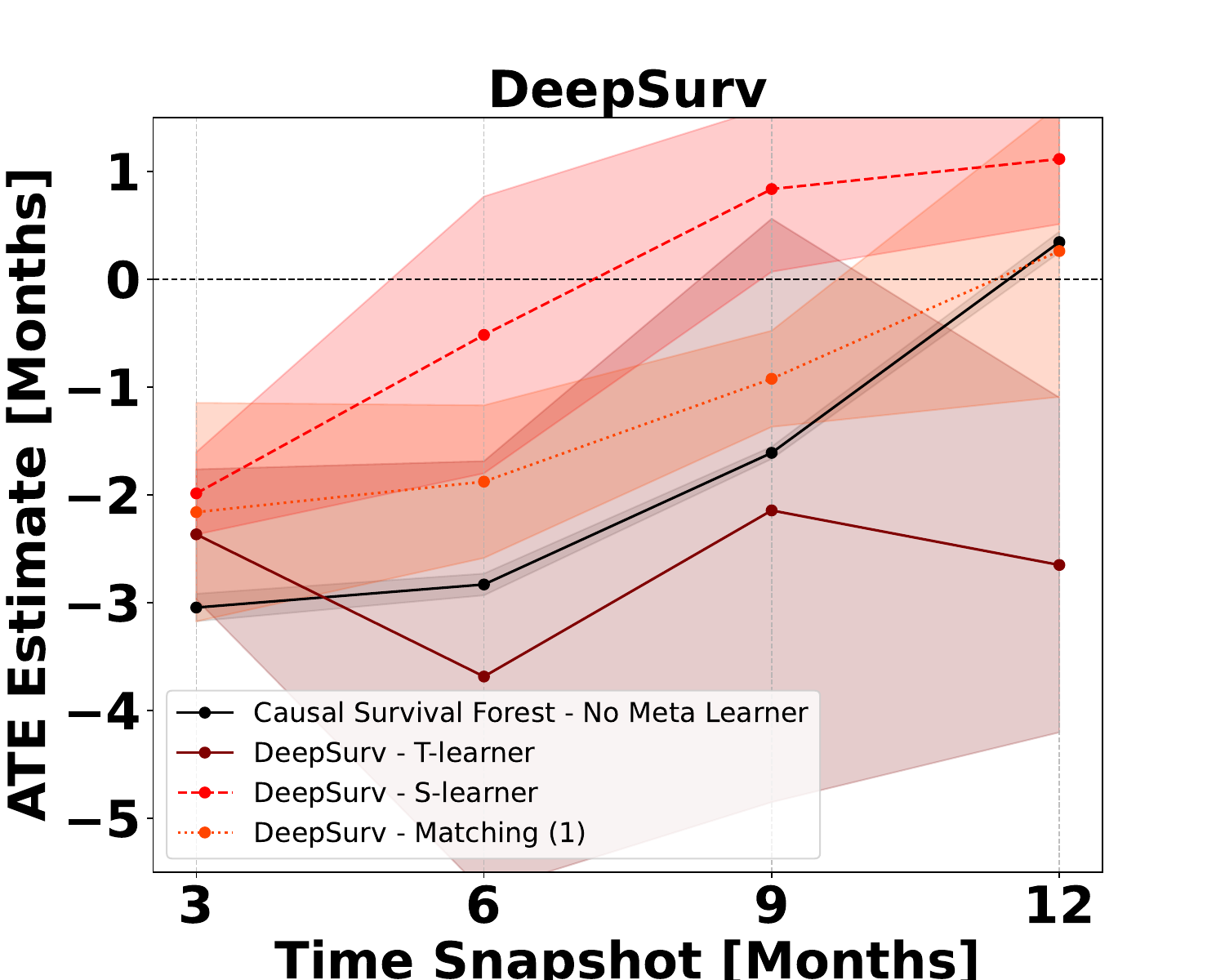}}%
    \subfigure[DeepHit]{\label{fig:graphical_ate_deephit}%
      \includegraphics[width=.46\linewidth]{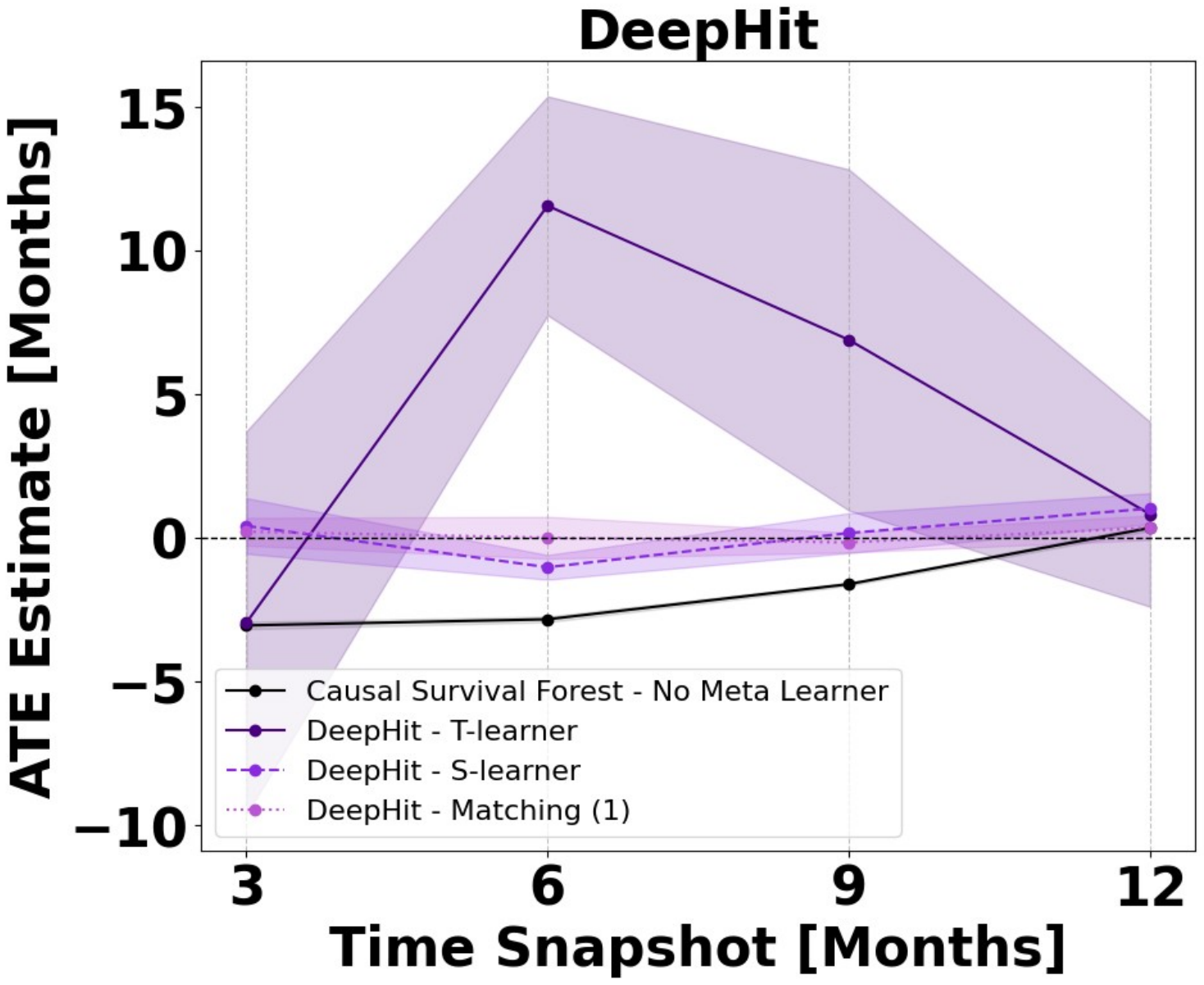}}%
  }
\end{figure*}

The graphical analysis of ATE estimates over time highlights consistent trends across different survival models and meta-learning strategies. Across CoxPH, Random Survival Forest (RSF), and DeepSurv, a clear ranking of ATE estimates emerges, with the T-learner consistently producing the most negative values, the S-learner yielding the least negative values, and the Matching-based estimator falling in between. This pattern holds across all time snapshots, suggesting that the T-learner amplifies the estimated effect of non-adherence, whereas the S-learner provides more conservative estimates. The Causal Survival Forest (CSF), included in all plots, provides relatively stable estimates that align closely with those from RSF and CoxPH.

As time progresses, a general upward trend is observed in ATE estimates, with some models—particularly RSF and CSF—approaching zero or slightly positive values at 12 months. However, this shift should be interpreted with caution, as it is likely influenced by \textit{survivor bias}, where the population at later time points consists only of individuals who have not yet experienced an adverse event. As discussed in Section~\ref{sec:04_results_causal}, this selection dynamic skews the estimates toward a subset of individuals who may inherently be at lower risk, leading to a potential underestimation of the true treatment effect. Even when an ATE estimate appears positive, the confidence intervals often extend into the negative range, indicating that the treatment effect of non-adherence could still be detrimental. The increasing standard deviation at later snapshots further suggests greater uncertainty, reinforcing the notion that these estimates become less reliable over longer time snapshots. While we account for observable covariates and risk scores, unmeasured resilience factors or external support systems may still differentiate the remaining population, contributing to the observed shift. Additionally, residual time-varying confounding may exacerbate this effect, as evolving risk factors influencing survival may not be fully captured by standard covariate adjustments. 

DeepHit continues to stand out as the least stable model for ATE estimation, producing extreme deviations from other methods, particularly at 6 months, where the T-learner reports an ATE of \(11.551 \pm 3.806\). As noted in Section~\ref{sec:04_results_causal}, this erratic behavior aligns with its poor predictive performance, where it consistently underperformed in concordance and calibration metrics. The instability in DeepHit’s survival predictions likely propagates into its ATE estimates, making it unreliable for causal inference in our experiments. In contrast, CoxPH and RSF provide more stable and interpretable estimates across time, reinforcing their reliability in treatment effect estimation. Lastly, CSF also gives estimates that remain within a reasonable range while avoiding the extreme fluctuations observed in DeepHit.

These results suggest that while ATE estimates generally indicate a detrimental effect of non-adherence at earlier time points, later snapshots introduce more uncertainty due to both selection dynamics and model instability. The consistent ordering of ATE estimates across meta-learners further underscores the systematic differences in how these approaches estimate treatment effects, with the T-learner capturing the largest negative effects, the S-learner being the most conservative, and Matching-based methods providing intermediate estimates. Given the growing standard deviations and the shifting ATE trends at later time points, care must be taken when interpreting long-term treatment effects, as unmeasured confounding and survival bias may increasingly distort the estimates.

\subsection{Unadjusted Survival Curves}
\label{apd:unadjusted_survival}

\begin{figure}[!tbp]
    \centering
    \includegraphics[width=\linewidth]{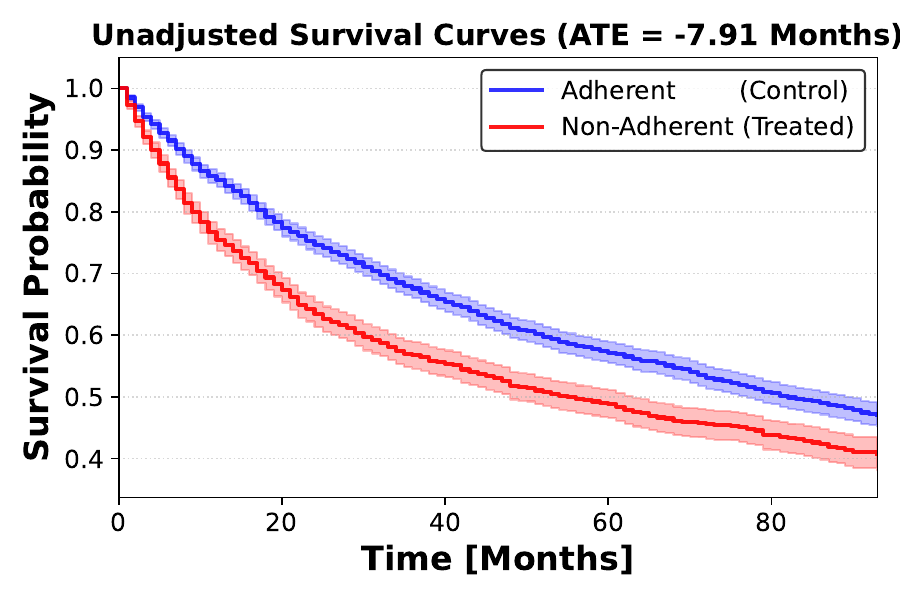}
    \caption{Unadjusted Survival Curves obtained from Kaplan-Meier Estimator for \textcolor{blue}{Adherent} (Control) and \textcolor{red}{Non-Adherent} (Treated) Patients. The Average Treatment Effect from the Restricted Mean Event Time difference between the groups is -7.91 Months.}
    \label{fig:unadjusted_survival_curve}
\end{figure}

To provide an initial assessment of the survival differences between adherent and non-adherent patients, we compute unadjusted survival curves using the Kaplan-Meier estimator. 
Figure~\ref{fig:unadjusted_survival_curve} illustrates the survival probability over time for both groups, highlighting the divergence in survival trajectories. 
Non-adherent patients exhibit a markedly shorter adverse event time, with an estimated Average Treatment Effect (ATE) of -7.91 months based on the difference in Restricted Mean Event Time (RMET) between the two groups. 
In Table~\ref{tab:time_of_first_event} we also show the raw average time to the first event or censoring across the cohort. 
While right-censoring occurs, on average, at 62.6 months, composite adverse events, including mortality, involuntary hospitalization, or jail booking, occur much earlier, averaging 25.2 months. 
These raw results of difference in adverse event time between the non-adherent and adherent groups set the stage for the survival modeling and causal inference analyses of our paper, which account for confounding factors and provide adjusted estimates of adherence's impact on time-to-first-event.

\begin{table}[hbtp]
    \setlength{\tabcolsep}{4pt}
    \footnotesize
    \floatconts
      {tab:time_of_first_event}
      {\caption{Average Time of the First Event or Censoring in Months across All Patients}}
      {\begin{tabular}{l c}
    \toprule
    & First Event Time \\
    \midrule
    Right Censoring & 62.6 ± 29.1 \\
    Composite Adverse Event &  25.2 ± 22.6  \\
    \bottomrule
    \end{tabular}}
\end{table}

\vspace{-1em}
\subsection{Complete Ablation Tables}
\label{apd:ablation}
\vspace{-0.5em}

The complete ablation tables summarize the impact of excluding county-provided risk scores on Average Treatment Effect (ATE) estimates across different models and time snapshots.
Table~\ref{tab:table_ablation_CoxPH} presents the ATE estimates for the Cox Proportional Hazards (CoxPH) model.
The removal of risk scores results in consistently more negative ATEs across all snapshots, with increases in magnitude most evident at earlier time points, such as 3 months, where the ATE shifts from \(-3.524 \pm 0.274\) to \(-5.432 \pm 0.344\) using the T-learner.
This suggests that risk scores effectively capture confounding factors that partially attenuate the observed impact of non-adherence.

Table~\ref{tab:table_ablation_RandomSurvivalForest} shows the ablation results for the Random Survival Forest (RSF) model.
Consistent with CoxPH, ablated models produce more negative ATE estimates, highlighting the importance of including risk scores to control for unmeasured confounding.
For instance, at 3 months, the ATE using the T-learner changes from \(-2.317 \pm 0.602\) to \(-5.915 \pm 0.433\), a substantial difference.
This trend persists across later snapshots, though the effect size diminishes as the cohort composition evolves.

DeepSurv ablation results are detailed in Table~\ref{tab:table_ablation_DeepSurv}.
The standard errors for DeepSurv ATE estimates remain large, both in full and ablated setups, reflecting its limited reliability for causal inference.
Despite this, the removal of risk scores amplifies the negative ATE values, indicating their critical role in adjusting for confounding.
For example, at 6 months, the S-learner's ATE shifts from \(-0.516 \pm 1.284\) to \(-3.130 \pm 1.551\) following ablation.

Table~\ref{tab:table_ablation_DeepHit} provides the results for DeepHit.
The ATE estimates show high variability, particularly in the ablated models, where standard errors are considerable.
For example, at 9 months, the T-learner ATE for the full model is \(6.887 \pm 5.930\), which reduces to \(2.075 \pm 3.780\) post-ablation.
This variability undermines the interpretability of DeepHit's estimates, underscoring its limitations for both survival and causal analyses in this dataset.

The Causal Survival Forest (CSF) results, presented in Table~\ref{tab:table_ablation_CausalSurvivalForest}, follow a similar pattern.
The magnitude of ATEs increases following ablation, particularly at earlier snapshots.
At 3 months, the ATE changes from \(-3.045 \pm 0.128\) to \(-7.816 \pm 0.064\), indicating a substantial impact of excluding risk scores.
This trend diminishes at later time points, where selection bias and survivor effects likely play a greater role.

Overall, the ablation results reveal consistent trends across all models.
The exclusion of county-provided risk scores leads to more pronounced negative ATE estimates, highlighting their utility as proxies for unmeasured confounders.
While the observed differences are most pronounced at earlier snapshots, the increasing standard errors at later time points suggest a need for cautious interpretation.
These findings underscore the importance of comprehensive feature inclusion in survival and causal modeling to improve robustness and accuracy.
The trends observed in later time snapshots align with the biases discussed in Appendix~\ref{apd:causal_inference}, where selection bias and declining survival model performance introduce additional challenges in interpreting ATE estimates.
This further reinforces the difficulty in drawing reliable causal conclusions at these later snapshots, as the subset of patients included becomes increasingly non-representative of the original cohort.

\begin{table*}[hbtp]
    \setlength{\tabcolsep}{4pt}
    % \footnotesize
    \floatconts
      {tab:table_ablation_CoxPH}
      {\caption{Causal Inference Ablation ATE Estimates for CoxPH}}
      {\adjustbox{max width=.825\textwidth}{\begin{tabular}{l c c c c c }
    \toprule
    &T-learner & S-learner & Matching (1) & Matching (5) & Matching (20) \\
    \midrule
    \textit{(3 months)}
 &  &  &  &  \\
    CoxPH (Full) & -3.524 ± 0.274 & -3.644 ± 0.393 & -4.245 ± 0.005 & -4.280 ± 0.003 & -4.406 ± 0.002 \\
    CoxPH (Ablation) & -5.432 ± 0.344 & -5.460 ± 0.429 & -6.020 ± 0.004 & -6.105 ± 0.004 & -6.184 ± 0.002 \\\\
    \textit{(6 months)}
 &  &  &  &  \\
    CoxPH (Full) & -2.910 ± 0.791 & -2.118 ± 0.751 & -2.578 ± 0.009 & -2.639 ± 0.006 & -2.900 ± 0.007 \\
    CoxPH (Ablation) & -4.933 ± 0.815 & -4.191 ± 0.720 & -4.220 ± 0.007 & -4.287 ± 0.003 & -4.453 ± 0.003 \\\\
    \textit{(9 months)}
 &  &  &  &  \\
    CoxPH (Full) & -2.285 ± 0.578 & -1.384 ± 0.512 & -1.658 ± 0.006 & -1.790 ± 0.002 & -2.032 ± 0.003 \\
    CoxPH (Ablation) & -3.397 ± 0.570 & -2.392 ± 0.495 & -2.148 ± 0.006 & -2.186 ± 0.003 & -2.387 ± 0.002 \\\\
    % \textit{(12 months)}
 % &  &  &  &  \\
    % CoxPH (Full) & -- & -- & -- & -- & -- \\
    % CoxPH (Ablation) & -- & -- & -- & -- & -- \\\\
    \bottomrule
    \end{tabular}}}
\end{table*}

\begin{table*}[hbtp]
    \setlength{\tabcolsep}{4pt}
    % \footnotesize
    \floatconts
      {tab:table_ablation_RandomSurvivalForest}
      {\caption{Causal Inference Ablation ATE Estimates for RandomSurvivalForest}}
      {\adjustbox{max width=0.9\textwidth}{\begin{tabular}{l c c c c c }
    \toprule
    &T-learner & S-learner & Matching (1) & Matching (5) & Matching (20) \\
    \midrule
    \textit{(3 months)}
 &  &  &  &  \\
    RandomSurvivalForest (Full) & -2.317 ± 0.602 & -1.183 ± 0.363 & -1.615 ± 0.073 & -1.538 ± 0.054 & -1.978 ± 0.050 \\
    RandomSurvivalForest (Ablation) & -5.915 ± 0.433 & -4.938 ± 0.370 & -5.668 ± 0.078 & -5.647 ± 0.103 & -5.712 ± 0.077 \\\\
    \textit{(6 months)}
 &  &  &  &  \\
    RandomSurvivalForest (Full) & -2.148 ± 0.957 & -0.925 ± 0.378 & -1.762 ± 0.191 & -1.569 ± 0.135 & -1.807 ± 0.109 \\
    RandomSurvivalForest (Ablation) & -5.210 ± 0.898 & -3.805 ± 0.767 & -4.007 ± 0.158 & -3.992 ± 0.163 & -4.307 ± 0.144 \\\\
    \textit{(9 months)}
 &  &  &  &  \\
    RandomSurvivalForest (Full) & -1.468 ± 0.518 & -0.546 ± 0.106 & -1.159 ± 0.042 & -1.242 ± 0.050 & -1.540 ± 0.055 \\
    RandomSurvivalForest (Ablation) & -3.848 ± 0.580 & -2.579 ± 0.394 & -2.554 ± 0.079 & -2.704 ± 0.077 & -2.907 ± 0.076 \\\\
    \textit{(12 months)}
 &  &  &  &  \\
    RandomSurvivalForest (Full) & 0.099 ± 0.430 & 0.050 ± 0.173 & -0.427 ± 0.148 & -0.544 ± 0.134 & -0.829 ± 0.127 \\
    RandomSurvivalForest (Ablation) & -2.238 ± 0.652 & -0.986 ± 0.545 & -1.090 ± 0.022 & -1.343 ± 0.044 & -1.811 ± 0.061 \\\\
    \bottomrule
    \end{tabular}}}
\end{table*}

\begin{table*}[hbtp]
    \setlength{\tabcolsep}{4pt}
    % \footnotesize
    \floatconts
      {tab:table_ablation_DeepSurv}
      {\caption{Causal Inference Ablation ATE Estimates for DeepSurv}}
      {\adjustbox{max width=.825\textwidth}{\begin{tabular}{l c c c c c }
    \toprule
    &T-learner & S-learner & Matching (1) & Matching (5) & Matching (20) \\
    \midrule
    \textit{(3 months)}
 &  &  &  &  \\
    DeepSurv (Full) & -2.366 ± 0.603 & -1.986 ± 0.382 & -2.160 ± 1.014 & -2.101 ± 1.051 & -2.312 ± 1.223 \\
    DeepSurv (Ablation) & -4.775 ± 0.853 & -3.842 ± 2.052 & -4.483 ± 2.070 & -4.529 ± 2.067 & -4.613 ± 2.061 \\\\
    \textit{(6 months)}
 &  &  &  &  \\
    DeepSurv (Full) & -3.685 ± 1.998 & -0.516 ± 1.284 & -1.877 ± 0.708 & -1.927 ± 0.594 & -2.152 ± 0.422 \\
    DeepSurv (Ablation) & -4.406 ± 2.343 & -3.130 ± 1.551 & -2.124 ± 1.442 & -2.129 ± 1.434 & -2.189 ± 1.534 \\\\
    \textit{(9 months)}
 &  &  &  &  \\
    DeepSurv (Full) & -2.144 ± 2.706 & 0.837 ± 0.767 & -0.923 ± 0.446 & -0.928 ± 0.436 & -1.021 ± 0.558 \\
    DeepSurv (Ablation) & -4.979 ± 2.941 & -2.165 ± 2.317 & -2.279 ± 1.823 & -2.366 ± 1.887 & -2.483 ± 2.043 \\\\
    \textit{(12 months)}
 &  &  &  &  \\
    DeepSurv (Full) & -2.650 ± 1.553 & 1.116 ± 0.605 & 0.262 ± 1.353 & -0.025 ± 1.381 & -0.347 ± 1.399 \\
    DeepSurv (Ablation) & -2.267 ± 1.834 & -0.593 ± 1.165 & -0.415 ± 1.082 & -0.700 ± 1.127 & -1.097 ± 1.174 \\\\
    \bottomrule
    \end{tabular}}}
\end{table*}

\begin{table*}[hbtp]
    \setlength{\tabcolsep}{4pt}
    % \footnotesize
    \floatconts
      {tab:table_ablation_DeepHit}
      {\caption{Causal Inference Ablation ATE Estimates for DeepHit}}
      {\adjustbox{max width=.825\textwidth}{\begin{tabular}{l c c c c c }
    \toprule
    &T-learner & S-learner & Matching (1) & Matching (5) & Matching (20) \\
    \midrule
    \textit{(3 months)}
 &  &  &  &  \\
    DeepHit (Full) & -2.956 ± 6.663 & 0.417 ± 0.967 & 0.215 ± 0.481 & 0.241 ± 0.476 & 0.200 ± 0.501 \\
    DeepHit (Ablation) & -5.495 ± 3.492 & -0.936 ± 1.746 & -0.774 ± 0.638 & -0.759 ± 0.622 & -0.806 ± 0.646 \\\\
    \textit{(6 months)}
 &  &  &  &  \\
    DeepHit (Full) & 11.551 ± 3.806 & -1.022 ± 0.428 & 0.007 ± 0.727 & 0.054 ± 0.804 & 0.140 ± 0.876 \\
    DeepHit (Ablation) & -0.249 ± 4.701 & -2.236 ± 1.570 & -1.969 ± 0.660 & -1.915 ± 0.664 & -1.927 ± 0.661 \\\\
    \textit{(9 months)}
 &  &  &  &  \\
    DeepHit (Full) & 6.887 ± 5.930 & 0.169 ± 0.692 & -0.160 ± 0.344 & -0.168 ± 0.328 & -0.163 ± 0.368 \\
    DeepHit (Ablation) & 2.075 ± 3.780 & -1.351 ± 0.895 & -1.273 ± 1.150 & -1.309 ± 1.178 & -1.407 ± 1.222 \\\\
    \textit{(12 months)}
 &  &  &  &  \\
    DeepHit (Full) & 0.822 ± 3.215 & 1.021 ± 0.535 & 0.355 ± 0.424 & 0.256 ± 0.420 & 0.148 ± 0.415 \\
    DeepHit (Ablation) & 6.172 ± 3.723 & 1.118 ± 0.347 & 0.680 ± 0.429 & 0.397 ± 0.315 & 0.078 ± 0.232 \\\\
    \bottomrule
    \end{tabular}}}
\end{table*}

\begin{table*}[hbtp]
    \setlength{\tabcolsep}{4pt}
    \footnotesize
    \floatconts
      {tab:table_ablation_CausalSurvivalForest}
      {\caption{Causal Inference Ablation ATE Estimates for CausalSurvivalForest}}
      {\begin{tabular}{l c }
    \toprule
    &ATE \\
    \midrule
    \textit{(3 months)}
 \\
    CausalSurvivalForest (Full) & -3.045 ± 0.128 \\
    CausalSurvivalForest (Ablation) & -7.816 ± 0.064 \\\\
    \textit{(6 months)}
 \\
    CausalSurvivalForest (Full) & -2.831 ± 0.101 \\
    CausalSurvivalForest (Ablation) & -5.189 ± 0.047 \\\\
    \textit{(9 months)}
 \\
    CausalSurvivalForest (Full) & -1.610 ± 0.057 \\
    CausalSurvivalForest (Ablation) & -2.833 ± 0.077 \\\\
    \textit{(12 months)}
 \\
    CausalSurvivalForest (Full) & 0.345 ± 0.097 \\
    CausalSurvivalForest (Ablation) & -1.561 ± 0.104 \\\\
    \bottomrule
    \end{tabular}}
\end{table*}
\newpage

\clearpage
\section{Medication Subgroup Analysis}
\label{apd:medication_subgroup_analysis}
\numberwithin{equation}{section}
\numberwithin{figure}{section}
\numberwithin{table}{section}

\subsection{Individual Treatment Effect for Medication Types}
\label{apd:ite_medication}

\begin{figure*}[htbp]
\floatconts
  {fig:ite_medication}
  {\caption{Distribution of Estimated Individual Treatment Effects (ITE) for the top four medications—\textcolor{blue}{risperidone}, \textcolor{orange}{aripiprazole}, \textcolor{ForestGreen}{olanzapine}, and \textcolor{red}{haloperidol}—evaluated using different survival models and T-learner at time snapshot \(\tau=3\) month. 
  The mean treatment effect (ATE) for each medication is provided in the legend.
  (a)-(c) show T-learner ITEs for (a) CoxPH, (b) DeepSurv, (c) Random Survival Forest. 
  (d) shows the ITE for Causal Survival Forest}}
  {%
    \subfigure[\footnotesize (CoxPH, T-Learner)]{\label{fig:ite_medication_coxph}%
      \includegraphics[width=.22\linewidth]{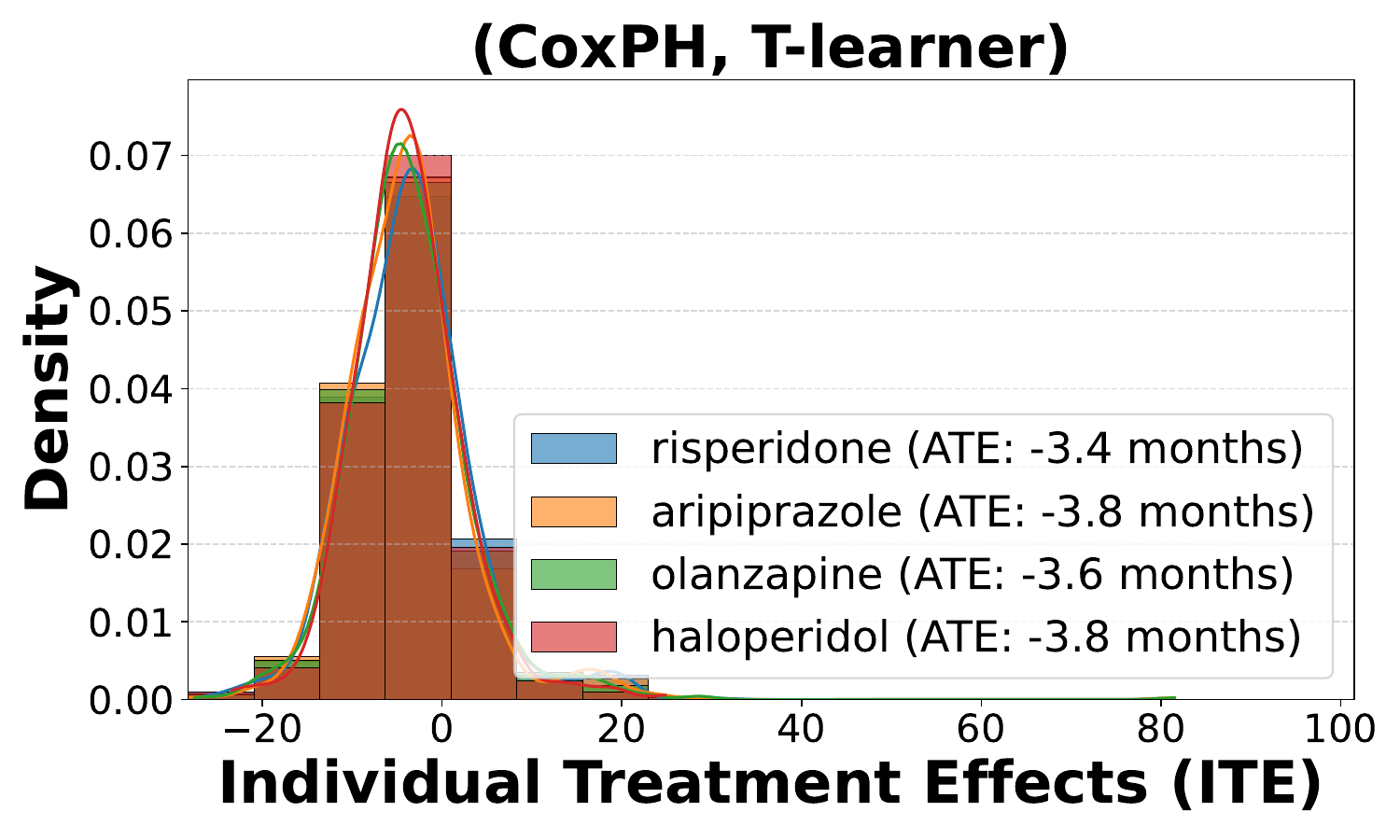}}%
    \qquad
    \subfigure[\footnotesize (DeepSurv, T-Learner)]{\label{fig:ite_medication_deepsurv}%
      \includegraphics[width=.22\linewidth]{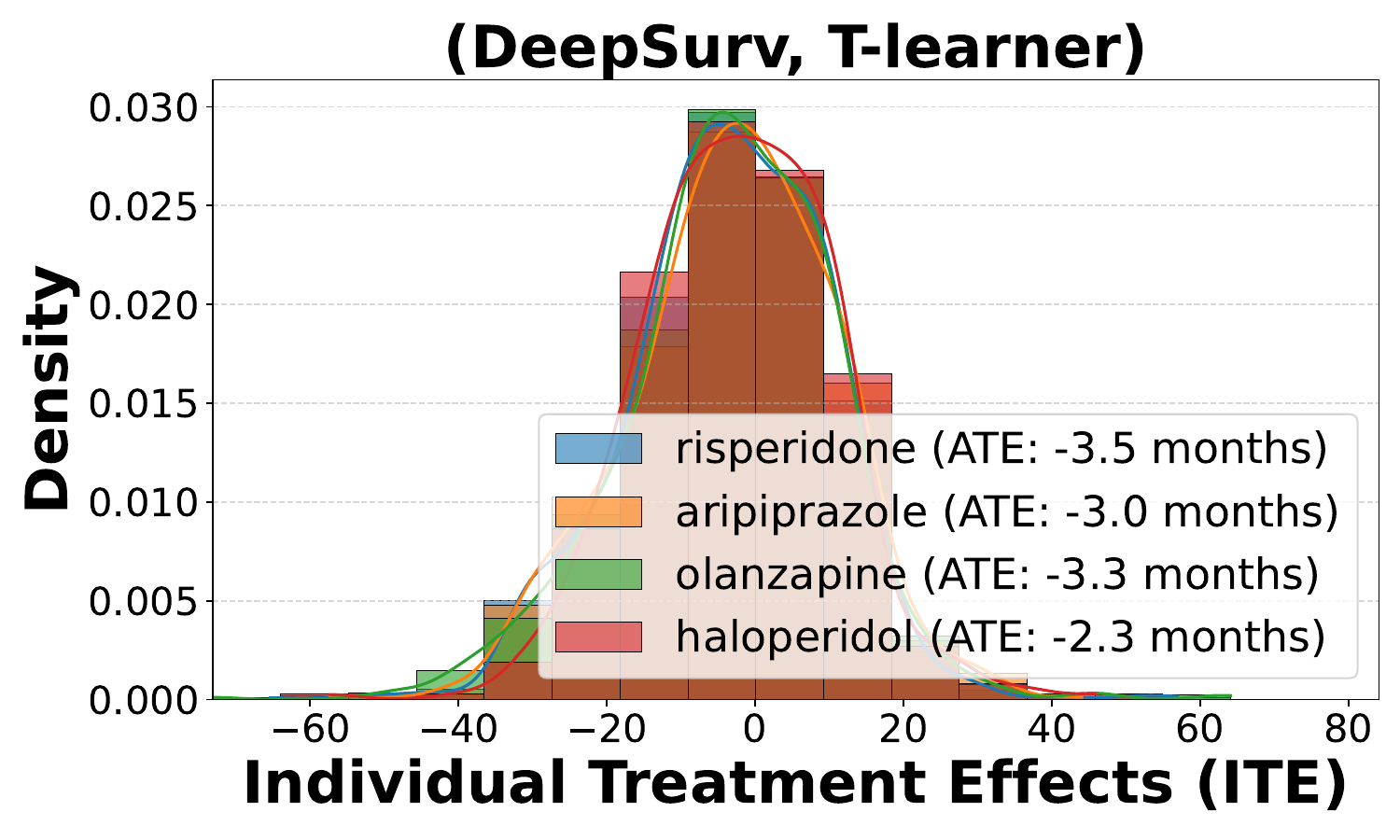}}%
    \qquad
    \subfigure[\footnotesize (RandomSurvivalForest, T-Learner)]{\label{fig:ite_medication_random_survival_forest}%
      \includegraphics[width=.22\linewidth]{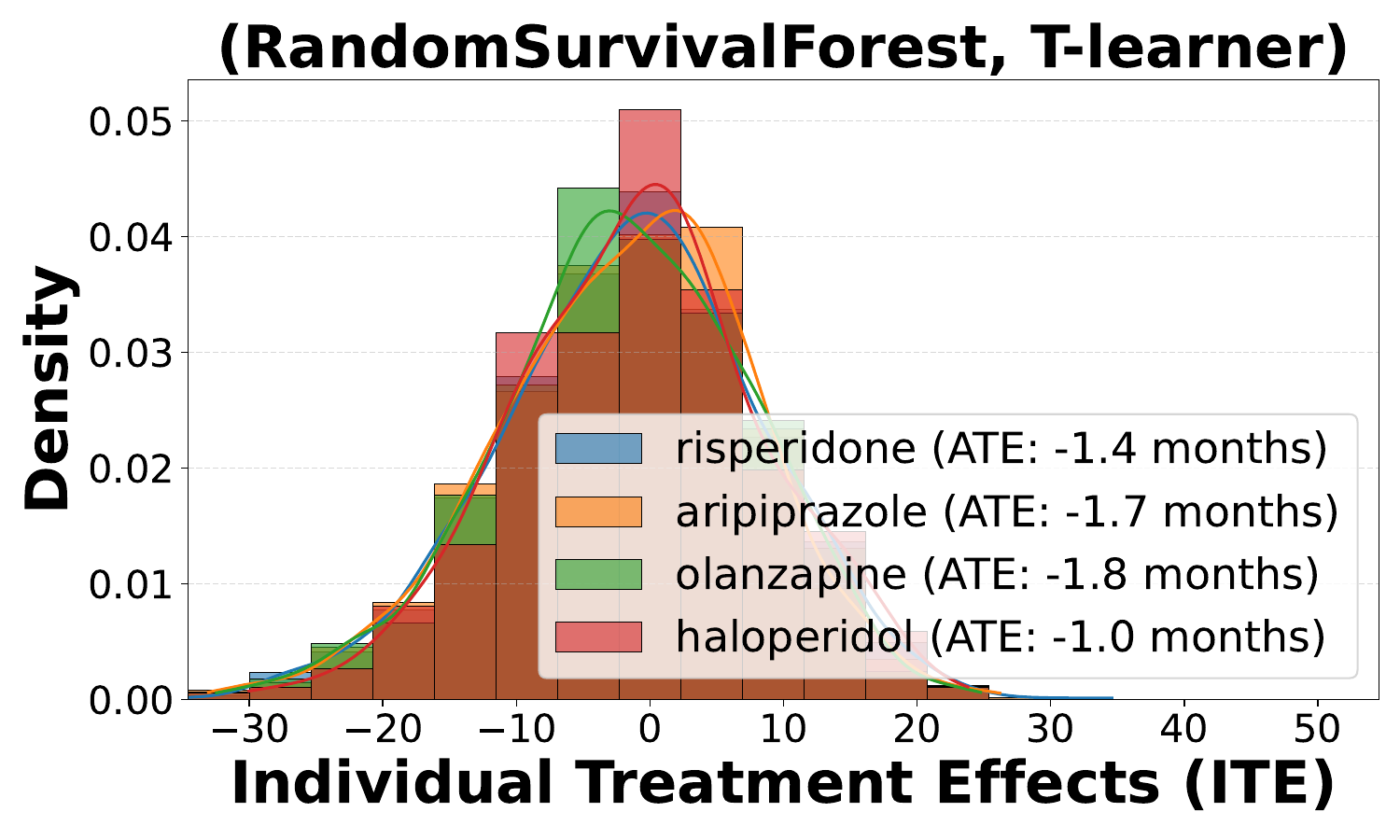}}%
    \qquad
    \subfigure[\footnotesize CausalSurvivalForest]{\label{fig:ite_medical_causal_survival_forest}%
      \includegraphics[width=.21\linewidth]{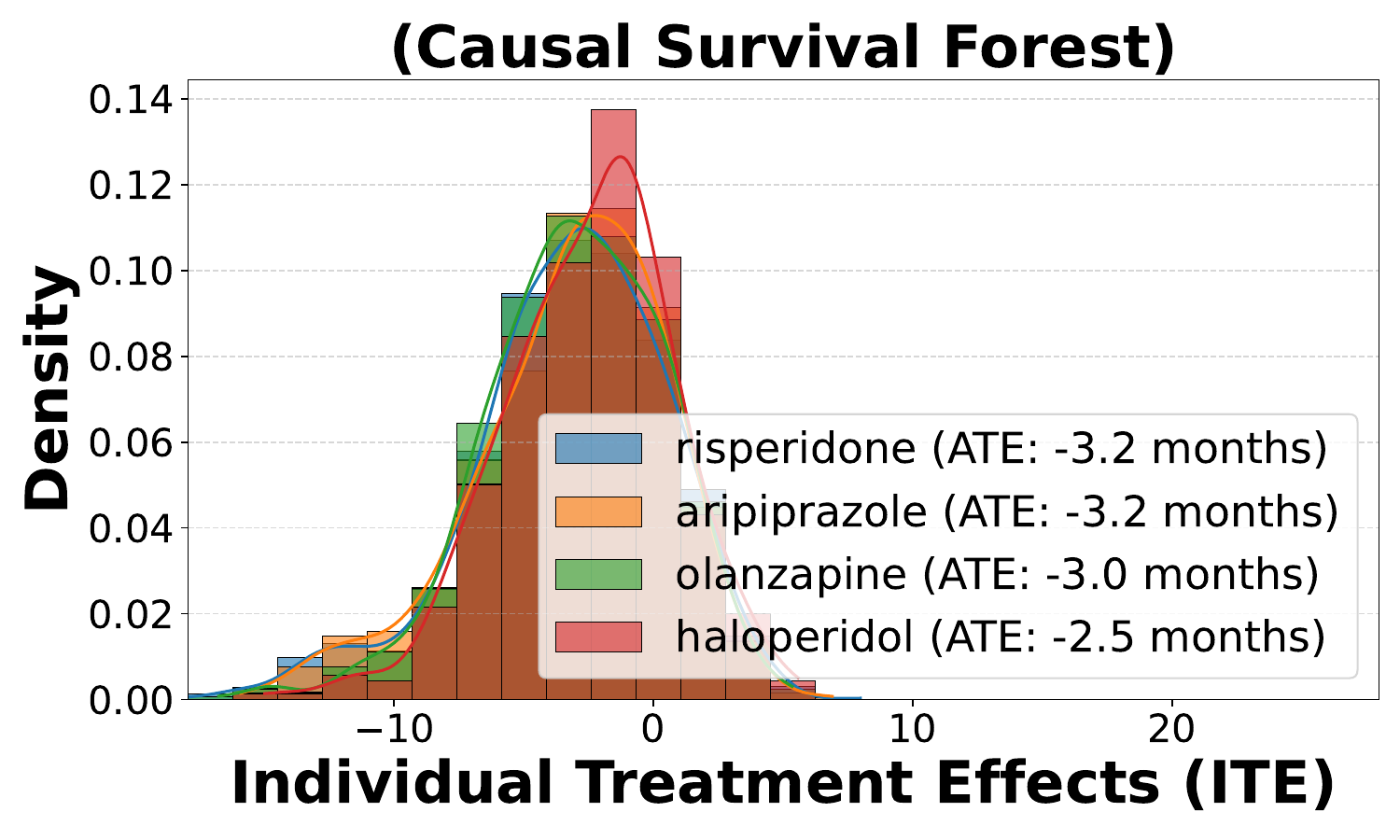}}%
  }
\end{figure*}

This section presents the distributions of Estimated Individual Treatment Effects (ITE) for the four most prevalent antipsychotic medications—risperidone, aripiprazole, olanzapine, and haloperidol—evaluated using various survival models and meta-learners.

Figure~\ref{fig:ite_medication} shows the ITE distributions at the time snapshot \(\tau=3\) months.
The plots compare the results from (a) Cox Proportional Hazards (CoxPH) model with T-learner, (b) DeepSurv with T-learner, (c) Random Survival Forest (RSF) with T-learner, and (d) Causal Survival Forest (CSF).
Each histogram is overlaid with a kernel density estimate to visualize the distribution shapes better.
The legend provides the average treatment effect (ATE) for each medication.

From Figure~\ref{fig:ite_medication}, the Causal Survival Forest (Figure~\ref{fig:ite_medical_causal_survival_forest}) produces the narrowest ITE distributions, indicating reduced variability and more consistent estimates compared to other models. 
The Random Survival Forest (Figure~\ref{fig:ite_medication_random_survival_forest}) shows slightly wider ITE distributions compared to the Causal Survival Forest but narrower distributions than the CoxPH (Figure~\ref{fig:ite_medication_coxph}) and DeepSurv (Figure~\ref{fig:ite_medication_deepsurv}) models, which exhibit the broadest ITE variability.

The ATE values across models exhibit different patterns and rankings for the medications. 
For CoxPH, haloperidol and aripiprazole share the most negative ATE values (\(-3.8\) months), followed by olanzapine (\(-3.6\)) and risperidone (\(-3.4\)). 
DeepSurv assigns the most negative ATE to risperidone (\(-3.5\)), followed by olanzapine (\(-3.3\)), aripiprazole (\(-3.0\)), and haloperidol (\(-2.3\)). 
In contrast, Random Survival Forest shows markedly smaller magnitude for all ATEs, with olanzapine (\(-1.8\)) being the most negative, followed by aripiprazole (\(-1.7\)), risperidone (\(-1.4\)), and haloperidol (\(-1.0\)). 
The Causal Survival Forest shows a pattern where risperidone and aripiprazole are tied at \(-3.2\), followed by olanzapine (\(-3.0\)) and haloperidol (\(-2.5\)).

These slight differences highlight that the ATE values for each medication within a given model are very similar, indicating no meaningful differences in treatment effects across the four most prevalent medications in our data.
While the ranking of ATEs for the medications varies slightly across models, the magnitudes are close within each model, further emphasizing the lack of significant differences among medications for any specific survival model. 
Additionally, we observe variability in ATE estimates across different models. 
The ATEs estimated by CoxPH, DeepSurv, and the Causal Survival Forest are closer to each other in magnitude, while the Random Survival Forest produces shorter ATE estimates compared to the other models. 
Another noteworthy observation is that the ATEs are consistently negative across all medications and models, which agrees with our previous results in Section~\ref{sec:experiments_results}, demonstrating shorter adverse event times for non-adherence regardless of the specific medication. 
This consistency in negative ATEs highlights the robustness of the association between non-adherence and reduced adverse event time in our analysis across the most prevalent antipsychotic medications.

\subsection{Individual Treatment Effects for Different Antipsychotic Administration Type}
\label{apd:ite_injectable_all}

\begin{figure*}[!ht]
\floatconts
  {fig:ite_injectable_all}
  {\vspace{-2em} \caption{
  Distribution of Estimated Individual Treatment Effects (ITE) for different medication adherence groups: \textcolor{ForestGreen}{injectable}, \textcolor{orange}{non-injectable}, and \textcolor{purple}{not covered} at time snapshot \(\tau=3\) months. 
  Each plot's legend highlights the Average Treatment Effect (ATE) in months for the groups covered by \textcolor{ForestGreen}{injectable} medication, \textcolor{orange}{non-injectable} medication, and those \textcolor{purple}{not covered} by any medication.
  (a)-(d) Show CoxPH survival model paired with various causal methods.
  (e)-(h) Show DeepSurv survival model paired with various causal methods.
  (i)-(l) Show Random Survival Forest paired with various causal methods.
  (m) shows Causal Survival Forest.
  Causal methods across each column is T-learner, S-learner, Matching method with 1 match, and Matching method with 5 matches.}}
  {%
    \subfigure[\footnotesize (CoxPH, T-Learner)]{\label{fig:ite_injectable_coxph_t_learner}%
      \includegraphics[width=.22\linewidth]{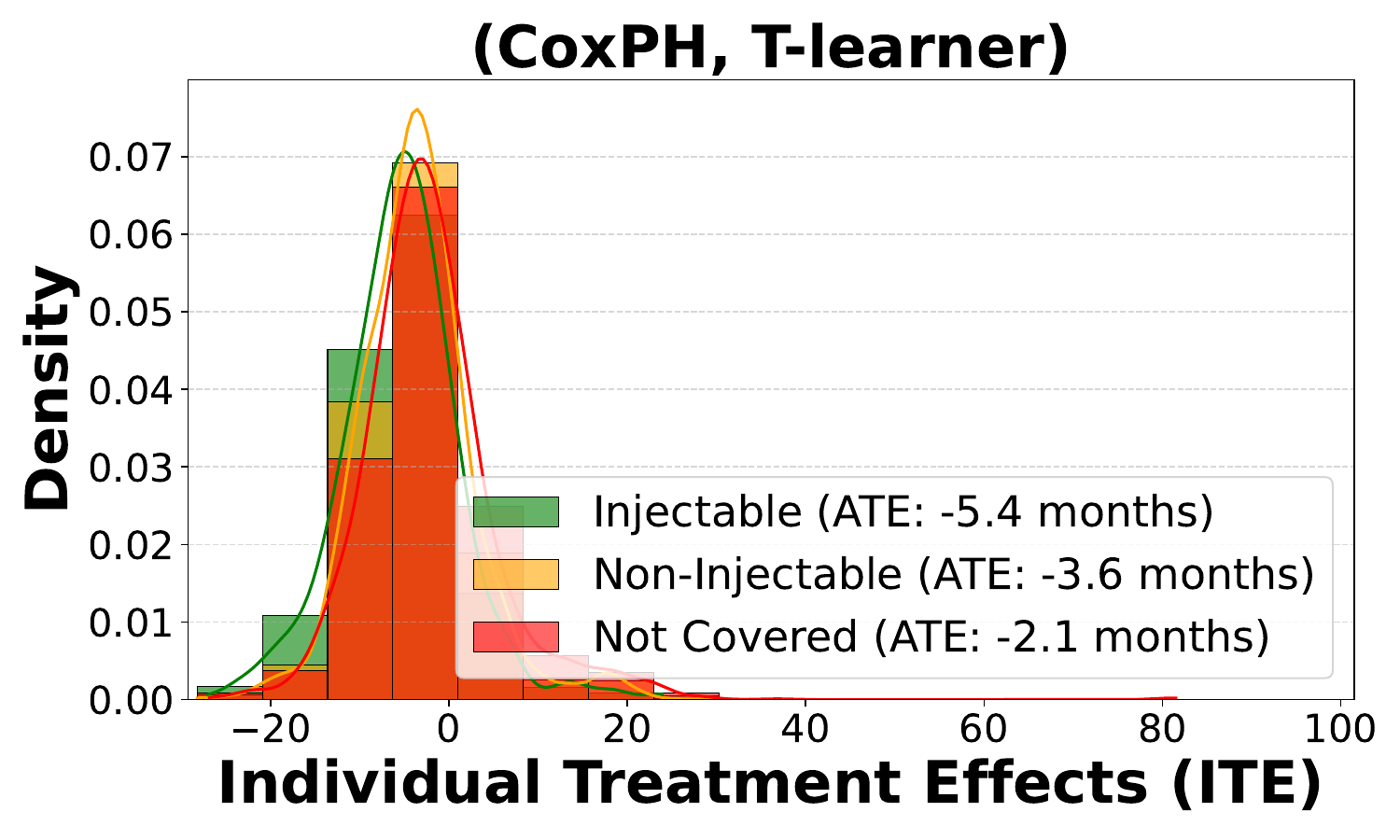}}%
    \qquad
    \subfigure[\footnotesize (CoxPH, S-Learner)]{\label{fig:ite_injectable_coxph_s_learner}%
      \includegraphics[width=.22\linewidth]{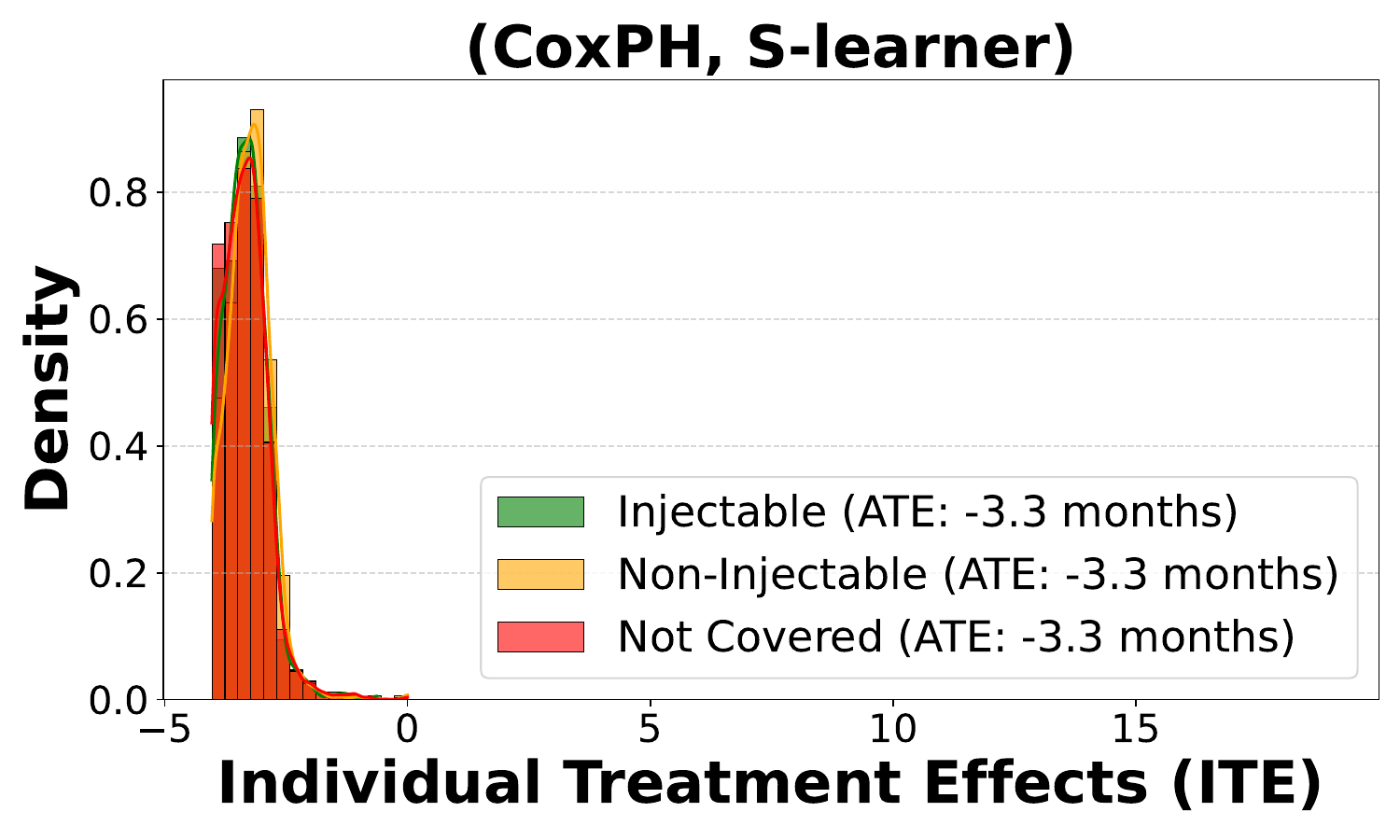}}%
    \qquad
    \subfigure[\footnotesize (CoxPH, Matching-1)]{\label{fig:ite_injectable_coxph_matching_1}%
      \includegraphics[width=.22\linewidth]{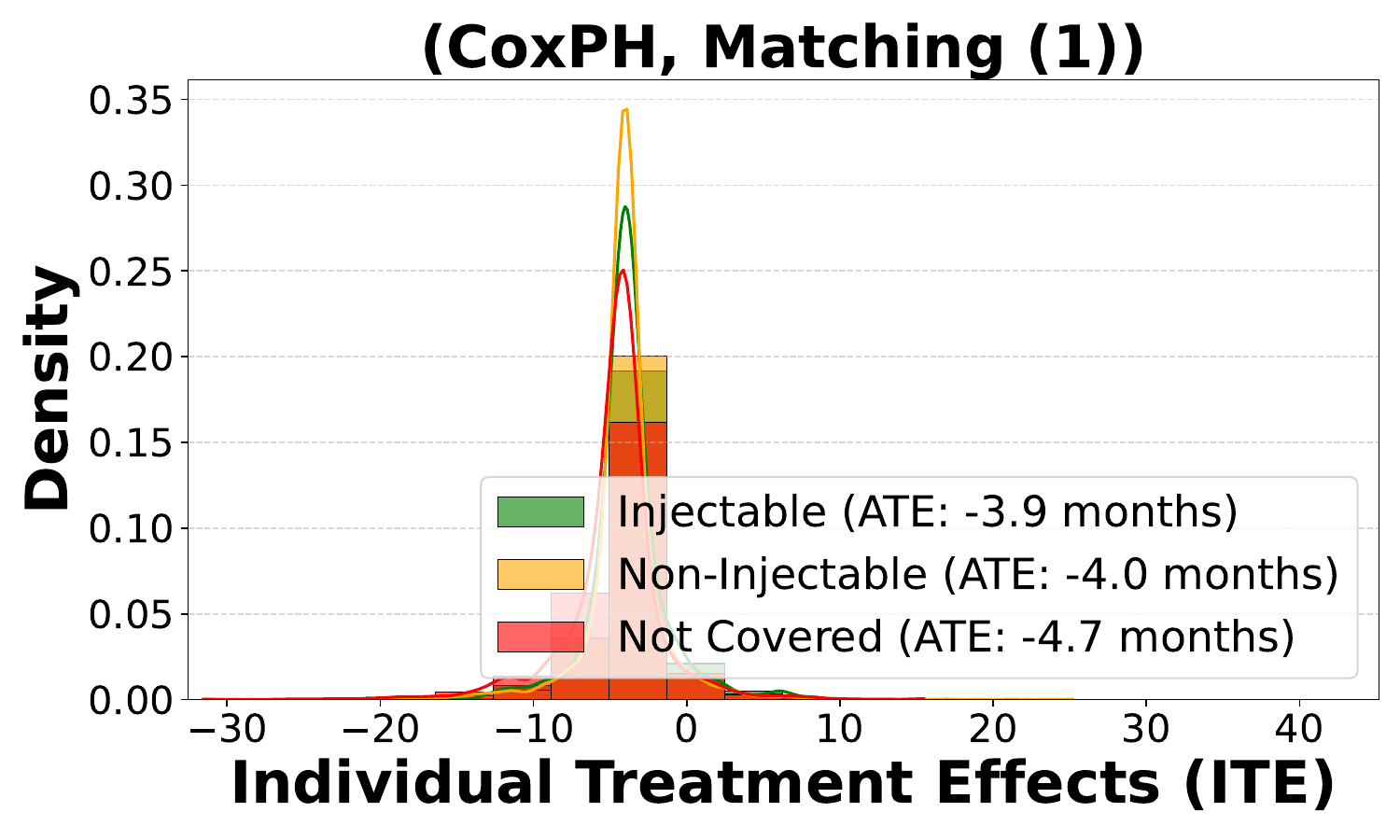}}%
    \qquad
    \subfigure[\footnotesize (CoxPH, Matching-5)]{\label{fig:ite_injectable_coxph_matching_5}%
      \includegraphics[width=.21\linewidth]{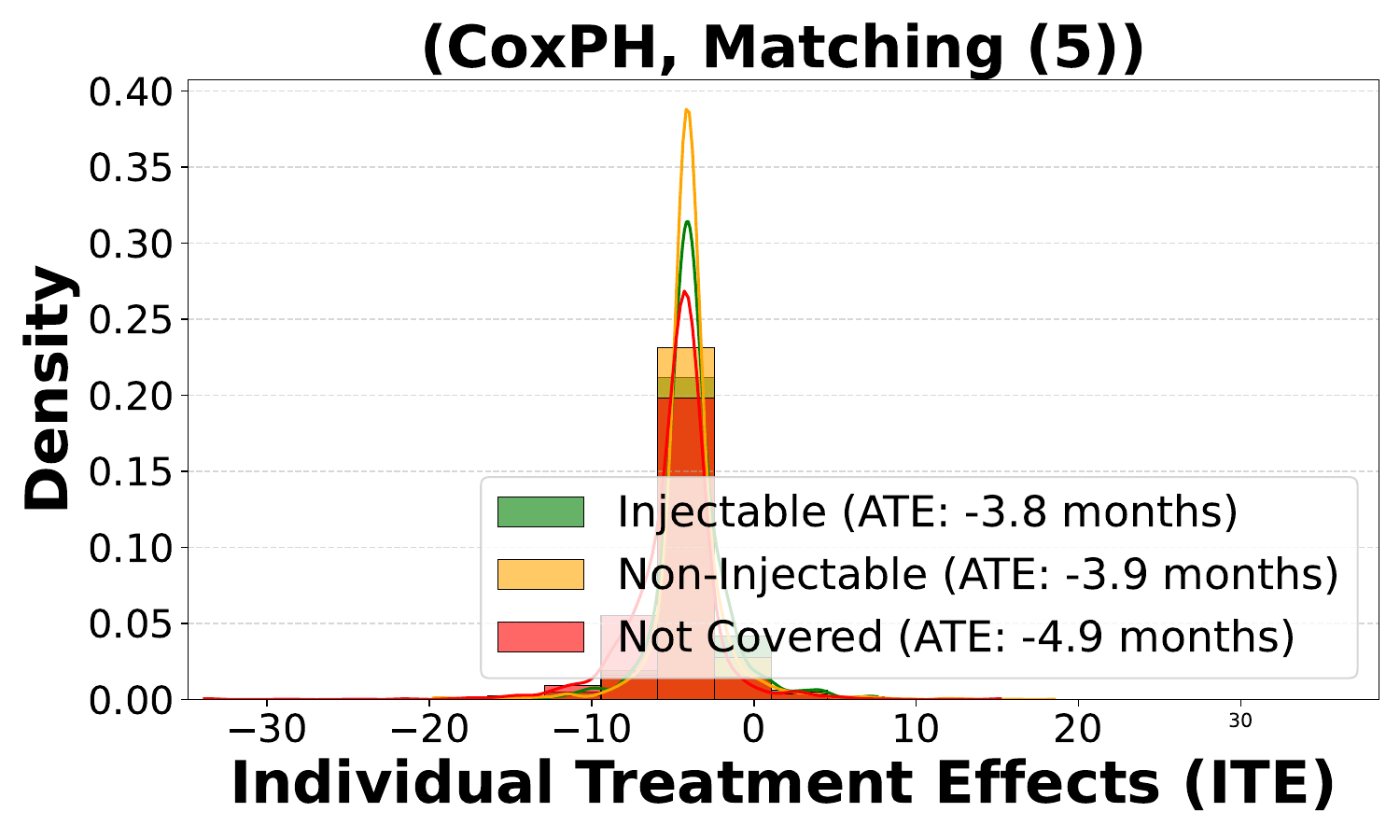}}%
    \\
    \subfigure[\footnotesize (DeepSurv, T-Learner)]{\label{fig:ite_injectable_deepsurv_t_learner}%
      \includegraphics[width=.22\linewidth]{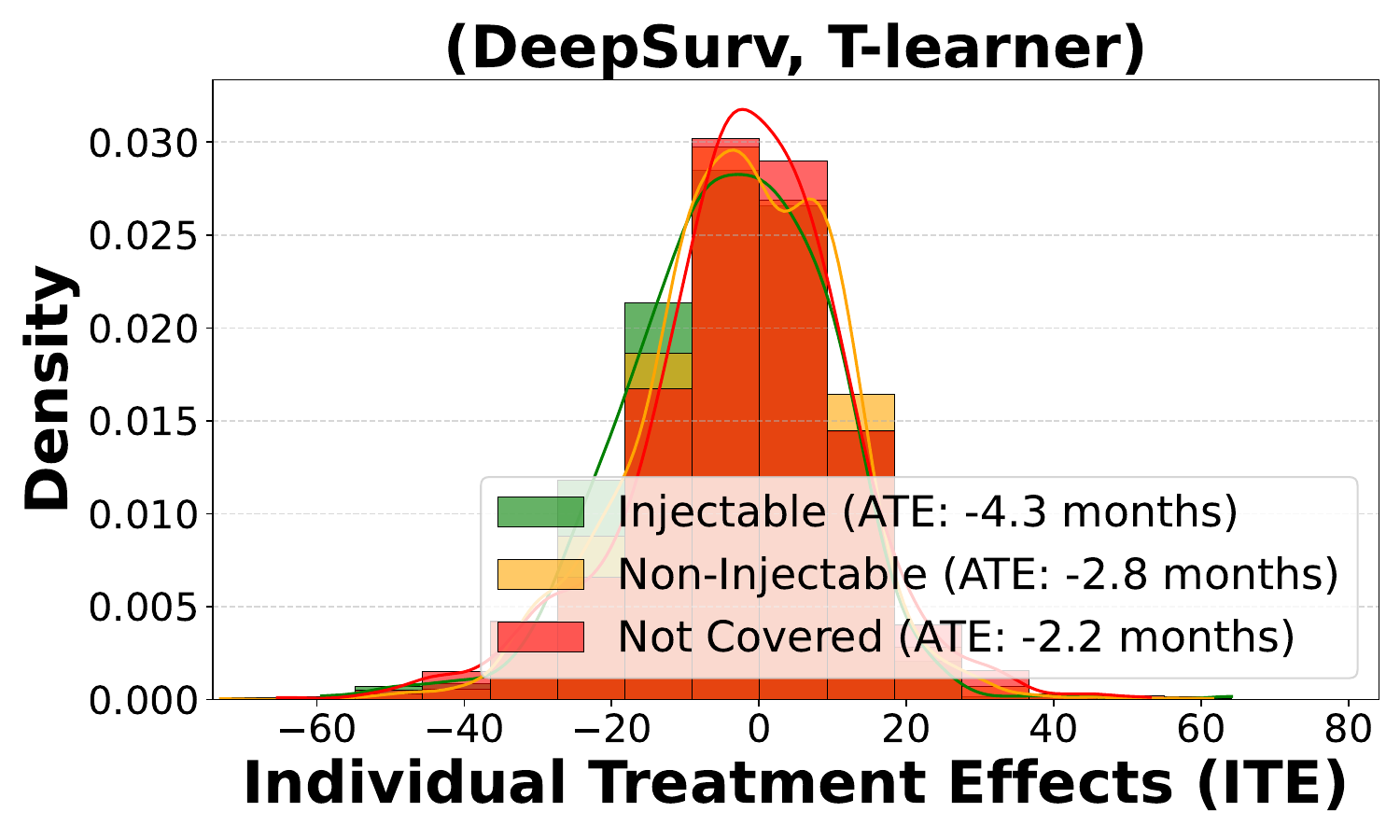}}%
    \qquad
    \subfigure[\footnotesize (DeepSurv, S-Learner)]{\label{fig:ite_injectable_deepsurv_s_learner}%
      \includegraphics[width=.22\linewidth]{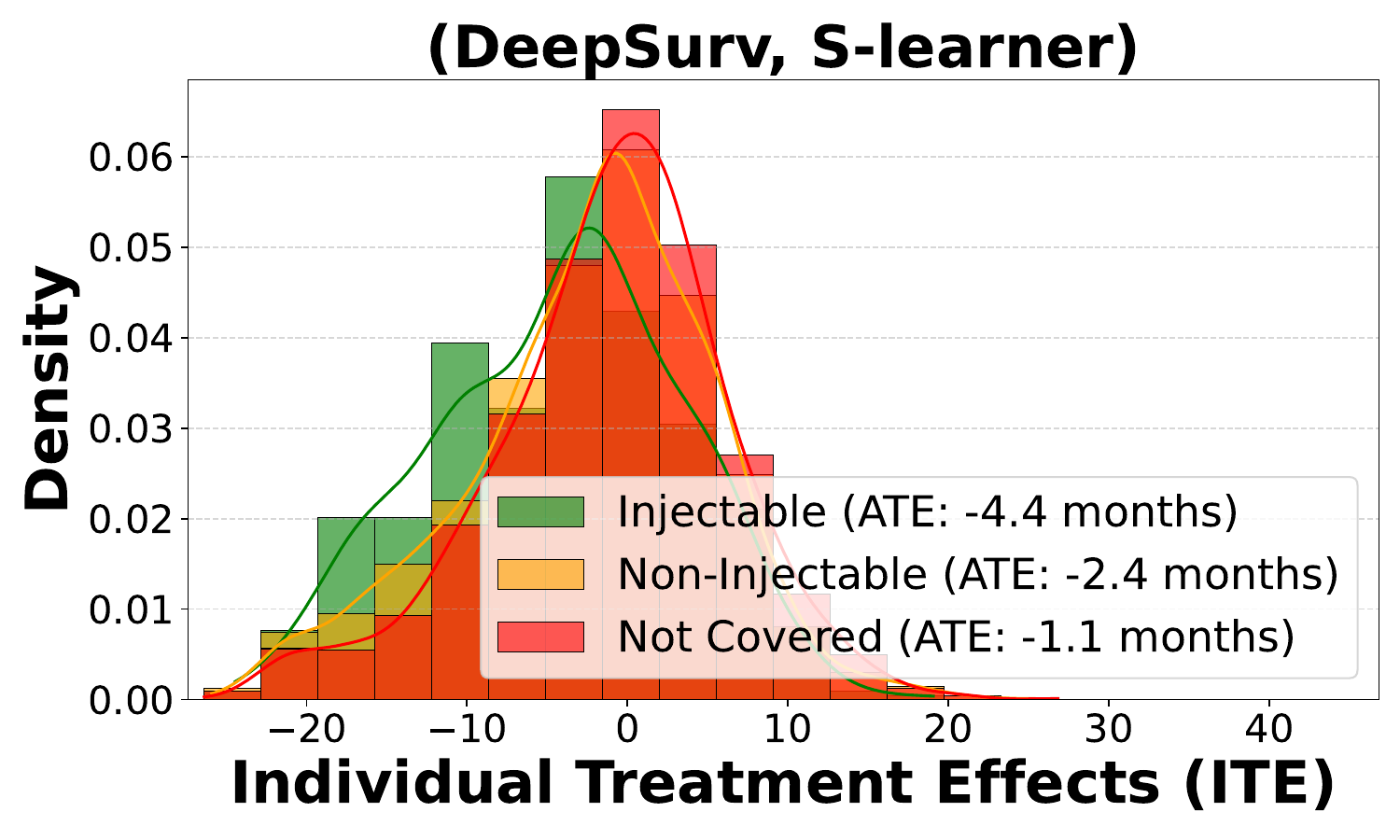}}%
    \qquad
    \subfigure[\footnotesize (DeepSurv, Matching-1)]{\label{fig:ite_injectable_deepsurv_matching_1}%
      \includegraphics[width=.22\linewidth]{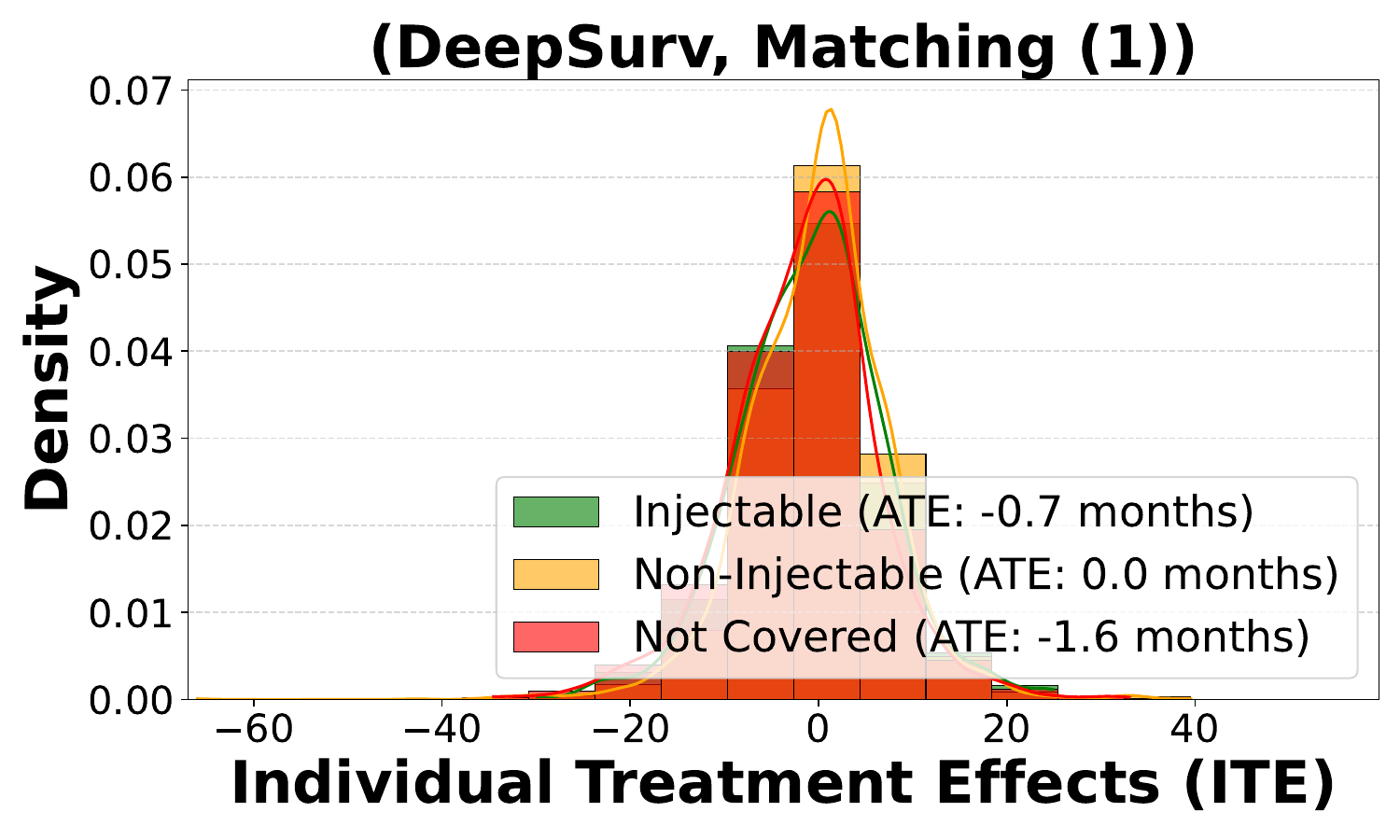}}%
    \qquad
    \subfigure[\footnotesize (DeepSurv, Matching-5)]{\label{fig:ite_injectable_deepsurv_matching_5}%
      \includegraphics[width=.21\linewidth]{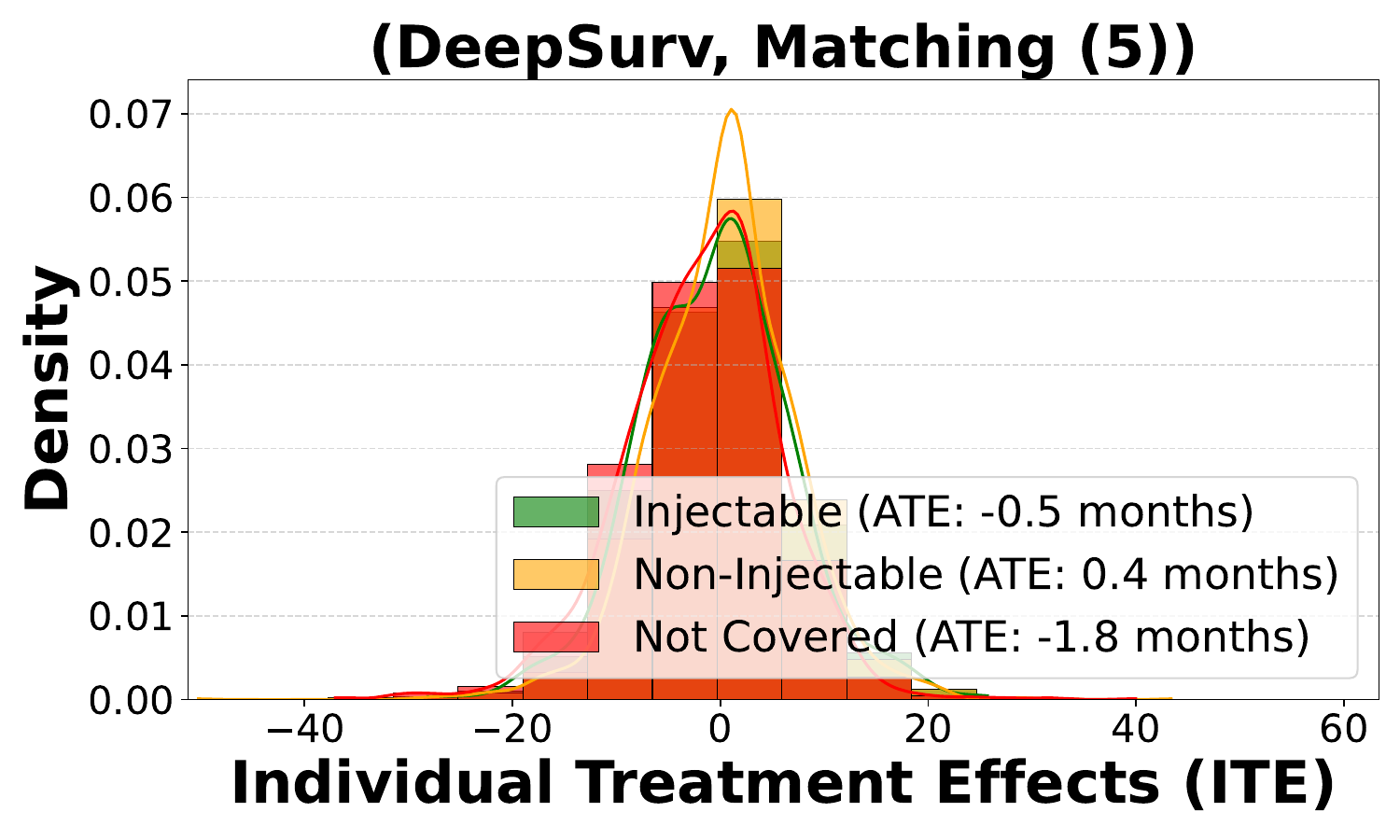}}%
    \\
    \subfigure[\footnotesize (RandomSurvivalForest, T-Learner)]{\label{fig:ite_injectable_rsf_t_learner}%
      \includegraphics[width=.22\linewidth]{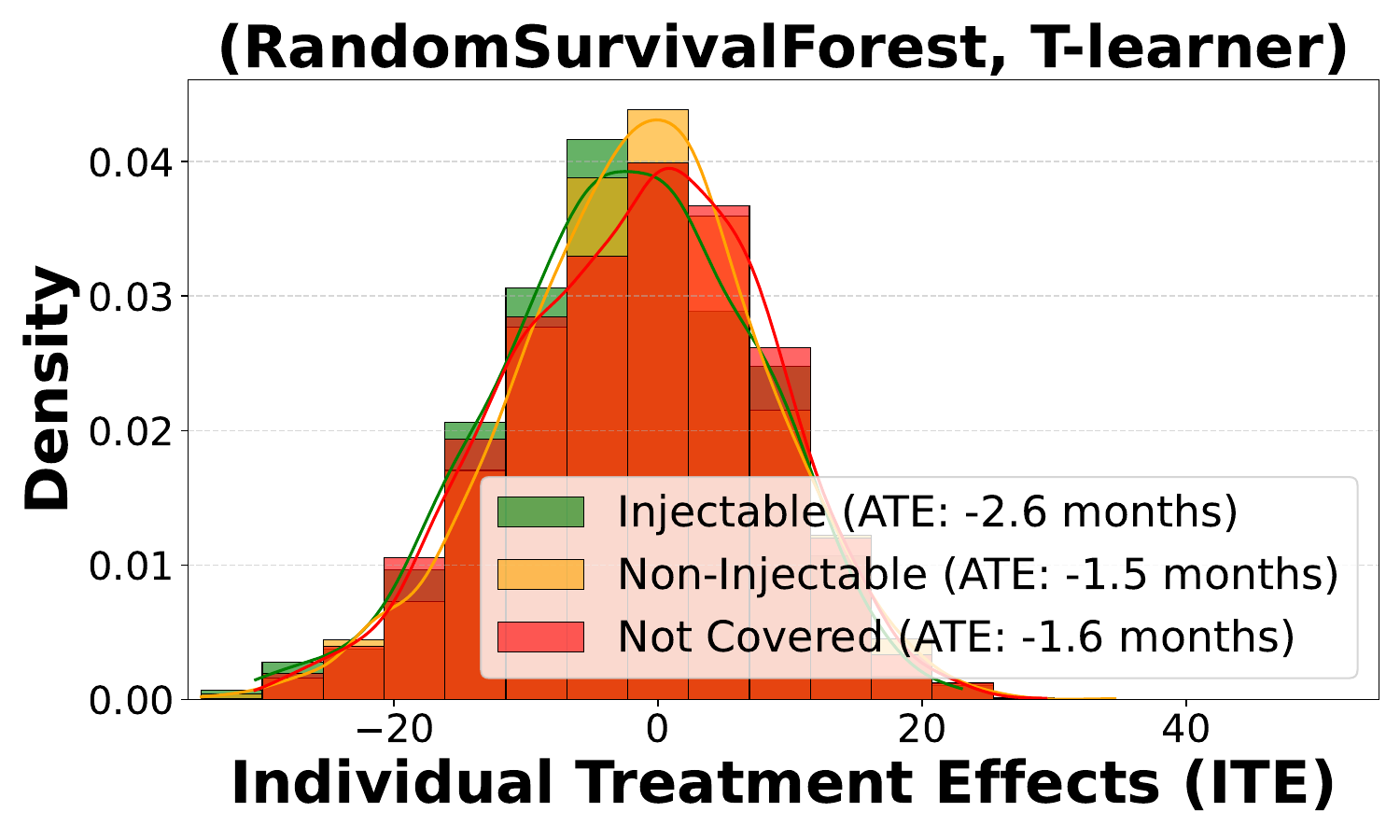}}%
    \qquad
    \subfigure[\footnotesize (RandomSurvivalForest, S-Learner)]{\label{fig:ite_injectable_rsf_s_learner}%
      \includegraphics[width=.22\linewidth]{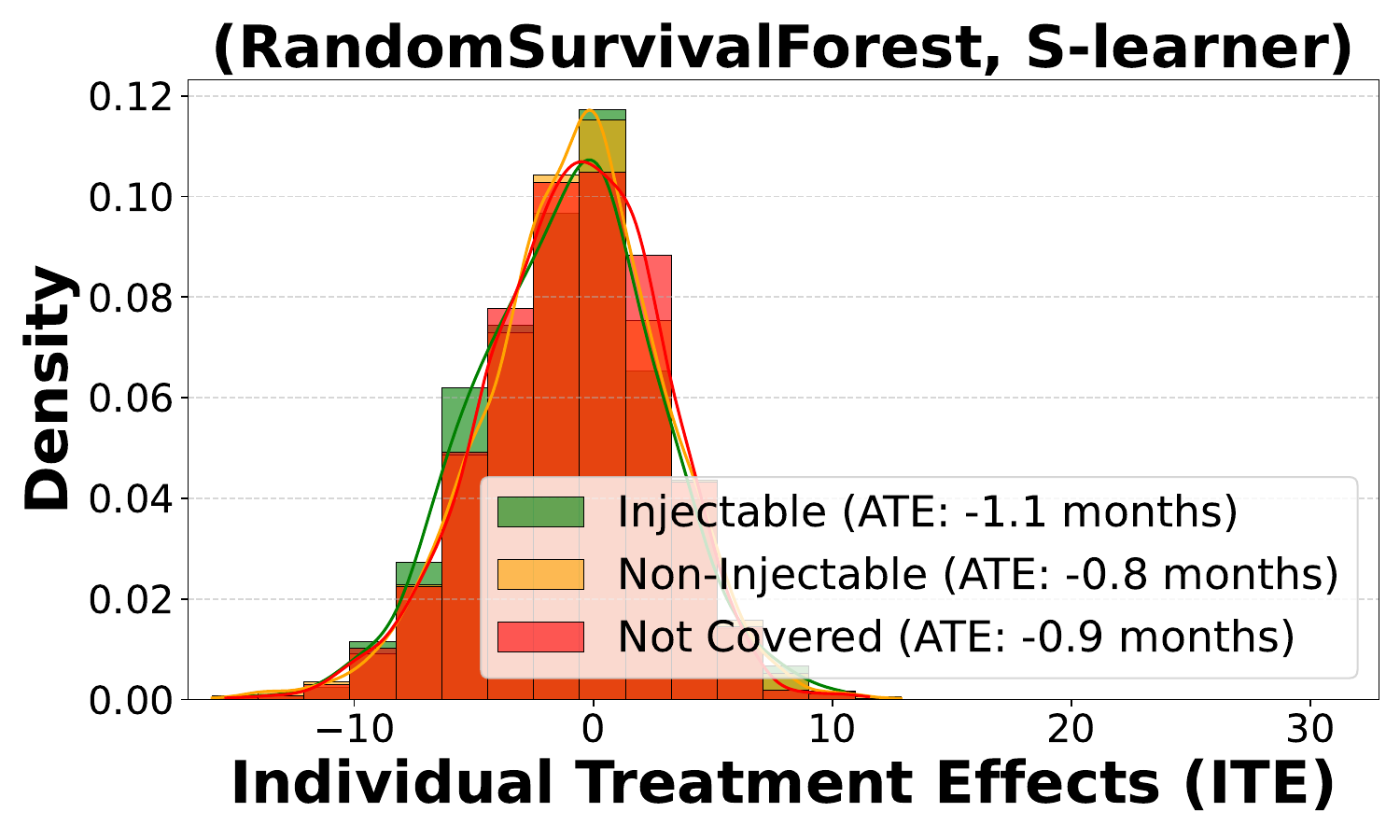}}%
    \qquad
    \subfigure[\footnotesize (RandomSurvivalForest, Matching-1)]{\label{fig:ite_injectable_rsf_matching_1}%
      \includegraphics[width=.22\linewidth]{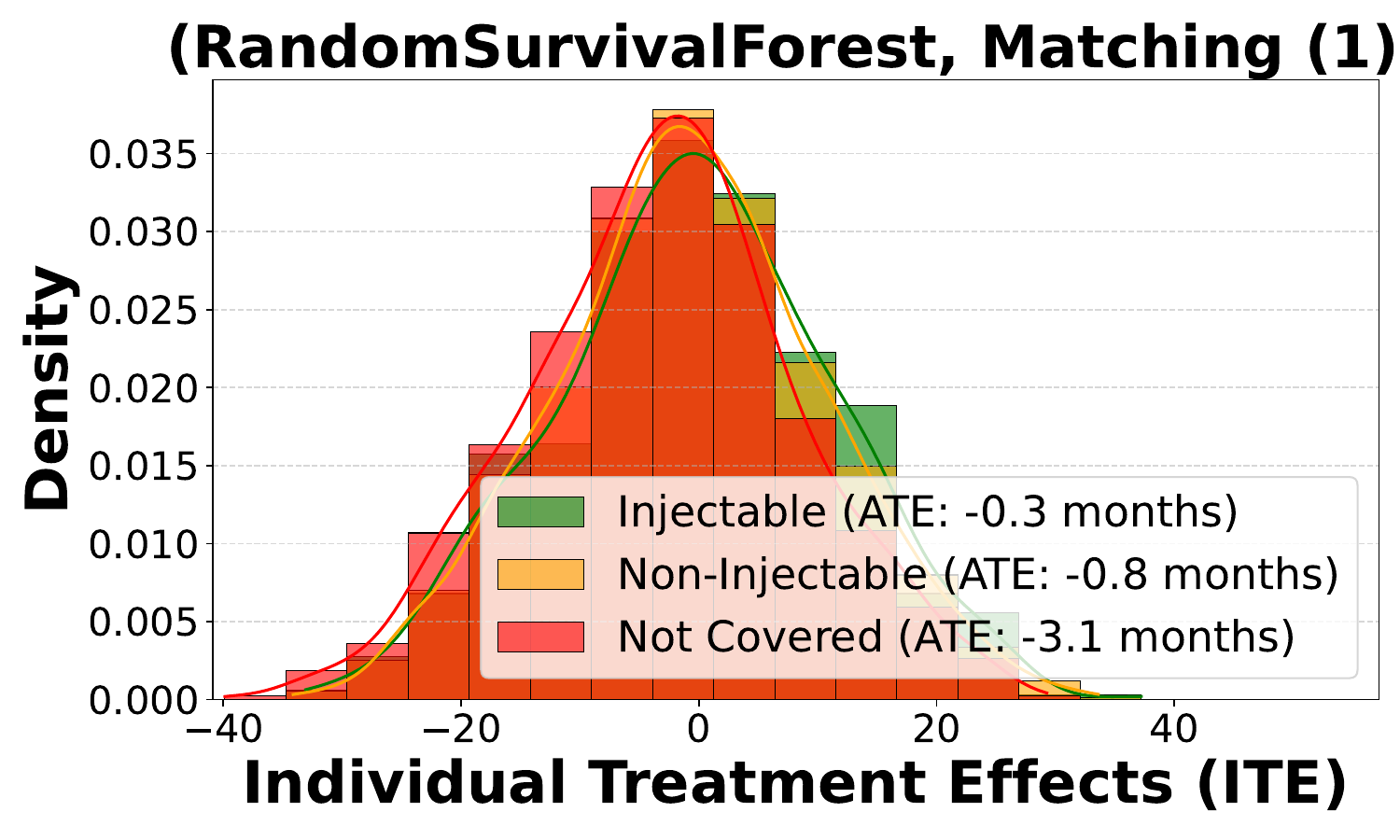}}%
    \qquad
    \subfigure[\footnotesize (RandomSurvivalForest, Matching-5)]{\label{fig:ite_injectable_rsf_matching_5}%
      \includegraphics[width=.21\linewidth]{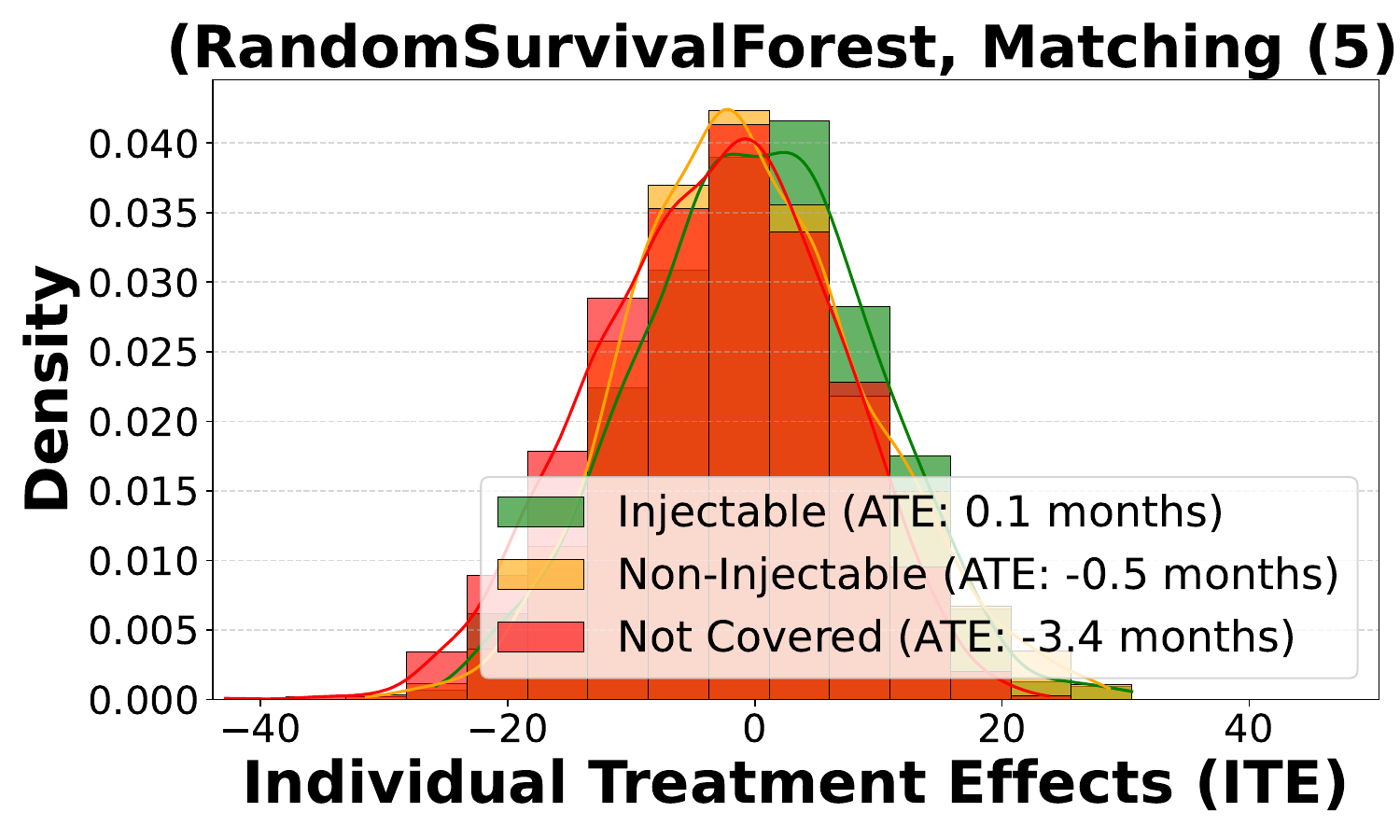}}%
    \\
    \subfigure[\footnotesize Causal Survival Forest]{\label{fig:ite_injectable_causal_survival_forest_all_3}%
      \includegraphics[width=.22\linewidth]{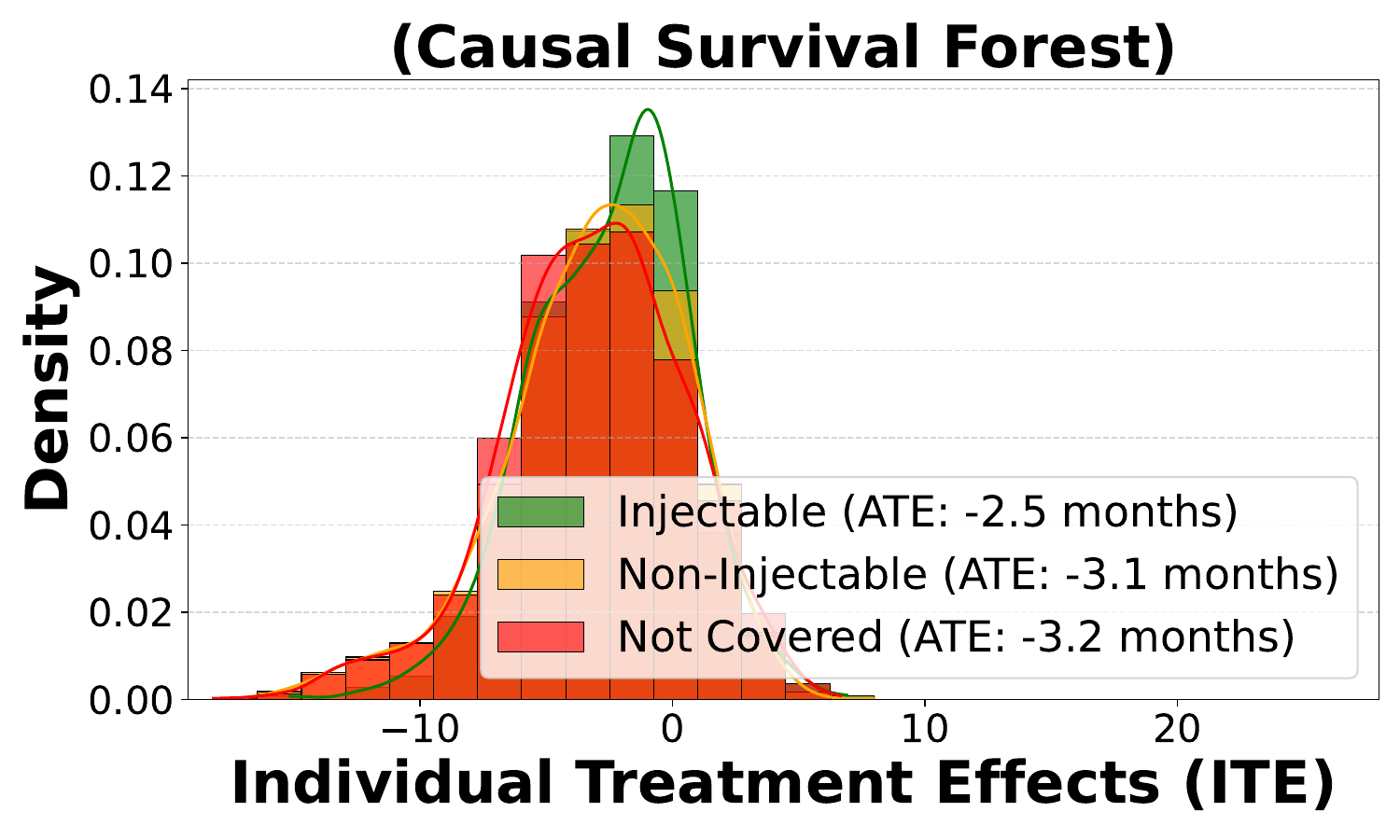}}
  }
\end{figure*}

In this section, we present a detailed analysis of the Estimated Individual Treatment Effects (ITE) distributions across three medication adherence groups—injectable, non-injectable, and not covered—using various survival models and causal methods at the time snapshot \(\tau=3\) months.
Figure~\ref{fig:ite_injectable_all} provides a comprehensive view, with subfigures corresponding to different combinations of survival models (CoxPH, DeepSurv, Random Survival Forest) and causal inference approaches (T-learner, S-learner, and Matching methods with one and five matches) and an additional inherently causal survival model (Causal Survival Forest).

Across the survival models and causal methods, the Average Treatment Effect (ATE) values for injectable medications generally appear slightly more negative than those for non-injectable or not-covered groups.
This trend is most evident in the T-learner and S-learner approaches for CoxPH (Figures~\ref{fig:ite_injectable_coxph_t_learner}, \ref{fig:ite_injectable_coxph_s_learner}) and DeepSurv (Figures~\ref{fig:ite_injectable_deepsurv_t_learner}, \ref{fig:ite_injectable_deepsurv_s_learner}).
However, this distinction diminishes in Matching-based methods (Figures~\ref{fig:ite_injectable_coxph_matching_1}-\ref{fig:ite_injectable_coxph_matching_5} for CoxPH and Figures~\ref{fig:ite_injectable_deepsurv_matching_1}-\ref{fig:ite_injectable_deepsurv_matching_5} for DeepSurv), where the ATEs for injectables are closer to (and sometimes slightly less negative than) those for the non-injectable and not-covered groups.

The Causal Survival Forest (Figure~\ref{fig:ite_injectable_causal_survival_forest_all_3}) demonstrates a distinct pattern compared to other methods, with relatively narrower ITE distributions compared to most other models and less pronounced differences between adherence groups.
This consistency in ATE values across groups indicates minimal heterogeneity in treatment effects based on medication adherence formulation within this model.

Overall, the analysis suggests that while some survival models and causal methods reveal slight differences in ATEs for injectable medications, these differences are not substantial or consistent across methods.
The results align with the main findings reported in Section~\ref{sec:04_results_ite_analysis}, reinforcing the observation that the adherence effect on survival outcomes is largely uniform across injectable, non-injectable, and not-covered groups in the context of our dataset.
This lack of meaningful variability in ITEs suggests that medication formulation may not be a significant factor influencing adherence-related survival outcomes.
Future analyses could expand this approach to include additional adherence categories and explore whether patient subgroups or contextual factors contribute to heterogeneity in treatment effects.
\newpage

\clearpage
\section{Dataset Information}
\label{apd:dataset_information}
\numberwithin{equation}{section}
\numberwithin{figure}{section}
\numberwithin{table}{section}

\subsection{Cohort Information}
\label{apd:cohort_information}
\begin{figure*}[thb]
\floatconts
  {fig:data_overview}
  {\caption{Overview of Schizophrenia Patient Trajectory in Allegheny County. (a) We look at the timing of the first composite outcome \(\in \{\text{\textcolor{orange}{Involuntary Hospitalization}, \textcolor{red}{Jail Stay}, \textbf{Death}}\}\). (b) Non-adherence at each month is calculated by looking at the daily equivalence of prescribed refills available for each month.}}
  {%
    \subfigure[Patient Trajectory]{\label{fig:patient_trajectory}%
      \includegraphics[width=.57\linewidth]{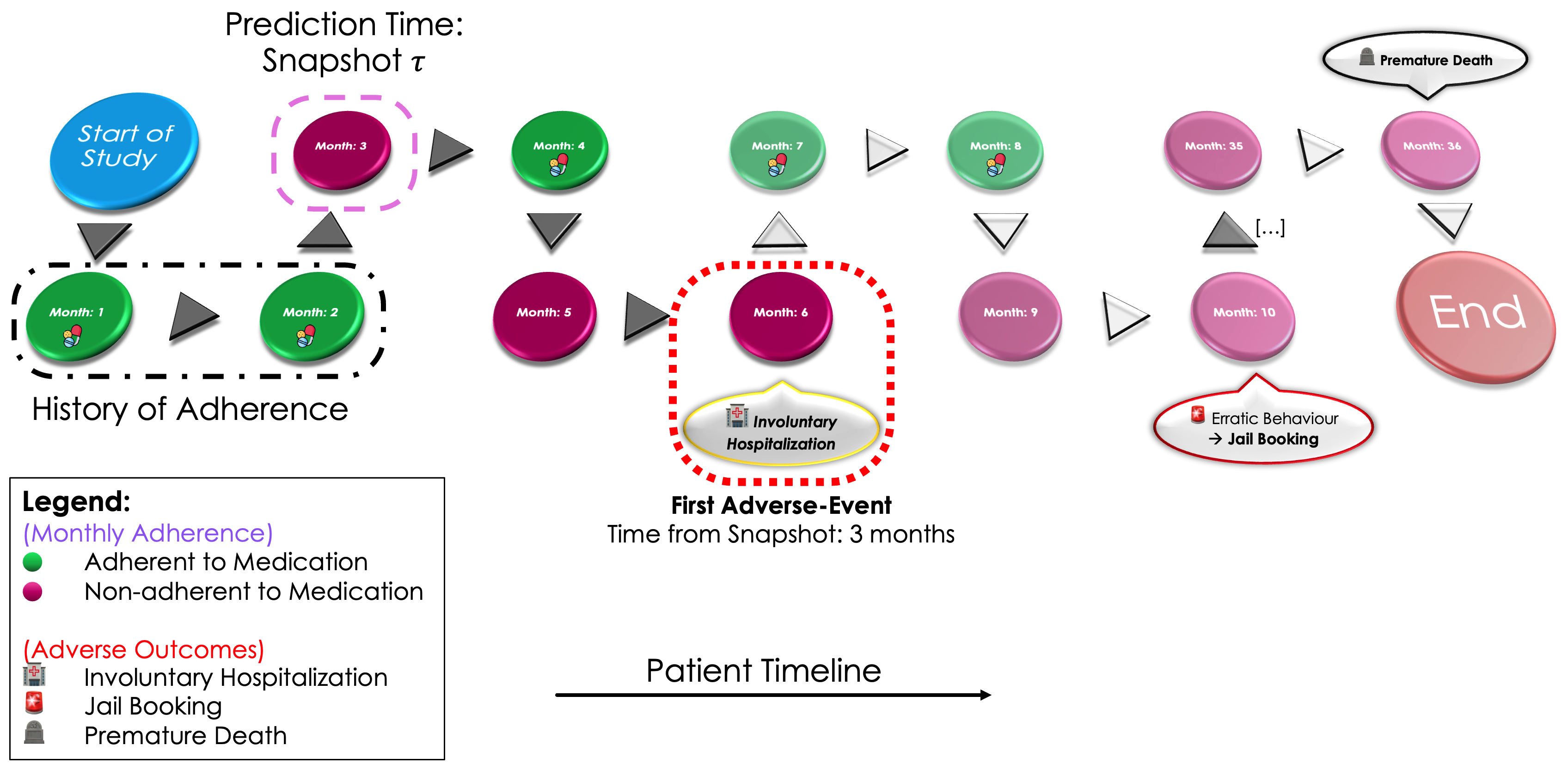}}%
    \hspace{2.5em}
    \subfigure[Gender]{\label{fig:adherence_indicator}%
      \includegraphics[width=.37\linewidth]{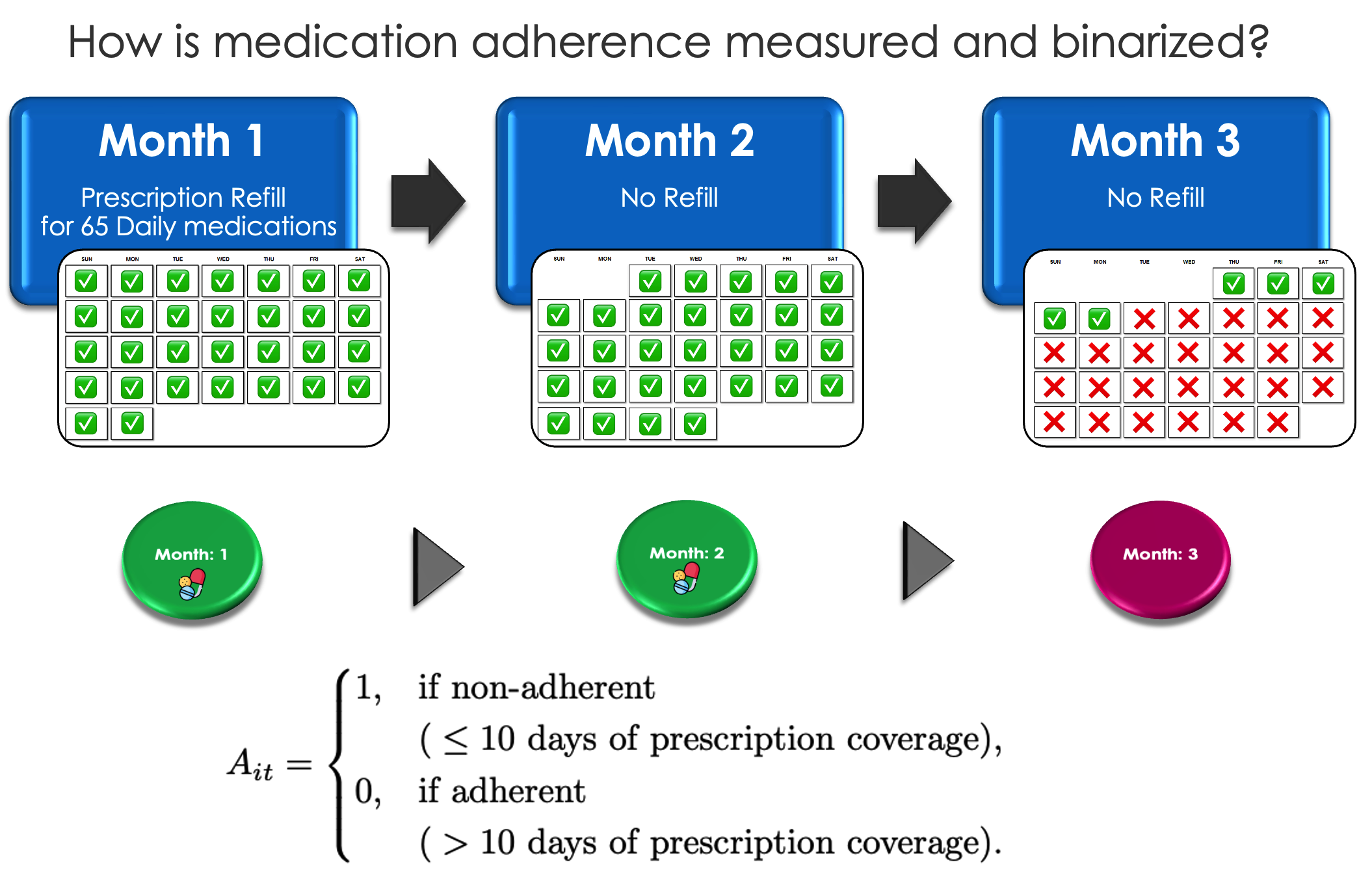}}%
    }
\end{figure*}

To provide further context about the temporal structure and adherence dynamics in our cohort, in Figure~\ref{fig:data_overview}, we include illustrative visualizations of a representative patient timeline and the operational definition of medication adherence. 
Figure~\ref{fig:patient_trajectory} shows an example trajectory of schizophrenia patient in our study, highlighting the structure of monthly adherence (adherent vs. non-adherent), the snapshot time \(\tau\) used for prediction, and the timing of the first observed adverse event, defined as the earliest occurrence of involuntary hospitalization, jail booking, or premature death. 
This figure illustrates how patient histories are segmented, how time-to-event is defined from the prediction snapshot, and the types of adverse outcomes we consider.
Figure~\ref{fig:adherence_indicator} illustrates how monthly medication adherence is binarized. 
Daily prescription coverage is computed based on refill history, and a patient-month is labeled non-adherent if medication availability covers 10 days or fewer. 
(For medications prescribed at different frequencies, such as weekly, adherence is defined analogously based on the number of covered doses within the month).
The choice of this threshold is discussed in Appendix~\ref{apd:cohort_nonadherence}.
These binarized monthly non-adherence indicators serve as time-varying covariates in our dataset.
Together, these visualizations clarify the temporal modeling setup, the construction of key covariates, and how the outcome is defined in our survival analysis framework.

The rest of this section provides a comprehensive overview of the study cohort, detailing its demographic composition, adverse event characteristics, adherence patterns, and prescribing trends for antipsychotic medications. 
This analysis aims to contextualize the characteristics of the patient population and identify key factors that may influence adherence behaviors. 
By examining static demographic covariates, the timing and type of adverse events, non-adherence trends, and medication-specific patterns, this section offers context for subsequent findings and their implications.

\subsubsection{Demographics}
\label{apd:cohort_demographics}

\begin{figure}[htbp]
\floatconts
  {fig:demographics}
  {\caption{Distribution of Static Demographic Covariates Across Patients in the Study Cohort.}}
  {%
    \subfigure[Age at the Beginning of the Study]{\label{fig:demographics_age}%
      \includegraphics[width=.9\linewidth]{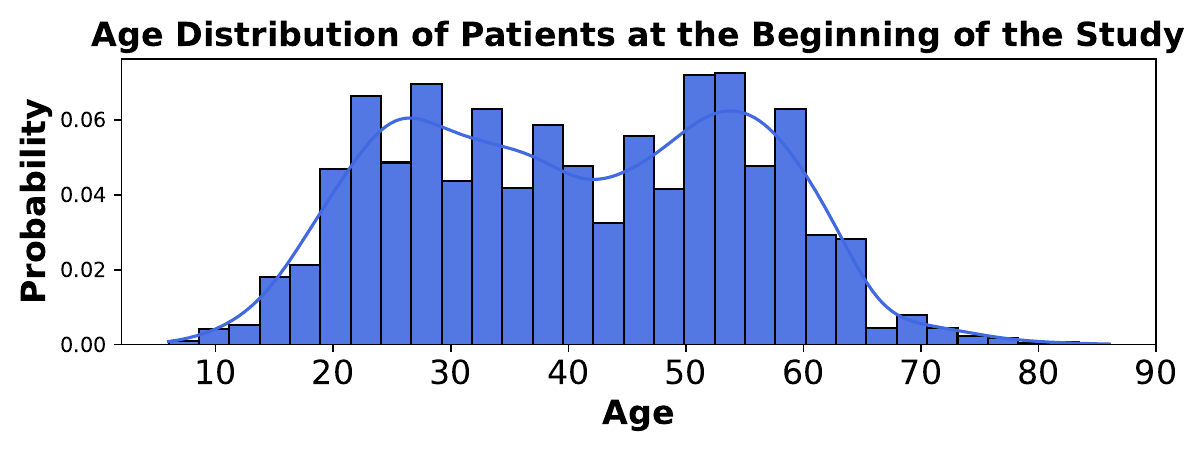}}%
    \\
    \subfigure[Gender]{\label{fig:dempgraphics_gender}%
      \includegraphics[width=.9\linewidth]{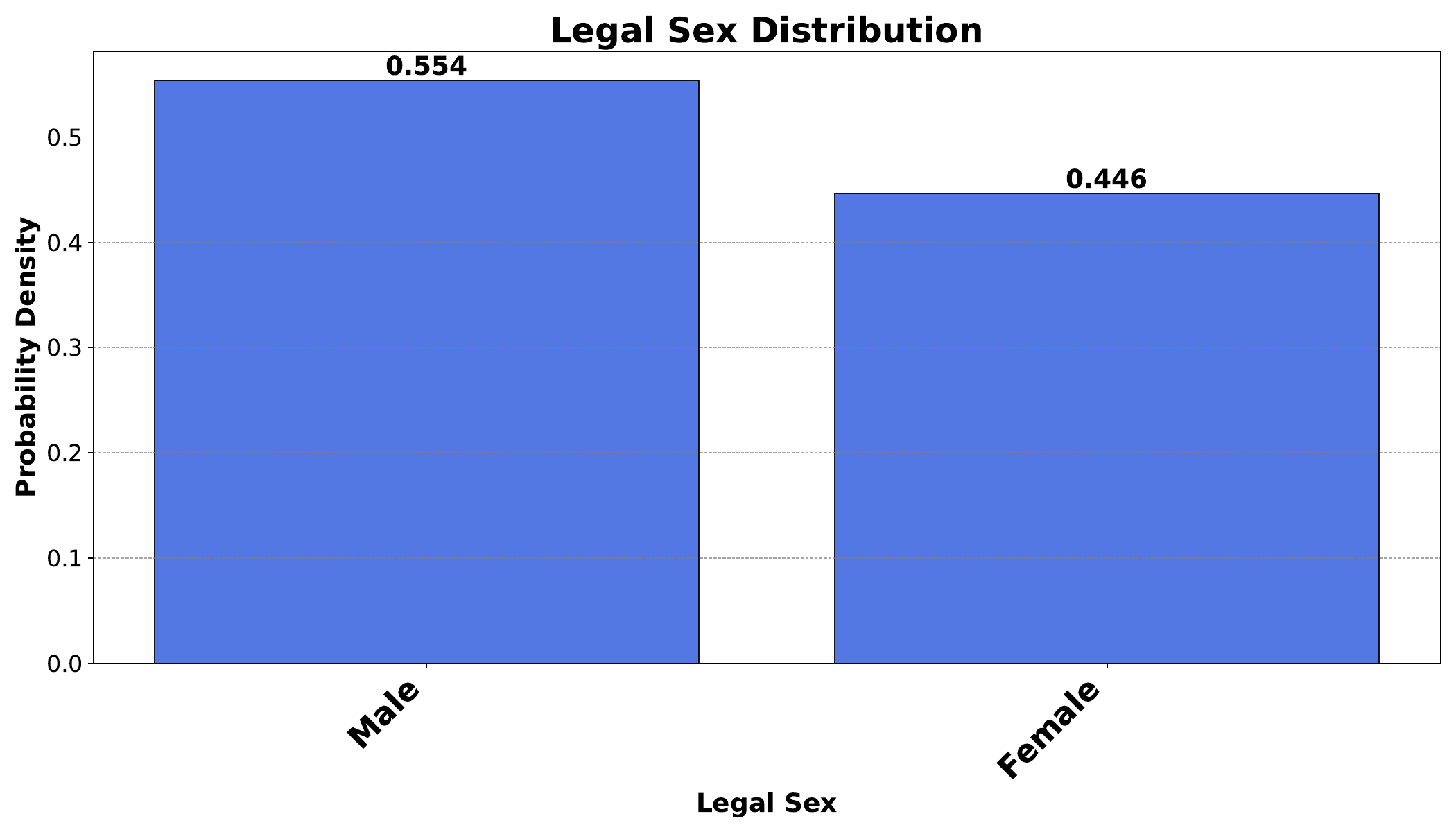}}%
    \\
    \subfigure[Race]{\label{fig:demographics_race}%
      \includegraphics[width=.9\linewidth]{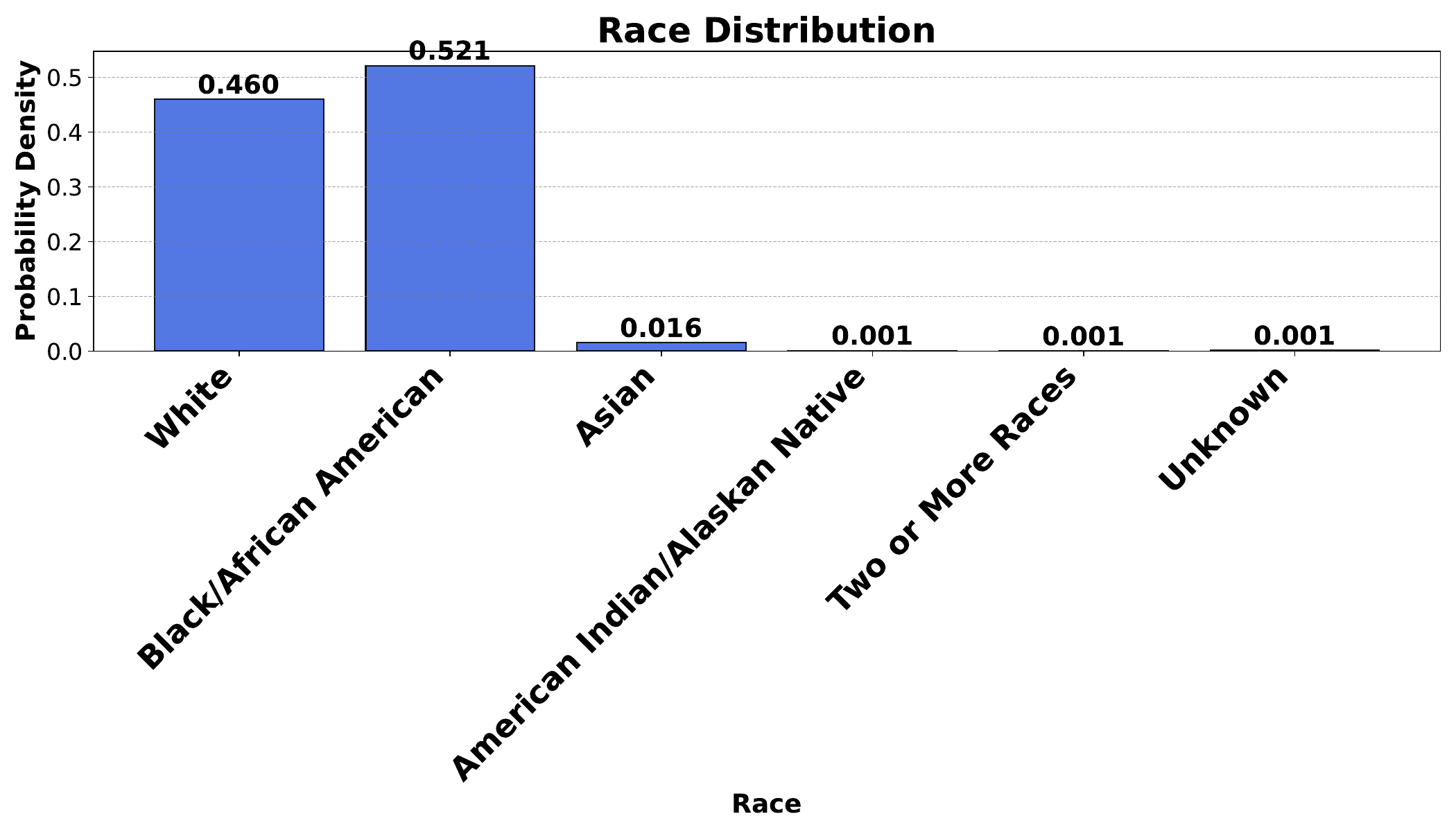}}%
    \\
    \subfigure[Education Level]{\label{fig:demographics_education}%
      \includegraphics[width=.9\linewidth]{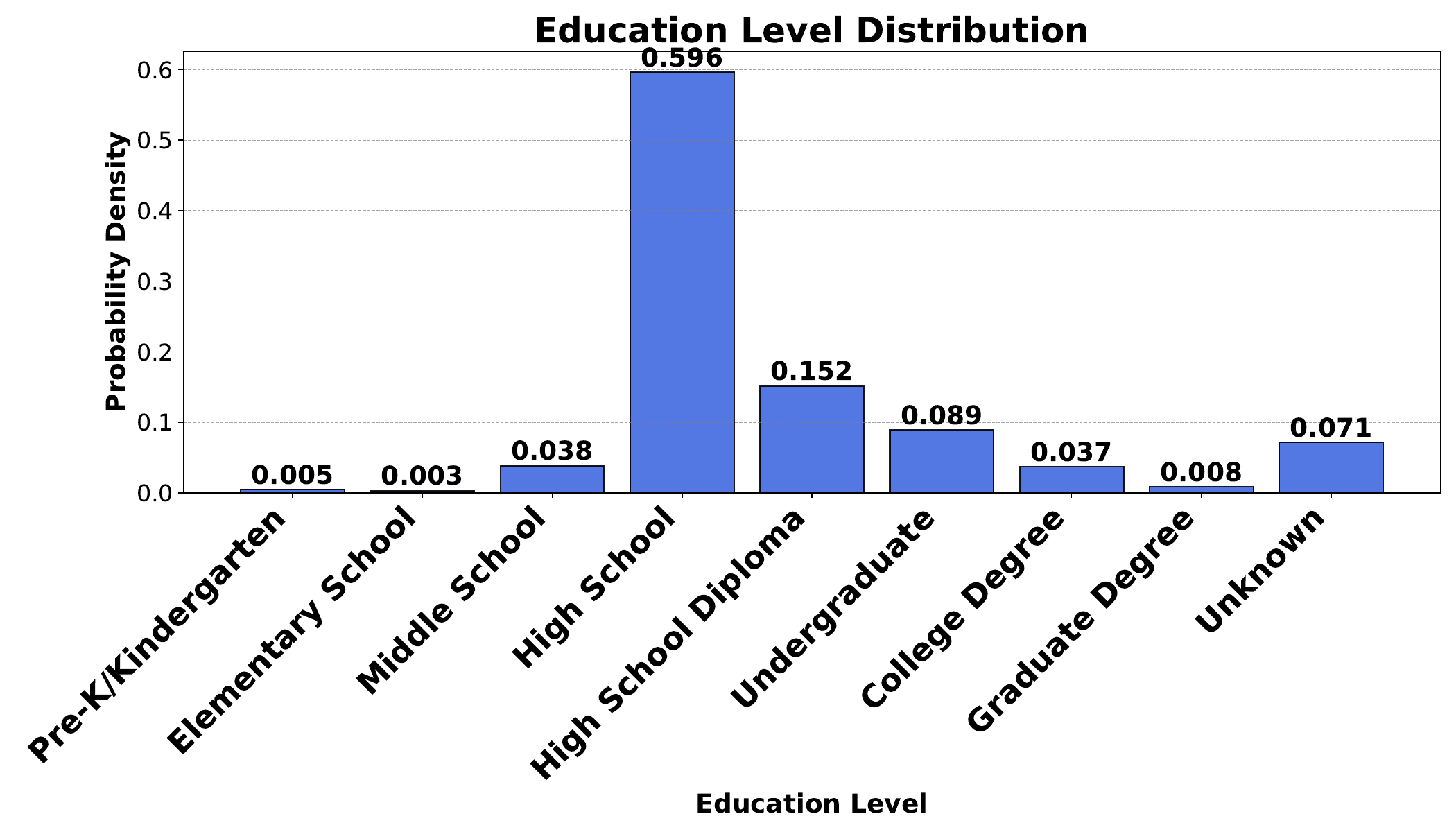}}%
  }
\end{figure}

The demographic characteristics of patients included in the study cohort are summarized in Figure~\ref{fig:demographics}.
 
Figure~\ref{fig:demographics_age} presents the age distribution of patients at the beginning of the study.
The cohort exhibits a broad age range, with the majority of patients aged between 30 and 60 years, and a peak density near 50 years.
This suggests that the study primarily captures middle-aged adults, though younger and older individuals are also represented.
 
The gender distribution is shown in Figure~\ref{fig:dempgraphics_gender}, with males comprising 55.4\% of the cohort and females making up 44.6\%.
This slight overrepresentation of males may reflect broader population trends or sampling biases in the study cohort.
 
Figure~\ref{fig:demographics_race} depicts the racial distribution of the cohort.
Black/African American patients form the largest racial group at 52.1\%, followed by White patients at 46.0\%.
Other racial groups, including Asian, American Indian/Alaskan Native, and multiracial individuals, are underrepresented, each contributing less than 2\% to the overall cohort.
 
The education level distribution, displayed in Figure~\ref{fig:demographics_education}, highlights that the majority of patients (59.6\%) have their education level at high school level.
A substantial proportion have high school diploma (15.2\%), while 8.9\% have enrolled in college.
The remaining patients have lower levels of formal education or their education status is unknown.

\subsubsection{First Adverse Event Type and Time}
\label{apd:cohort_first_event}

The timing and type of the first composite adverse event for patients in the study cohort are depicted in Figure~\ref{fig:first_event_time}. 
In the cohort of our study, the average time to a composite adverse event is 18.1 months. 
The average censoring time is 60.7 months.

\begin{figure*}[!htbp]
\floatconts
  {fig:first_event_time}
  {\caption{ (a)-(b) Distribution of Time of First Composite Adverse Event (in Months) across Patients Stacked by Each Adverse Event Type. 
  (a) includes patients with no adverse event (Right-Censored). 
  (b) presents the histogram of times only for patients with an adverse event recorded (Excluding Right-Censored Patients).
  (c)-(d) Distribution of Event Types within each Composite Adverse Event.}}
  {%
    \subfigure[All Patients]{\label{fig:first_event_time_with_censoring}%
      \includegraphics[width=0.9\linewidth]{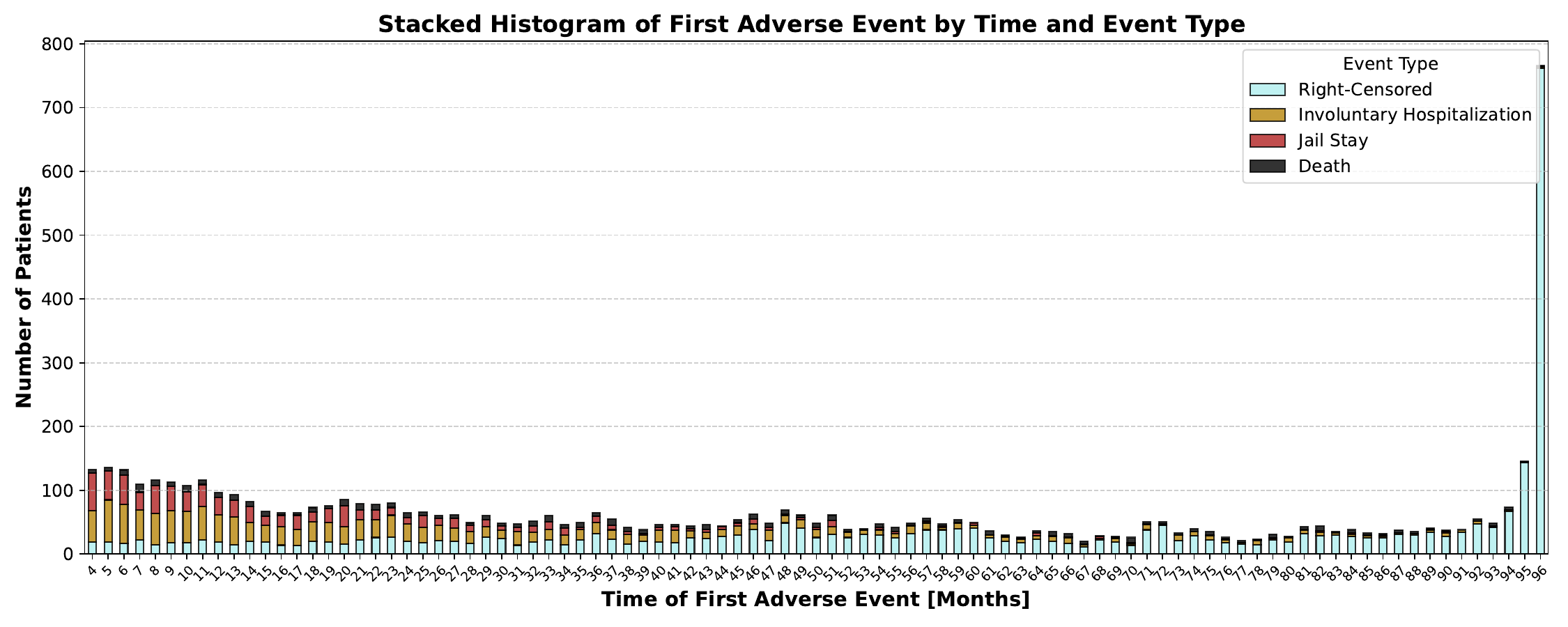}}%
    \\
    \subfigure[Patients with an Adverse Event]{\label{fig:first_event_time_without_censored_data}%
      \includegraphics[width=0.9\linewidth]{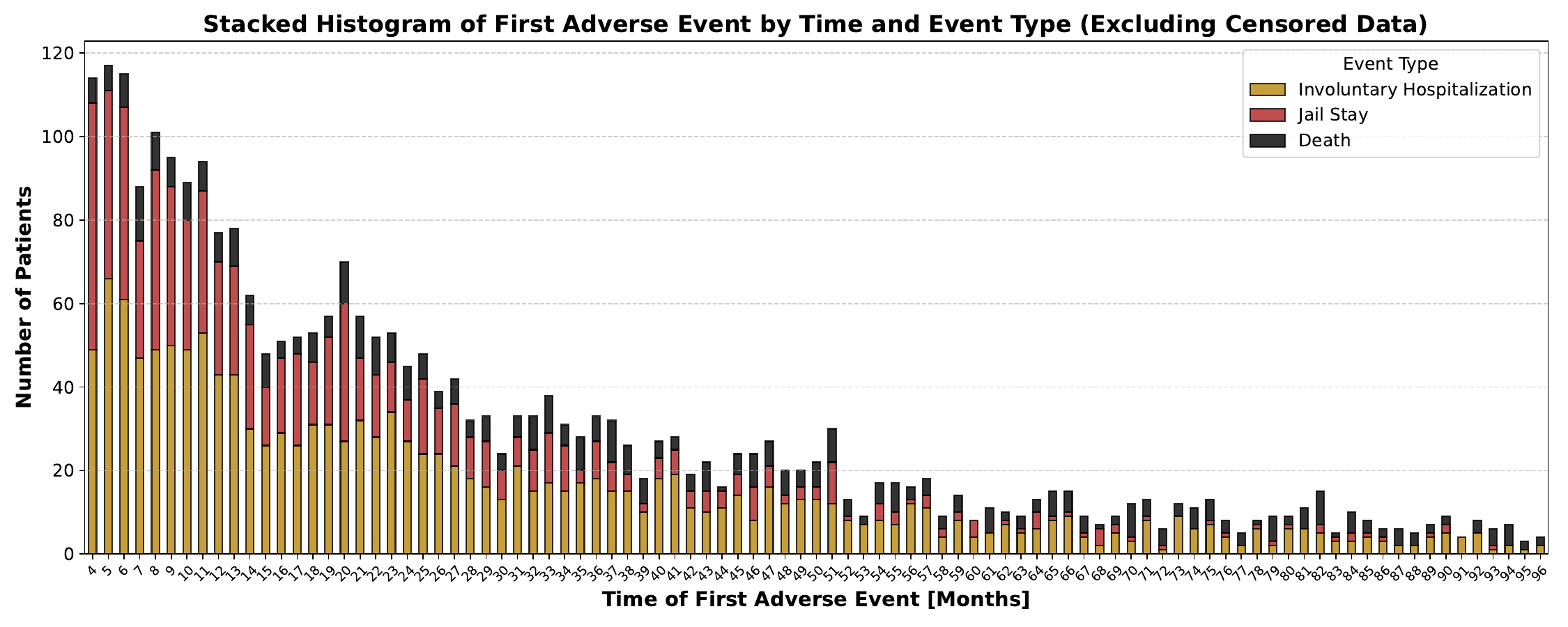}}%
    \\
    \subfigure[Event Type Distribution - All Patients]{\label{fig:event_type_with_censoring}%
      \includegraphics[width=0.4\linewidth]{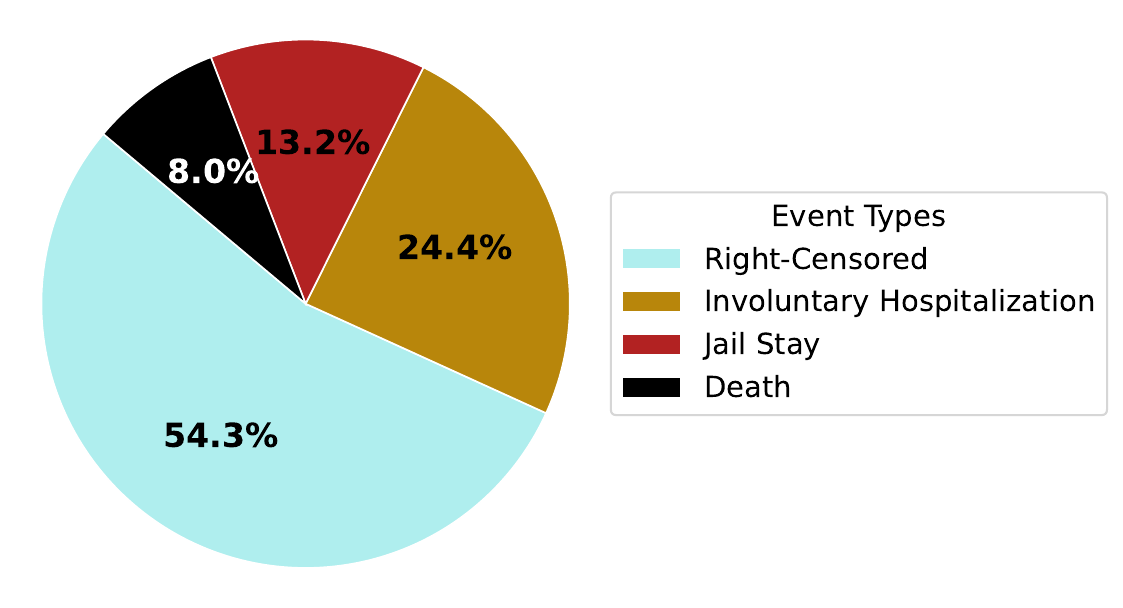}}%
    \qquad
    \subfigure[Event Type Distribution - Patients with an Adverse Event]{\label{fig:event_type_without_censored_data}%
      \includegraphics[width=0.4\linewidth]{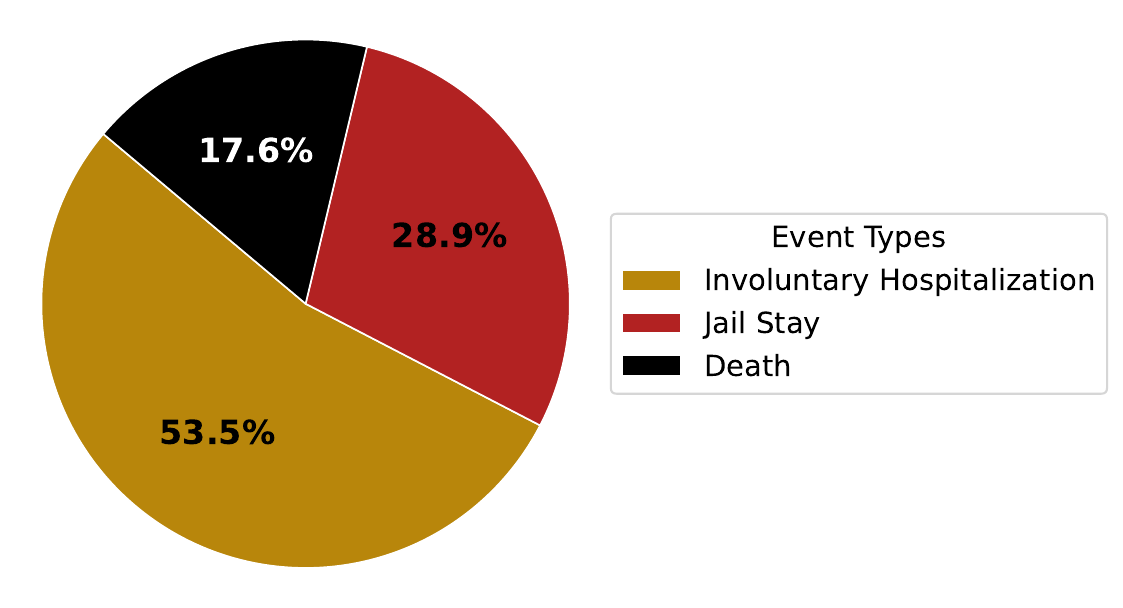}}%
  }
\end{figure*}

Figure~\ref{fig:first_event_time_with_censoring} illustrates the distribution of the time to the first adverse event in months for all patients, including those who were right-censored. 
The histogram reveals that a substantial number of patients experience their first adverse event within the first 12 months of the study. 
The peak in adverse events at later months, specifically near the 96-month mark, coincides with patients who are censored at the end of the follow-up period. 

Figure~\ref{fig:first_event_time_without_censored_data} presents a similar distribution but excludes right-censored patients. 
The histogram demonstrates that the majority of adverse events occur within the first two years of the study, with a gradual decline in frequency over time. 
This decline reflects the decreasing number of at-risk patients as the study progresses. 

The distribution of event types for the first composite adverse event is summarized in Figures~\ref{fig:event_type_with_censoring} and \ref{fig:event_type_without_censored_data}. 
Figure~\ref{fig:event_type_with_censoring} includes all patients and highlights that over half of the cohort (54.3\%) were right-censored, with involuntary hospitalization comprising 24.4\% of events, followed by jail stays (13.2\%) and deaths (8.0\%). 

For patients who experienced an adverse event, Figure~\ref{fig:event_type_without_censored_data} shows that involuntary hospitalizations account for the majority (53.5\%), followed by jail stays (28.9\%) and deaths (17.6\%). 
The relative proportions of event types underscore the higher prevalence of involuntary hospitalization compared to other adverse events within the cohort of our study. 

\subsubsection{Non-Adherence across Patients}
\label{apd:cohort_nonadherence}

Figure~\ref{fig:non_adherence_level} presents the distribution of average continuous non-adherence across patients in the cohort, highlighting differences in adherence patterns based on medication administration.

\begin{figure}[htbp]
\floatconts
  {fig:non_adherence_level}
  {\caption{Distribution of Average Continuous Non-Adherence across Patients. (a) All Patients, (b) Patients prescribed with Injectable Medications (at least once in their trajectory), (c) Patients prescribed with Non-Injectable Medications. The x-axis shows the number of days in a month that a patient goes non-adherent on average. Non-coverage of less than 10 days in month is considered \textcolor{blue}{Adherent}, and for more than 10 days is considered \textcolor{red}{Non-Adherent} in our setup.}}
  {%
    \subfigure[All Patients]{\label{fig:non_adherence_all}%
      \includegraphics[width=\linewidth]{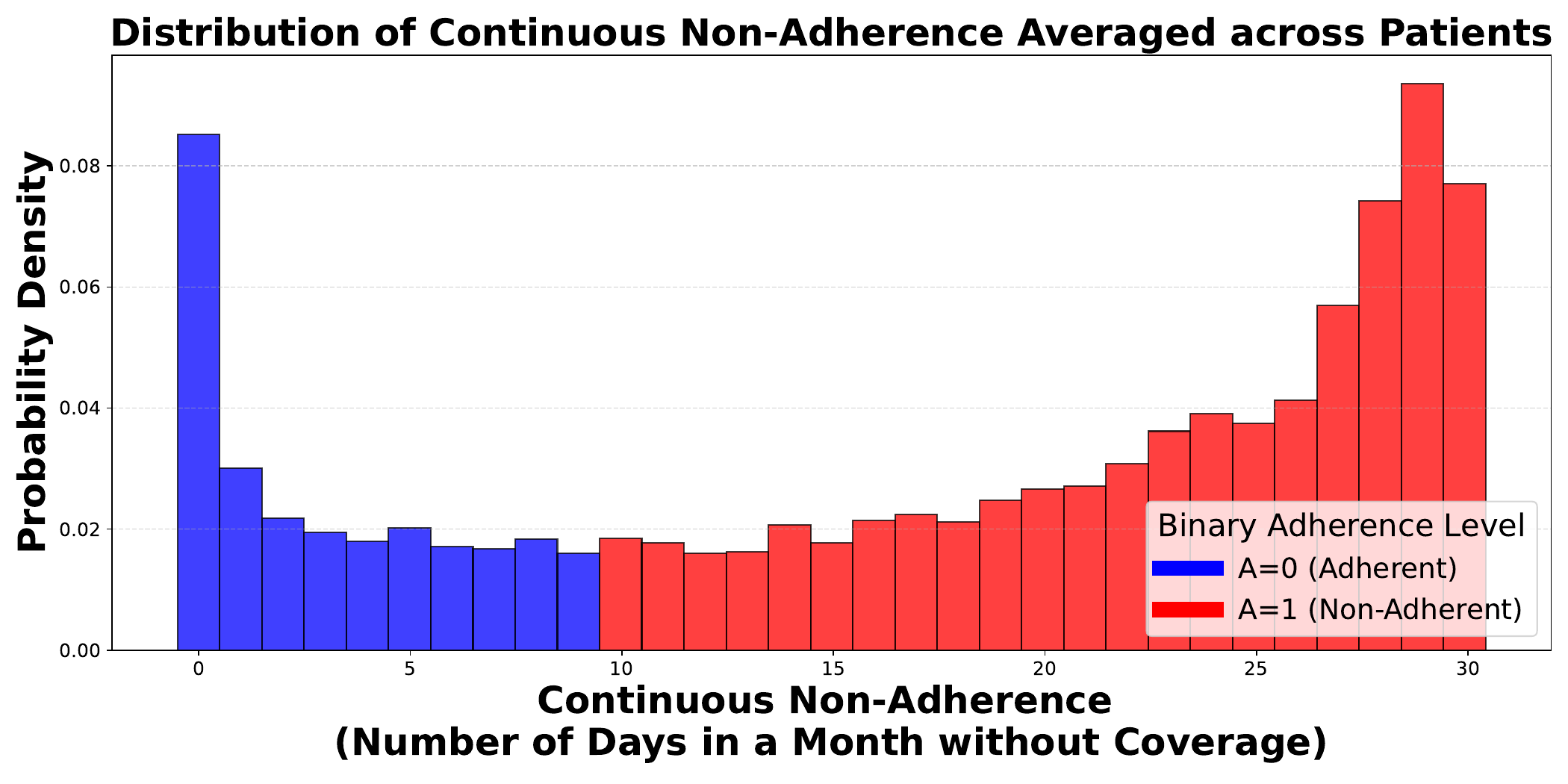}}%
    \\
    \subfigure[Patients with Injectable Medication]{\label{fig:non_adherence_injectable}%
      \includegraphics[width=\linewidth]{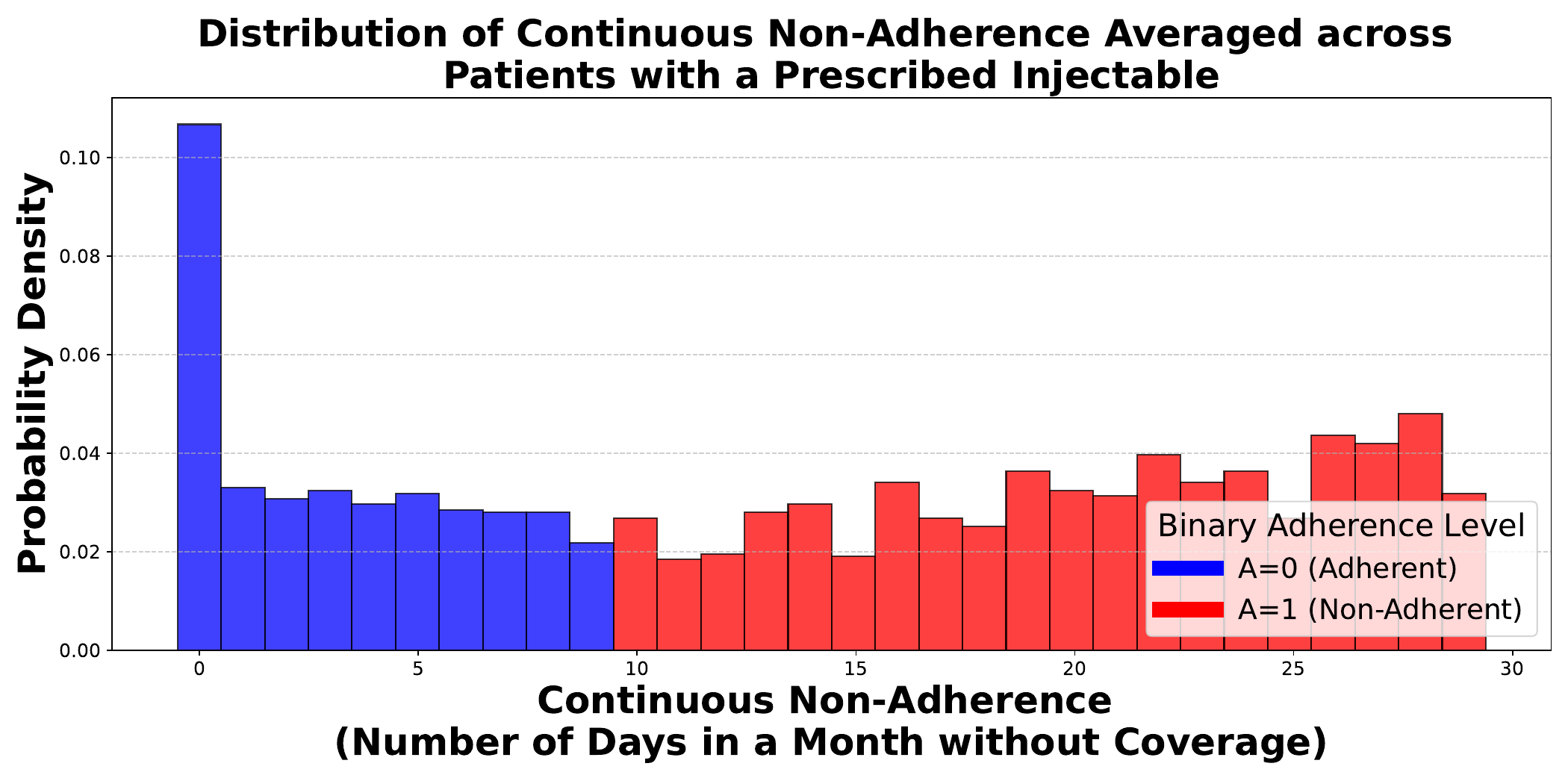}}%
    \\
    \subfigure[Patients with Non-Injectable Medication]{\label{fig:non_adherence_non_injectable}%
      \includegraphics[width=\linewidth]{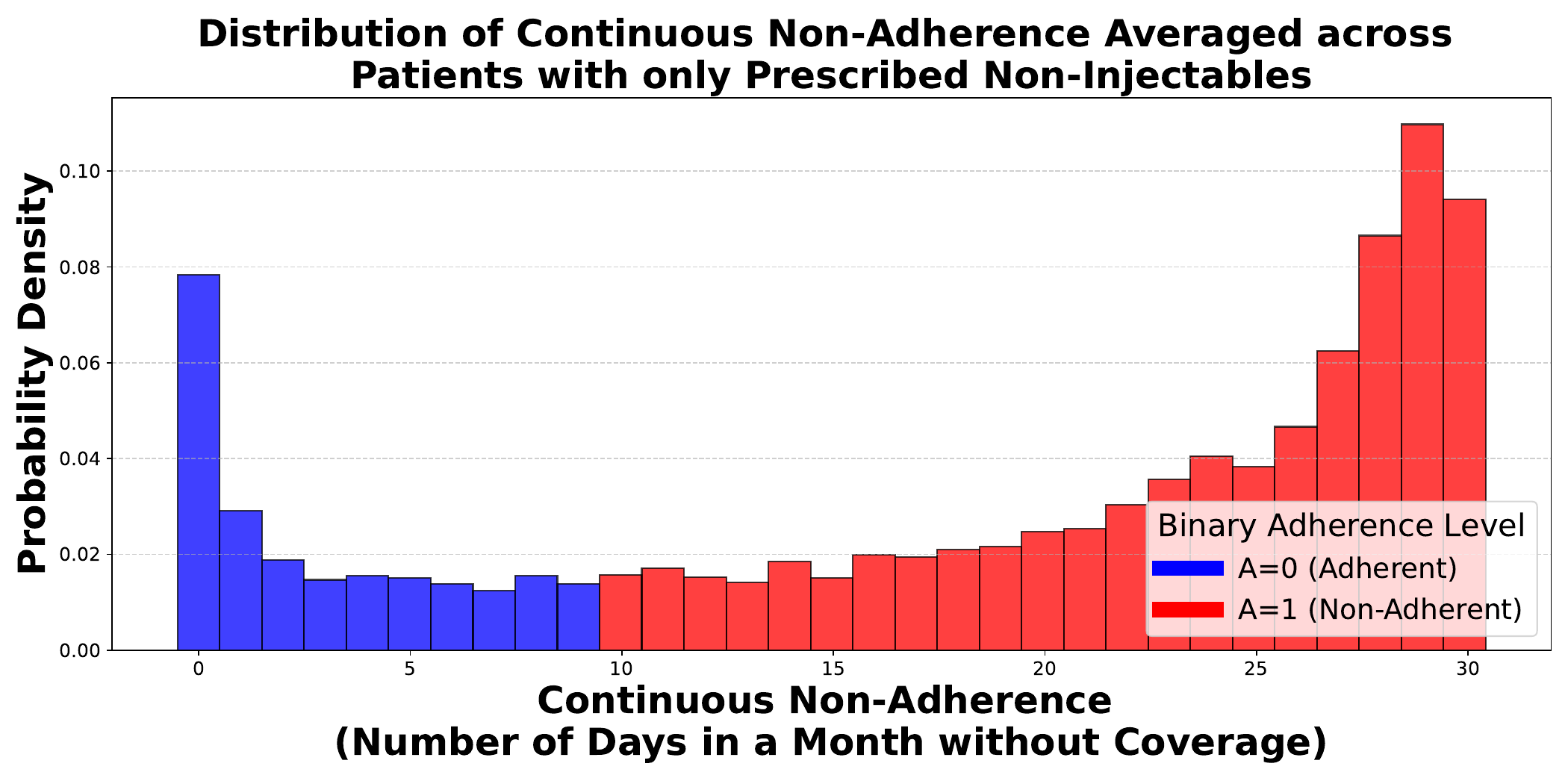}}%
  }
\end{figure}

In Figure~\ref{fig:non_adherence_all}, the histogram illustrates non-adherence across all patients, with a bimodal distribution.
A significant proportion of patients are on average fully adherent, averaging zero days of non-adherence per month.
Conversely, another cluster of patients demonstrates complete non-adherence, averaging nearly 30 days of non-adherence per month.
As per the approach taken in Section~\ref{sec:methods_data}, we use a binary treatment indicator in our study for non-adherence. 
Patients with less than 10 days of non-coverage in a month are considered to be adherent to their medication in that month and subsequently non-adherent if they have more than 10 days of non-coverage in that month (represented in blue and red bars respectively in Figure~\ref{fig:non_adherence_level}).

Figure~\ref{fig:non_adherence_injectable} focuses on patients prescribed injectable medications at least once in their trajectory.
These patients exhibit more balanced adherence patterns compared to the overall cohort, with fewer patients at the extremes of complete non-adherence and relatively the same proportion of patients show full adherence on average in their month as the complete cohort.
It should be noted that if a patient is administered with an injectable in a month, they cannot be fully non-adhered in that month (hence there is no bar at day 30).

In contrast, Subfigure~\ref{fig:non_adherence_non_injectable} examines patients prescribed only non-injectable medications.
This group shows a higher tendency toward complete non-adherence, with a more pronounced peak near 30 days of non-adherence. 
The complete adherence pattern however is similar to that of the full cohort with 8\% of patients showing complete adherence to their prescribed drugs.

The comparison between subgroups suggests that the type of prescribed medication plays a role in influencing adherence behavior.
Injectable medications may be associated with better adherence due to their administration requirements and monitoring, while non-injectable medications exhibit greater variability in patient adherence patterns.
However, in our cohort only 26\% of patients were ever administered an injectable medication throughout the time they were involved in this study.
For this reason, in future work, it is equally important to focus on interventions and strategies that address adherence challenges specific to patients prescribed non-injectable medications, as they constitute the majority of the cohort.

\subsubsection{Antipsychotic Medications}
\label{adp:cohort_medications}

\begin{figure*}[htbp]
\floatconts
  {fig:medications}
  {\caption{Prevalence of Antipsychotic Medications in the Patient Cohort. (a) Illustrates the percentage of patients prescribed each antipsychotic medication within the study cohort. The x-axis lists the medication names, and the y-axis represents the proportion of patients, expressed as a percentage. This visualization highlights the most frequently prescribed antipsychotic medications, with risperidone, aripiprazole, and olanzapine being the most prevalent. 
  (b), (c), and (d) Show the demographic composition of patients using each medication for (b) Legal Sex, (c) Race (c), and (d) Education Level. 
  (e) Shows the average age of patients using each antipsychotic medication.
  (f) Presents the mean continuous non-adherence scores for each antipsychotic medication. The y-axis values can range between 0 and 1 where a score of 1 indicates complete non-adherence, and a score of 0 represents full adherence.
}}
  {%
    \subfigure[Antipsychotic Medications Prevalence]{\label{fig:medication_prevalence}%
      \includegraphics[width=0.458\linewidth]{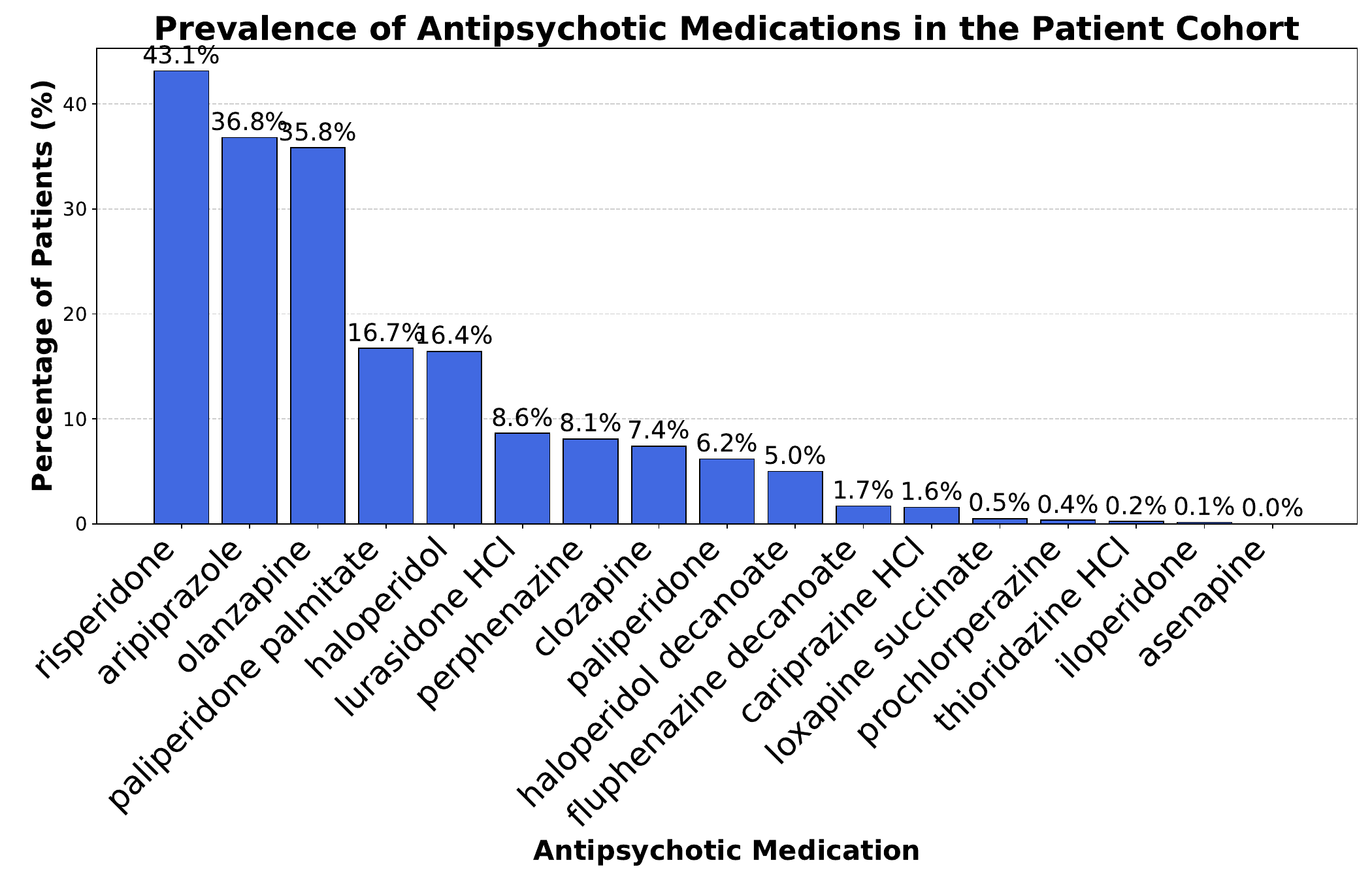}}%
    \qquad
    \subfigure[Gender]{\label{fig:medication_gender}%
      \includegraphics[width=0.458\linewidth]{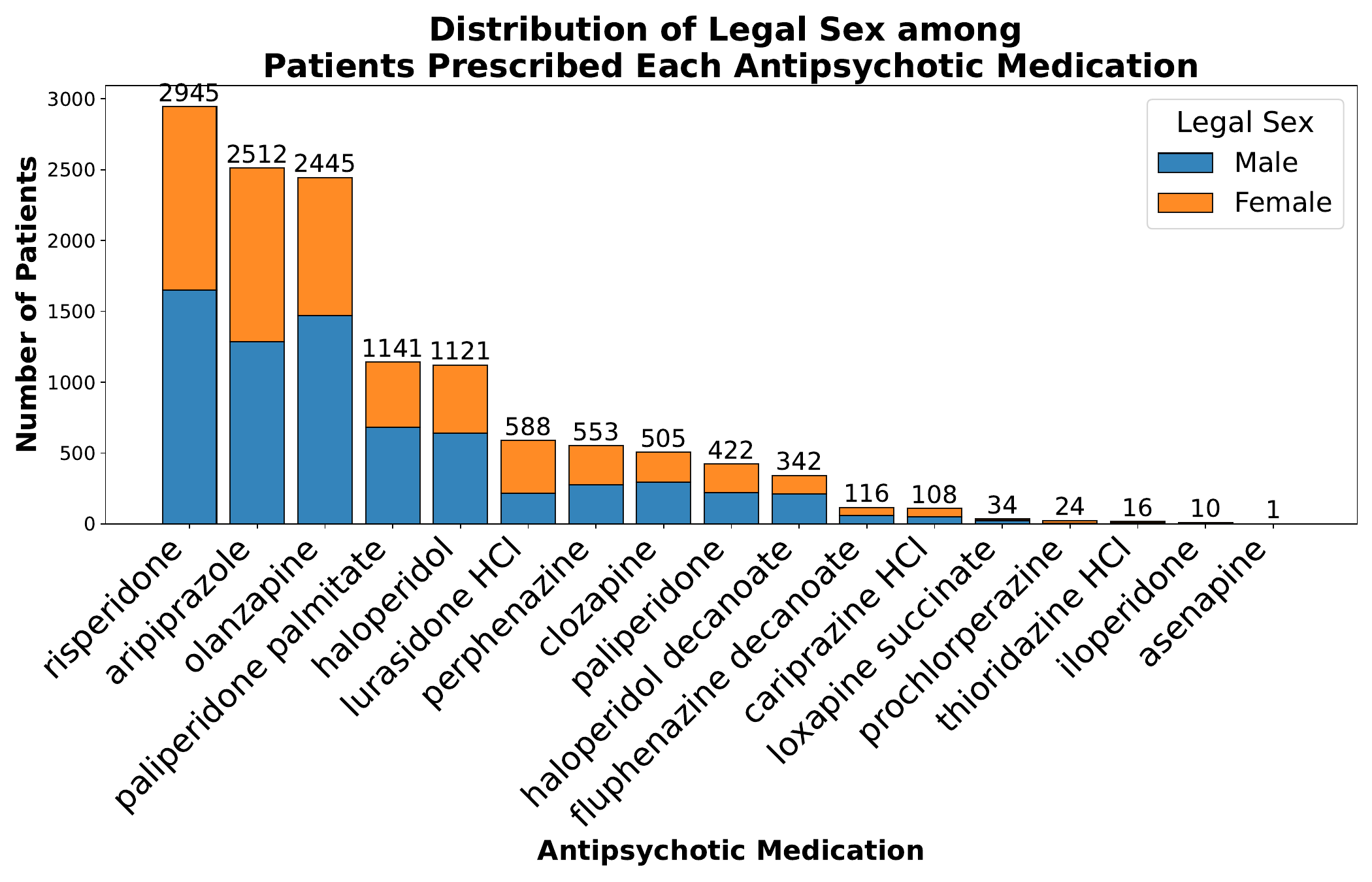}}%
    \\
    \subfigure[Race]{\label{fig:medication_race}%
      \includegraphics[width=0.458\linewidth]{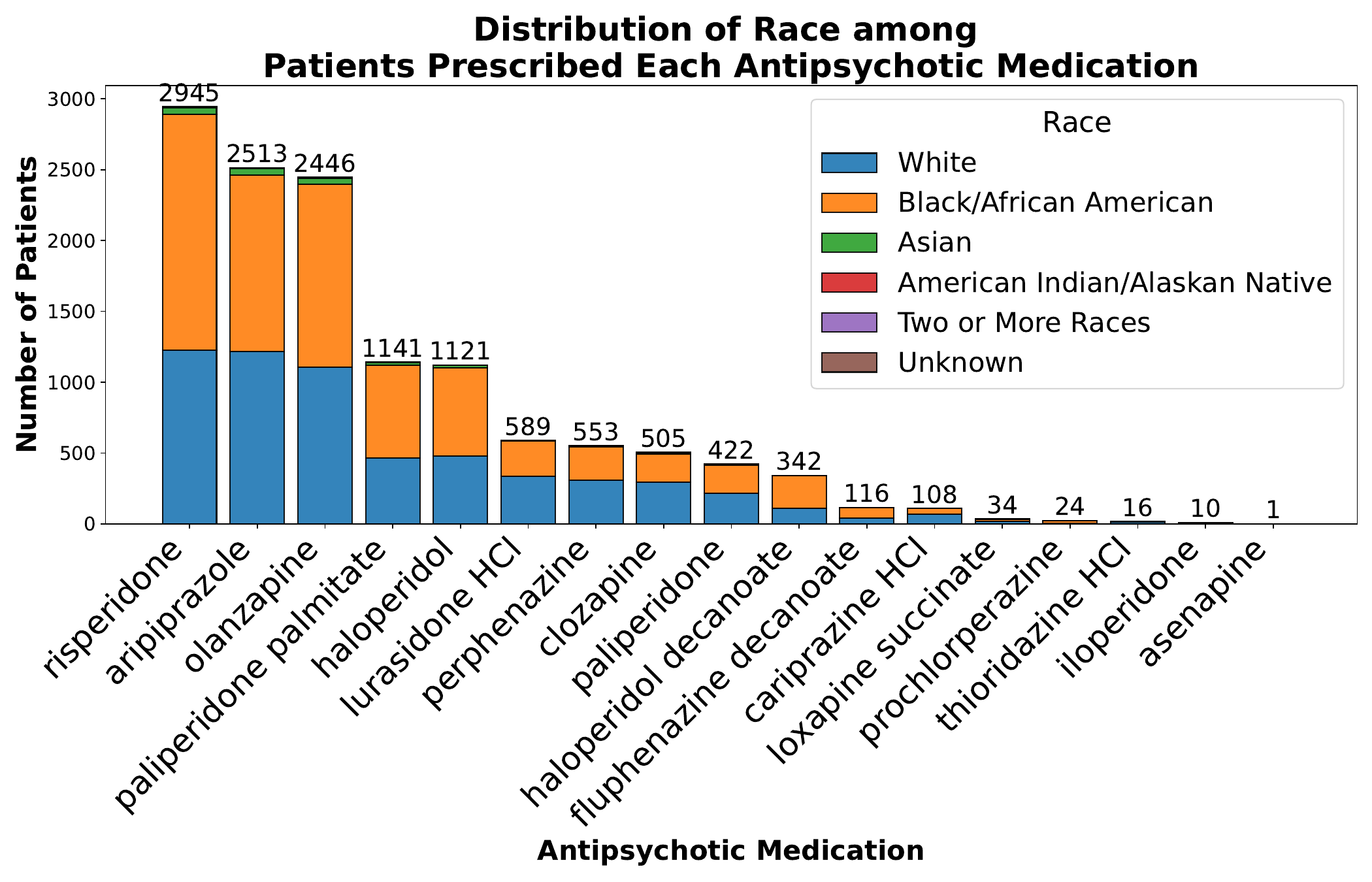}}%
    \qquad
    \subfigure[Education Level]{\label{fig:medication_education}%
      \includegraphics[width=0.458\linewidth]{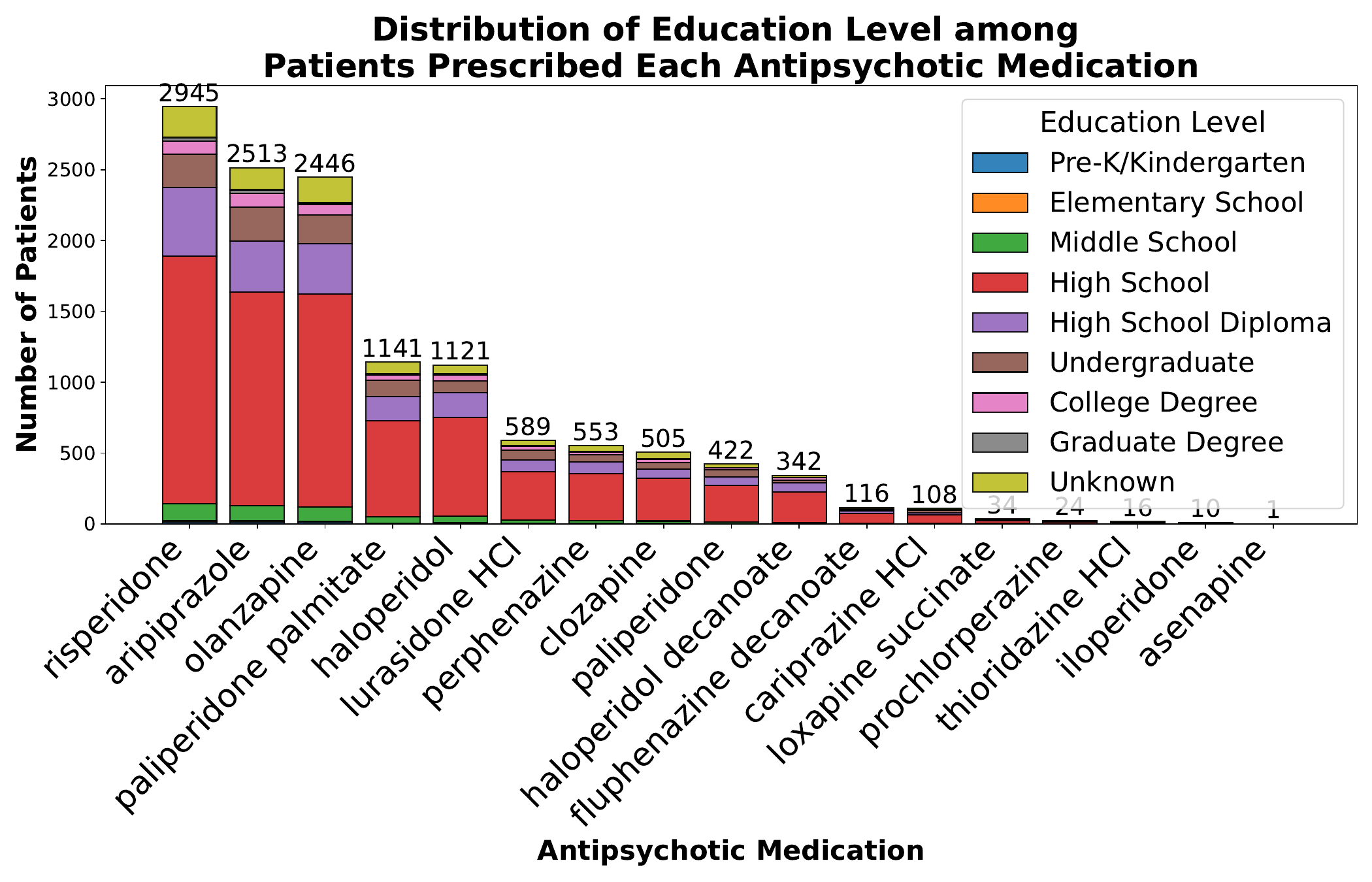}}%
    \\
    \subfigure[Age]{\label{fig:medication_age}%
      \includegraphics[width=0.458\linewidth]{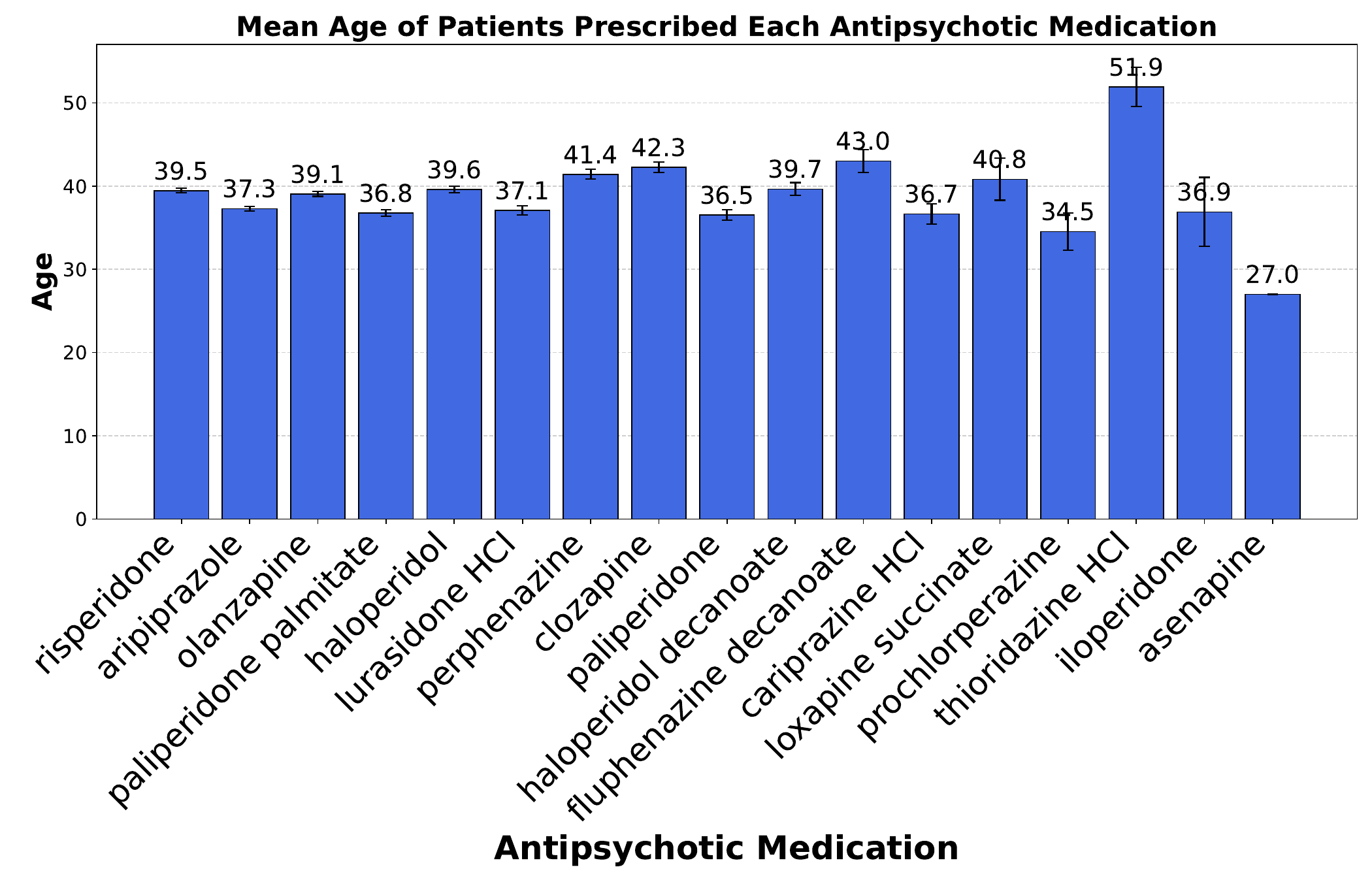}}%
    \qquad
        \subfigure[Non-Adherence Level]{\label{fig:medication_non_adherence}%
      \includegraphics[width=0.458\linewidth]{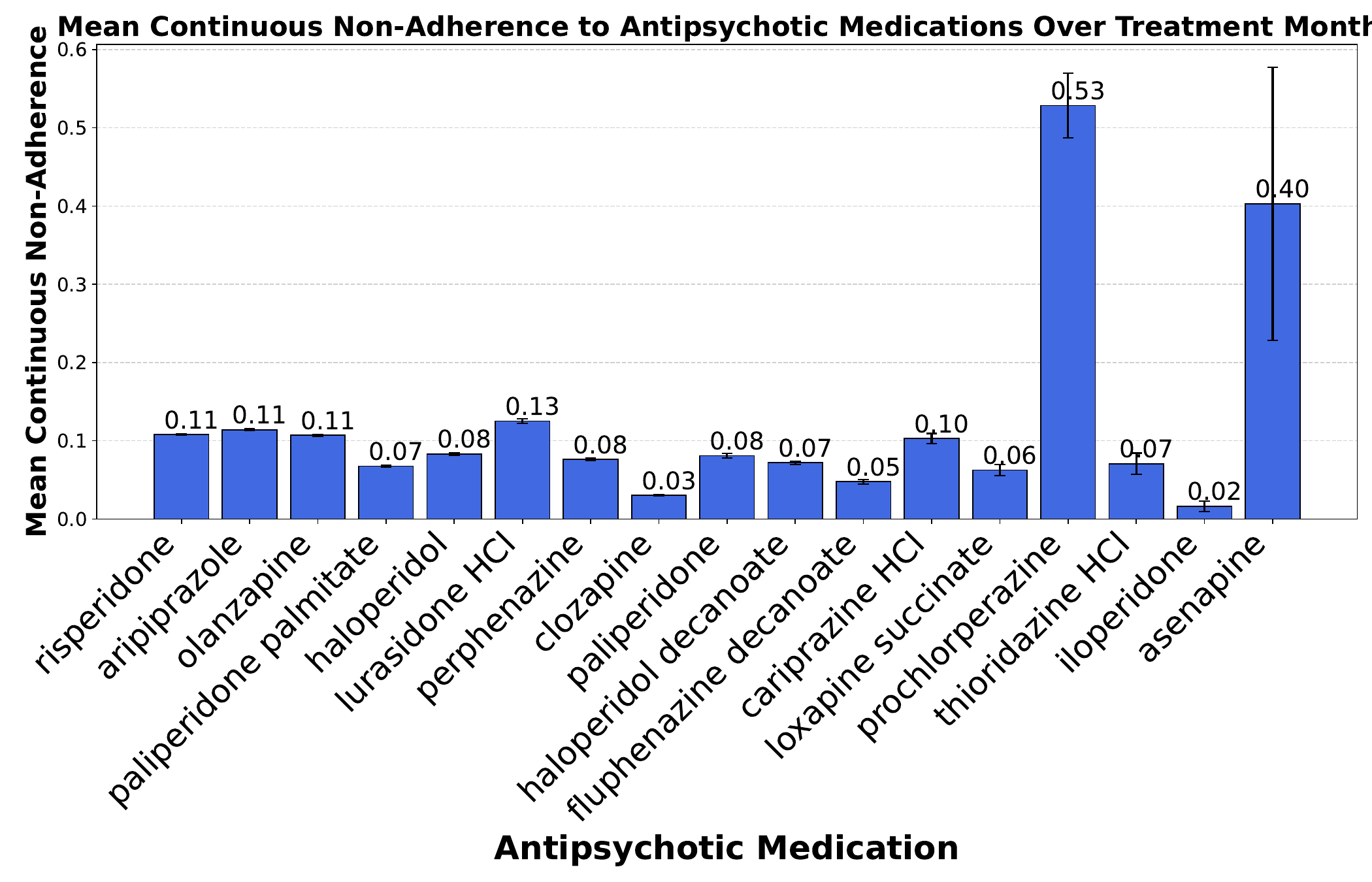}}%
  }
\end{figure*}

We examine the usage patterns of the antipsychotic medications within the patient cohort, considering factors such as prevalence of each medication, demographic distribution, adherence behavior, and other relevant characteristics in Figure~\ref{fig:medications}. 
Understanding these patterns provides insights into prescribing trends but also highlights potential challenges in treatment adherence and accessibility across diverse patient populations.
It should be noted that for all subfigure of Figure~\ref{fig:medications}, the x-axis is ordered from left to right based on the frequency of each medication being prescribed to the patients in our cohort with the risperidone being the most prevalent and asenapine being the least prevalent medication in our cohort of study.

Figure~\ref{fig:medication_prevalence} highlights the prevalence of antipsychotic medications in the study cohort. 
Risperidone emerges as the most commonly prescribed medication, with 43.1\% of patients receiving this treatment. 
It is closely followed by aripiprazole (36.8\%) and olanzapine (35.8\%). 
Paliperidone palmitate and haloperidol are also frequently prescribed, though at a lower prevalence of 16.7\% and 6.4\%, respectively. 
Other medications, such as loxapine and asenapine, exhibit minimal usage within the cohort, with fewer than 1\% of patients receiving these prescriptions. 
These results reflect the dominance of specific medications in treatment strategies for psychiatric conditions within this cohort.

Figure~\ref{fig:medication_gender} illustrates the distribution of legal sex among patients prescribed each antipsychotic medication. 
For most medications, the proportion of male patients exceeds that of female patients, consistent with the distribution seen in Appendix~\ref{apd:cohort_demographics}. 
Medications such as risperidone, aripiprazole, and olanzapine, which have the highest overall usage, exhibit similar male-to-female ratios. 
Interestingly, lurasidone HCl and perphenazine show slightly different distributions between the two genders (with the former being more predominant among females), potentially reflecting variations in prescribing practices or differences in the target patient populations for these drugs.

Figure~\ref{fig:medication_race} illustrates the racial demographics of patients prescribed various antipsychotic medications.
Black/African American patients, similar to the demographical composition of our cohort, represent the largest group for most medications, with particularly high representation for risperidone, aripiprazole, and olanzapine.
White patients constitute a sizable proportion of patients for many medications, including risperidone and aripiprazole, where their representation is nearly equal to that of Black/African American patients.
For certain medications, such as haloperidol and paliperidone palmitate, Black/African American patients are the clear majority group, demonstrating variations in prescribing trends.
Patients identifying as Asian, American Indian/Alaskan Native, or Two or More Races are minimally represented across all medications, and the proportion of patients with unknown racial information is similarly low.
These patterns may reflect differences in prescribing practices, patient demographics, or healthcare access within the study cohort.

Figure~\ref{fig:medication_education} shows the distribution of education levels among patients prescribed each antipsychotic medication. 
High school eduation are the most prevalent group for nearly all medications consistent with education level composition of our cohort with high school diploma also represented prominently. 
College education, graduate degree, and lower education levels, such as elementary or middle school, are much less frequent among the cohort. 
We do not see a particular trend suggesting a possible direct association between education level and medication type in our cohort.

Figure~\ref{fig:medication_age} presents the mean age of patients prescribed each antipsychotic medication. 
The mean age for most medications falls within the range of 36 to 43 years, consistent with the typical age of onset for psychiatric conditions treated with antipsychotics. 
Interestingly, patients prescribed thioridazine HCl tend to be older, with a mean age exceeding 50 years. 
In contrast, asenapine is associated with a significantly younger patient population, with a mean age of 27 years. 
These age differences may reflect the specific clinical indications or prescribing patterns for these medications.

Figure~\ref{fig:medication_non_adherence} explores the mean continuous non-adherence scores for each antipsychotic medication. 
Most medications, including risperidone, aripiprazole, and olanzapine, have low non-adherence scores (below 0.15), indicating relatively high adherence levels. 
However, certain medications, such as prochlorperazine and asenapine, exhibit markedly higher non-adherence scores of 0.53 and 0.40, respectively. 
It should be noted that these two medications are among the least frequent medications in our study and hence there is high uncertainty involved with their adherence levels in our analysis. 
Nevertheless, these preliminary findings highlight some challenges associated with maintaining adherence for specific medications, potentially due to side effects, dosing regimens, or other factors influencing patient compliance. 
Future research could focus on understanding the reasons behind these adherence disparities, including the role of patient characteristics, medication accessibility, and the severity of side effects. 

\vspace{-0.75em}
\subsection{Predictability of the Risk Scores in Our Data}
\label{apd:risk_score_evaluations}
\vspace{-0.25em}

\begin{figure}[htbp]
\floatconts
  {fig:roc_predictability_of_risks}
  {\caption{ROC curves assessing the accuracy of county-provided risk scores for predicting adverse events within 12 months. The AUC quantifies performance for (a) \textcolor{orange}{Involuntary Hospitalization} (b) \textcolor{red}{Jail Stay} and (c) \textbf{Mortality}.}}
  {%
    \subfigure[Involuntary Hospitalization]{\label{fig:roc_302_risk}%
      \includegraphics[width=.67\linewidth]{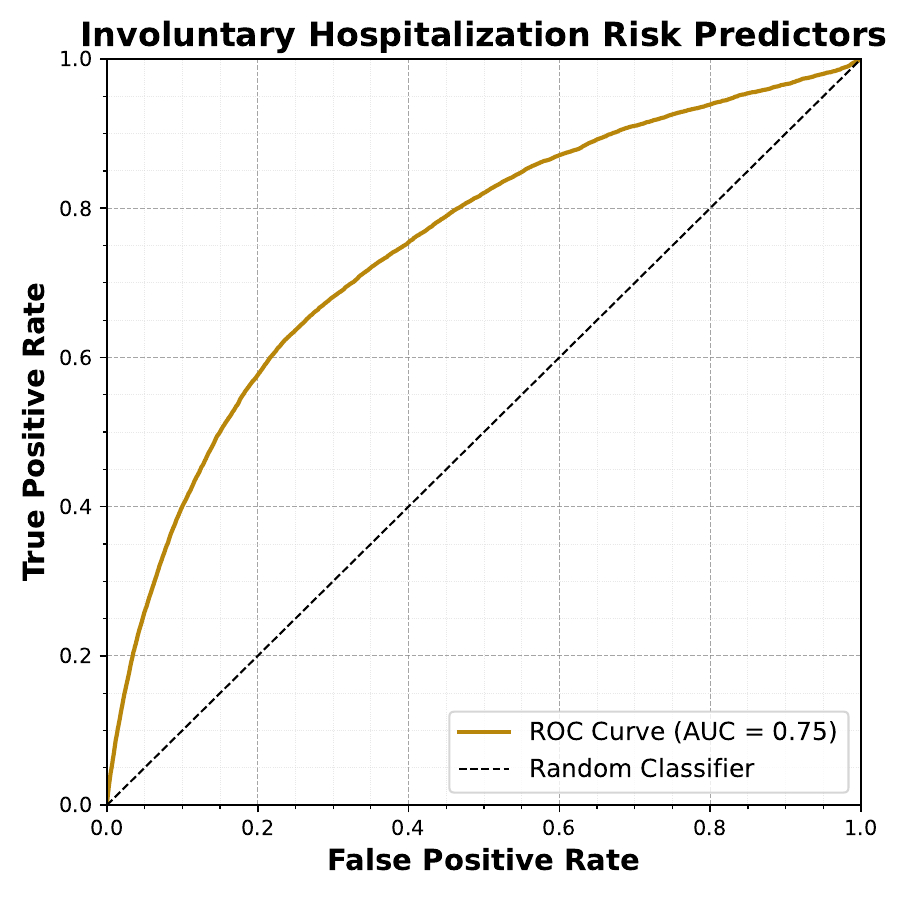}}%
    \\
    \subfigure[Jail Stay]{\label{fig:roc_jail_risk}%
      \includegraphics[width=.67\linewidth]{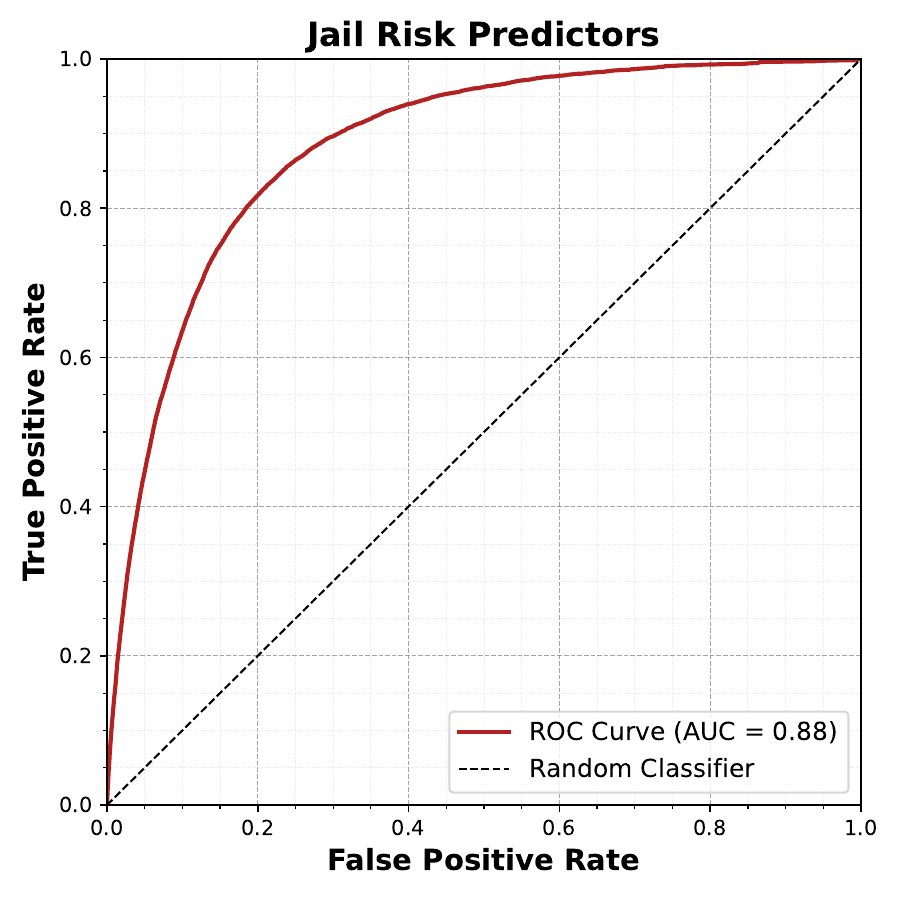}}
    \\
    \subfigure[Mortality]{\label{fig:roc_death_risk}%
      \includegraphics[width=.67\linewidth]{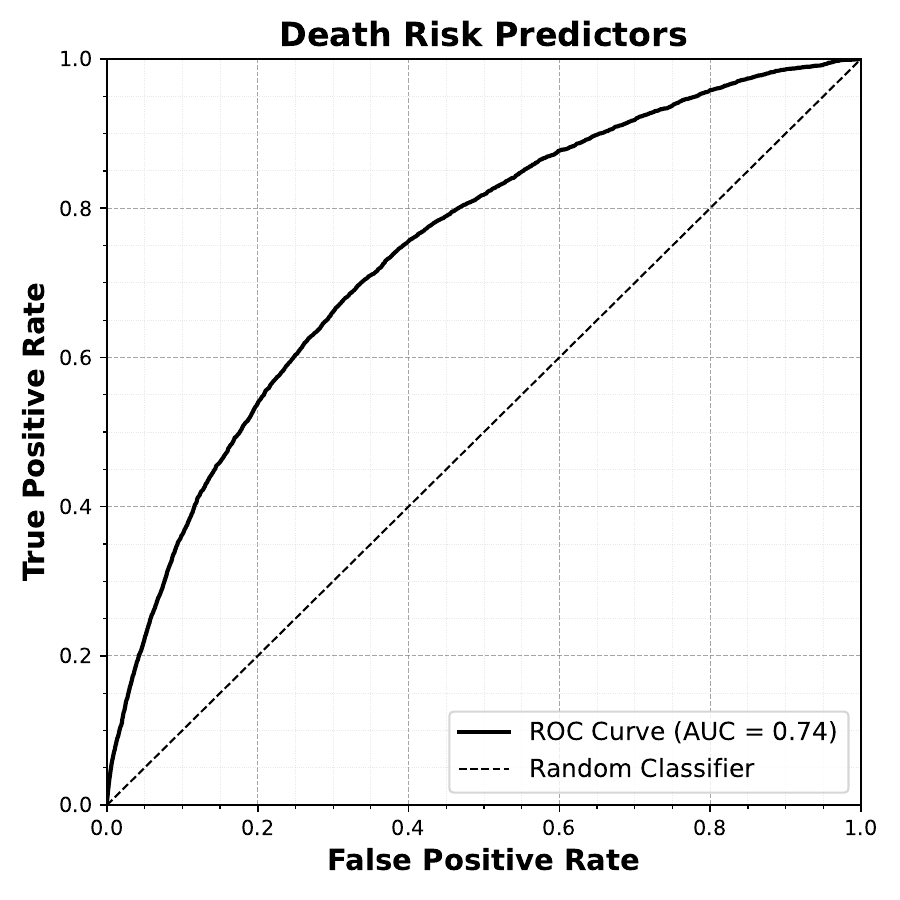}}
  }
\end{figure}

To evaluate the predictive accuracy of county-provided risk scores described in Section~\ref{sec:methods_data} in our data, we assess their ability to predict adverse events within 12 months.
The county-provided risk scores that we evaluate correspond to probabilities of three specific adverse events that we have access to: involuntary hospitalization, jail stay, and mortality.
Using ground truth outcomes available in our dataset, we compute Receiver Operating Characteristic (ROC) curves for each risk score and quantify performance using the Area Under the Curve (AUC).

Figure~\ref{fig:roc_predictability_of_risks} presents the ROC curves for the three adverse events.
For involuntary hospitalization (Figure~\ref{fig:roc_302_risk}), the risk score achieves an AUC of 0.75, indicating moderate predictive performance.
The jail stay risk score (Figure~\ref{fig:roc_jail_risk}) performs better, with an AUC of 0.88, reflecting strong discrimination between individuals at higher versus lower risk.
In contrast, the mortality risk score (Figure~\ref{fig:roc_death_risk}) achieves an AUC of 0.74, comparable to involuntary hospitalization, but still reflecting only moderate predictive accuracy.

These results suggest that while the county-provided risk scores capture some signal related to adverse events, their predictive performance varies significantly across outcomes.
Jail stay predictions are relatively reliable, whereas involuntary hospitalization and mortality scores exhibit room for improvement.
These results indicate that the county-provided risk scores capture some signal related to adverse events, though their performance varies across outcomes.
In this study, we incorporate these risk scores as covariates in our models, where they serve as proxies for unmeasured confounders.
Their inclusion helps to isolate the Average Treatment Effect (ATE) of non-adherence by accounting for confounding factors that would otherwise remain unobserved.
\newpage

\clearpage
\section{Preprocessing and Modeling Hyperparameters}
\label{apd:preprocessing_and_modeling_hyperparameters}
\numberwithin{equation}{section}
\numberwithin{figure}{section}
\numberwithin{table}{section}

\subsection{Data Preprocessing}
\label{apd:preprocessing}
This section outlines the detailed steps involved in preprocessing the data for our analysis.

The study cohort includes trajectories for 6,827 patients, where we identify the first composite adverse event for each patient.
For patients who experienced no adverse event, the time of the last available timestep was recorded.
At specific time snapshots \(\tau=\{3, 6, 9, 12\}\) months, we filtered the cohort to include only those patients with data available up to the snapshot time plus one additional month.
The survival analysis event or censoring time was defined as the time of the recorded event or censoring in the subset of the cohort, minus the snapshot time (i.e. time from the snapshot to the event/censoring).
We also generated a binary event indicator, where a value of 1 indicated the occurrence of a composite event (either involuntary hospitalization, jail stay, or death), and a value of 0 represented censoring.

Adherence levels were calculated for each month based on the methodology described in Section~\ref{sec:methods_data}.
Patients were classified as adherent ($A_{it}=0$) if they had less than 10 days of non-coverage in a month, and non-adherent ($A_{it}=1$) if they had more than 10 days of non-coverage in a month.

Static demographic covariates were extracted to generate the covariate matrix for patients.
We retained education level, race, and gender as static covariates while excluding ethnicity.
Any patient with a static covariate combination that appeared only once in the dataset (i.e., unique combinations) was removed from the analysis, resulting in the exclusion of 11 patients.
Age at the beginning of the study was included as an additional covariate.
We also included five county-provided risk scores at the snapshot time, the history of adherence up to (but not including) the snapshot time, and the binary adherence indicator for the snapshot time.

Continuous covariates were standardized using the training data, with the same normalization applied to validation and test sets.
Categorical covariates (those in the static covariate matrix), were one-hot encoded with the first category dropped for each variable to mitigate collinearity.

Lastly, we performed trimming to exclude patients whose static demographic information fully determined their treatment indicator (non-adherence at the snapshot time).
This trimming step resulted in cohort sizes of 5951 patients for \(\tau=3\) months, 5550 patients for \(\tau=6\) months, 5211 patients for \(\tau=9\) months, and 4893 patients for \(\tau=12\) months.

For the training and testing splits, we used an 80-20 split and repeated this process for five different random seeds to ensure robustness in experimental results.
The splits were applied consistently across all snapshots to enable a fair evaluation of our methods.

\subsection{Model Hyperparameters}
\label{apd:model_hyperparams}

This section provides the hyperparameters sets used for each survival analysis model implemented in our study.
We selected the best hyperparameter combination based on performance on a validation set.
The results presented in Section~\ref{sec:04_results_survival} are from a different test set using the best hyperparameter combination found.
The survival models in our study include Cox Proportional Hazards (CoxPH), Random Survival Forest (RSF), DeepSurv, and DeepHit.
The details of the hyperparameters grid for each model are summarized in Table~\ref{tab:model_hyperparameters}.

\begin{table}[htbp]
\centering
\scriptsize
\setlength{\tabcolsep}{4pt}
\caption{Set of Hyperparameters for Survival Analysis Models}
\label{tab:model_hyperparameters}
\begin{tabular}{l l c}
\hline
\textbf{Model}         & \textbf{Hyperparameter}             & \textbf{Values} \\ \hline
\multirow{1}{*}{CoxPH} & Penalizer                         & \{0, 0.01, 0.1, 0.5\} \\ \hline
\multirow{4}{*}{RSF}   & Number of estimators                & \{100, 250, 500\}   \\
                       & Minimum samples per split           & \{5, 10, 20\}       \\
                       & Minimum samples per leaf            & \{2, 5, 10\}        \\ \hline
\multirow{8}{*}{DeepHit} 
                       & Number of nodes per layer           & \{32, 64, 128, 256\} \\
                       & Batch normalization                 & \{True, False\}       \\
                       & Dropout rate                        & \{0.0, 0.1, 0.2, 0.3\} \\
                       & Learning rate                       & \{0.001, 0.01, 0.05\} \\
                       & Batch size                          & \{128, 256, 512\}    \\
                       & Epochs                              & \{200, 512, 1000\}   \\
                       & Alpha                               & \{0.1, 0.2, 0.3, 0.5\} \\
                       & Sigma                               & \{0.05, 0.1, 0.2, 0.3\} \\ \hline
\multirow{6}{*}{DeepSurv} 
                       & Number of nodes per layer           & \{32, 64, 128, 256\} \\
                       & Batch normalization                 & \{True, False\}       \\
                       & Dropout rate                        & \{0.0, 0.1, 0.2, 0.3\} \\
                       & Learning rate                       & \{0.001, 0.01, 0.05\} \\
                       & Batch size                          & \{128, 256, 512\}    \\
                       & Epochs                              & \{200, 512, 1000\}   \\ \hline
\end{tabular}
\end{table}

The hyperparameters in Table~\ref{tab:model_hyperparameters} were selected based on empirical tuning and prior research.
For all neural network-based models, early stopping was employed during training to prevent overfitting.
Random seeds were set to ensure reproducibility in experiments.

\end{document}